\documentclass{article}

\PassOptionsToPackage{square, numbers, compress}{natbib}
\usepackage[final]{neurips_data_2022}

\usepackage[utf8]{inputenc} % allow utf-8 input
\usepackage[T1]{fontenc}    % use 8-bit T1 fonts
\usepackage{hyperref}       % hyperlinks
\usepackage{url}            % simple URL typesetting
\usepackage{booktabs}       % professional-quality tables
\usepackage{amsfonts}       % blackboard math symbols
\usepackage{nicefrac}       % compact symbols for 1/2, etc.
\usepackage{microtype}      % microtypography
\usepackage{xcolor}         % colors
\usepackage{placeins}
\usepackage{hyperref}

\usepackage[draft]{minted}
\usepackage{listings}
\usepackage{xcolor}
\definecolor{codegreen}{rgb}{0,0.6,0}
\definecolor{codegray}{rgb}{0.5,0.5,0.5}
\definecolor{codepurple}{rgb}{0.58,0,0.82}
\definecolor{backcolour}{rgb}{0.95,0.95,0.92}

\lstdefinestyle{mystyle}{
    backgroundcolor=\color{backcolour},   
    commentstyle=\color{codegreen},
    keywordstyle=\color{magenta},
    numberstyle=\tiny\color{codegray},
    stringstyle=\color{codepurple},
    basicstyle=\ttfamily\footnotesize,
    breakatwhitespace=false,         
    breaklines=true,                 
    captionpos=b,                    
    keepspaces=true,                 
    numbers=left,                    
    numbersep=5pt,                  
    showspaces=false,                
    showstringspaces=false,
    showtabs=false,                  
    tabsize=2
}

\usepackage{amsmath}
\usepackage{amsfonts}
\usepackage{amssymb}
\usepackage{amsthm}
\usepackage{xcolor}
\usepackage{mathrsfs}
\usepackage{mathtools}
\usepackage{comment}

\usepackage{widetext}
\usepackage{gensymb}
\usepackage{amsmath}
\usepackage{amsmath,amsfonts,amssymb}
\usepackage{graphicx}
\usepackage{bm}
\usepackage{algorithm,algpseudocode,float}

\usepackage{threeparttable}
\usepackage{multirow}
\usepackage{booktabs}
\usepackage{tablefootnote}
\usepackage{array}
\usepackage{caption}
\usepackage{subcaption}
\usepackage{tabularx}
\usepackage{longtable}

\usepackage{mathtools}

\usepackage{bbm}

\usepackage{mathtools}

\usepackage{wrapfig}
\usepackage{tikz}
\usetikzlibrary{arrows.meta,positioning}

\usepackage{adjustbox}
\usepackage{array}

\usepackage{enumitem}
\newlist{eqlist}{enumerate*}{1}
\setlist[eqlist]{itemjoin=\quad,mode=unboxed,label=(\roman*),ref=\theequation(\roman*)}

\usepackage{enumitem}% http://ctan.org/pkg/enumitem

\usepackage[normalem]{ulem}

\usetikzlibrary{backgrounds}

\usepackage[para]{footmisc}

\usepackage[disable]{todonotes} 
\newcommand{\namednote}[3]{\todo[color=#1!20!white]{\textbf{#2:} #3}}
\newcommand{\mathias}[1]{\namednote{red}{Mathias}{#1}}

\newcommand{\dan}[1]{\namednote{purple}{Dan}{#1}}

\newcommand{\projectname}{\textsc{PDEBench}}

\usepackage{booktabs}

\makeatletter
\def\PYG@reset{\let\PYG@it=\relax \let\PYG@bf=\relax%
    \let\PYG@ul=\relax \let\PYG@tc=\relax%
    \let\PYG@bc=\relax \let\PYG@ff=\relax}
\def\PYG@tok#1{\csname PYG@tok@#1\endcsname}
\def\PYG@toks#1+{\ifx\relax#1\empty\else%
    \PYG@tok{#1}\expandafter\PYG@toks\fi}
\def\PYG@do#1{\PYG@bc{\PYG@tc{\PYG@ul{%
    \PYG@it{\PYG@bf{\PYG@ff{#1}}}}}}}
\def\PYG#1#2{\PYG@reset\PYG@toks#1+\relax+\PYG@do{#2}}

\@namedef{PYG@tok@w}{\def\PYG@tc##1{\textcolor[rgb]{0.73,0.73,0.73}{##1}}}
\@namedef{PYG@tok@c}{\let\PYG@it=\textit\def\PYG@tc##1{\textcolor[rgb]{0.24,0.48,0.48}{##1}}}
\@namedef{PYG@tok@cp}{\def\PYG@tc##1{\textcolor[rgb]{0.61,0.40,0.00}{##1}}}
\@namedef{PYG@tok@k}{\let\PYG@bf=\textbf\def\PYG@tc##1{\textcolor[rgb]{0.00,0.50,0.00}{##1}}}
\@namedef{PYG@tok@kp}{\def\PYG@tc##1{\textcolor[rgb]{0.00,0.50,0.00}{##1}}}
\@namedef{PYG@tok@kt}{\def\PYG@tc##1{\textcolor[rgb]{0.69,0.00,0.25}{##1}}}
\@namedef{PYG@tok@o}{\def\PYG@tc##1{\textcolor[rgb]{0.40,0.40,0.40}{##1}}}
\@namedef{PYG@tok@ow}{\let\PYG@bf=\textbf\def\PYG@tc##1{\textcolor[rgb]{0.67,0.13,1.00}{##1}}}
\@namedef{PYG@tok@nb}{\def\PYG@tc##1{\textcolor[rgb]{0.00,0.50,0.00}{##1}}}
\@namedef{PYG@tok@nf}{\def\PYG@tc##1{\textcolor[rgb]{0.00,0.00,1.00}{##1}}}
\@namedef{PYG@tok@nc}{\let\PYG@bf=\textbf\def\PYG@tc##1{\textcolor[rgb]{0.00,0.00,1.00}{##1}}}
\@namedef{PYG@tok@nn}{\let\PYG@bf=\textbf\def\PYG@tc##1{\textcolor[rgb]{0.00,0.00,1.00}{##1}}}
\@namedef{PYG@tok@ne}{\let\PYG@bf=\textbf\def\PYG@tc##1{\textcolor[rgb]{0.80,0.25,0.22}{##1}}}
\@namedef{PYG@tok@nv}{\def\PYG@tc##1{\textcolor[rgb]{0.10,0.09,0.49}{##1}}}
\@namedef{PYG@tok@no}{\def\PYG@tc##1{\textcolor[rgb]{0.53,0.00,0.00}{##1}}}
\@namedef{PYG@tok@nl}{\def\PYG@tc##1{\textcolor[rgb]{0.46,0.46,0.00}{##1}}}
\@namedef{PYG@tok@ni}{\let\PYG@bf=\textbf\def\PYG@tc##1{\textcolor[rgb]{0.44,0.44,0.44}{##1}}}
\@namedef{PYG@tok@na}{\def\PYG@tc##1{\textcolor[rgb]{0.41,0.47,0.13}{##1}}}
\@namedef{PYG@tok@nt}{\let\PYG@bf=\textbf\def\PYG@tc##1{\textcolor[rgb]{0.00,0.50,0.00}{##1}}}
\@namedef{PYG@tok@nd}{\def\PYG@tc##1{\textcolor[rgb]{0.67,0.13,1.00}{##1}}}
\@namedef{PYG@tok@s}{\def\PYG@tc##1{\textcolor[rgb]{0.73,0.13,0.13}{##1}}}
\@namedef{PYG@tok@sd}{\let\PYG@it=\textit\def\PYG@tc##1{\textcolor[rgb]{0.73,0.13,0.13}{##1}}}
\@namedef{PYG@tok@si}{\let\PYG@bf=\textbf\def\PYG@tc##1{\textcolor[rgb]{0.64,0.35,0.47}{##1}}}
\@namedef{PYG@tok@se}{\let\PYG@bf=\textbf\def\PYG@tc##1{\textcolor[rgb]{0.67,0.36,0.12}{##1}}}
\@namedef{PYG@tok@sr}{\def\PYG@tc##1{\textcolor[rgb]{0.64,0.35,0.47}{##1}}}
\@namedef{PYG@tok@ss}{\def\PYG@tc##1{\textcolor[rgb]{0.10,0.09,0.49}{##1}}}
\@namedef{PYG@tok@sx}{\def\PYG@tc##1{\textcolor[rgb]{0.00,0.50,0.00}{##1}}}
\@namedef{PYG@tok@m}{\def\PYG@tc##1{\textcolor[rgb]{0.40,0.40,0.40}{##1}}}
\@namedef{PYG@tok@gh}{\let\PYG@bf=\textbf\def\PYG@tc##1{\textcolor[rgb]{0.00,0.00,0.50}{##1}}}
\@namedef{PYG@tok@gu}{\let\PYG@bf=\textbf\def\PYG@tc##1{\textcolor[rgb]{0.50,0.00,0.50}{##1}}}
\@namedef{PYG@tok@gd}{\def\PYG@tc##1{\textcolor[rgb]{0.63,0.00,0.00}{##1}}}
\@namedef{PYG@tok@gi}{\def\PYG@tc##1{\textcolor[rgb]{0.00,0.52,0.00}{##1}}}
\@namedef{PYG@tok@gr}{\def\PYG@tc##1{\textcolor[rgb]{0.89,0.00,0.00}{##1}}}
\@namedef{PYG@tok@ge}{\let\PYG@it=\textit}
\@namedef{PYG@tok@gs}{\let\PYG@bf=\textbf}
\@namedef{PYG@tok@gp}{\let\PYG@bf=\textbf\def\PYG@tc##1{\textcolor[rgb]{0.00,0.00,0.50}{##1}}}
\@namedef{PYG@tok@go}{\def\PYG@tc##1{\textcolor[rgb]{0.44,0.44,0.44}{##1}}}
\@namedef{PYG@tok@gt}{\def\PYG@tc##1{\textcolor[rgb]{0.00,0.27,0.87}{##1}}}
\@namedef{PYG@tok@err}{\def\PYG@bc##1{{\setlength{\fboxsep}{\string -\fboxrule}\fcolorbox[rgb]{1.00,0.00,0.00}{1,1,1}{\strut ##1}}}}
\@namedef{PYG@tok@kc}{\let\PYG@bf=\textbf\def\PYG@tc##1{\textcolor[rgb]{0.00,0.50,0.00}{##1}}}
\@namedef{PYG@tok@kd}{\let\PYG@bf=\textbf\def\PYG@tc##1{\textcolor[rgb]{0.00,0.50,0.00}{##1}}}
\@namedef{PYG@tok@kn}{\let\PYG@bf=\textbf\def\PYG@tc##1{\textcolor[rgb]{0.00,0.50,0.00}{##1}}}
\@namedef{PYG@tok@kr}{\let\PYG@bf=\textbf\def\PYG@tc##1{\textcolor[rgb]{0.00,0.50,0.00}{##1}}}
\@namedef{PYG@tok@bp}{\def\PYG@tc##1{\textcolor[rgb]{0.00,0.50,0.00}{##1}}}
\@namedef{PYG@tok@fm}{\def\PYG@tc##1{\textcolor[rgb]{0.00,0.00,1.00}{##1}}}
\@namedef{PYG@tok@vc}{\def\PYG@tc##1{\textcolor[rgb]{0.10,0.09,0.49}{##1}}}
\@namedef{PYG@tok@vg}{\def\PYG@tc##1{\textcolor[rgb]{0.10,0.09,0.49}{##1}}}
\@namedef{PYG@tok@vi}{\def\PYG@tc##1{\textcolor[rgb]{0.10,0.09,0.49}{##1}}}
\@namedef{PYG@tok@vm}{\def\PYG@tc##1{\textcolor[rgb]{0.10,0.09,0.49}{##1}}}
\@namedef{PYG@tok@sa}{\def\PYG@tc##1{\textcolor[rgb]{0.73,0.13,0.13}{##1}}}
\@namedef{PYG@tok@sb}{\def\PYG@tc##1{\textcolor[rgb]{0.73,0.13,0.13}{##1}}}
\@namedef{PYG@tok@sc}{\def\PYG@tc##1{\textcolor[rgb]{0.73,0.13,0.13}{##1}}}
\@namedef{PYG@tok@dl}{\def\PYG@tc##1{\textcolor[rgb]{0.73,0.13,0.13}{##1}}}
\@namedef{PYG@tok@s2}{\def\PYG@tc##1{\textcolor[rgb]{0.73,0.13,0.13}{##1}}}
\@namedef{PYG@tok@sh}{\def\PYG@tc##1{\textcolor[rgb]{0.73,0.13,0.13}{##1}}}
\@namedef{PYG@tok@s1}{\def\PYG@tc##1{\textcolor[rgb]{0.73,0.13,0.13}{##1}}}
\@namedef{PYG@tok@mb}{\def\PYG@tc##1{\textcolor[rgb]{0.40,0.40,0.40}{##1}}}
\@namedef{PYG@tok@mf}{\def\PYG@tc##1{\textcolor[rgb]{0.40,0.40,0.40}{##1}}}
\@namedef{PYG@tok@mh}{\def\PYG@tc##1{\textcolor[rgb]{0.40,0.40,0.40}{##1}}}
\@namedef{PYG@tok@mi}{\def\PYG@tc##1{\textcolor[rgb]{0.40,0.40,0.40}{##1}}}
\@namedef{PYG@tok@il}{\def\PYG@tc##1{\textcolor[rgb]{0.40,0.40,0.40}{##1}}}
\@namedef{PYG@tok@mo}{\def\PYG@tc##1{\textcolor[rgb]{0.40,0.40,0.40}{##1}}}
\@namedef{PYG@tok@ch}{\let\PYG@it=\textit\def\PYG@tc##1{\textcolor[rgb]{0.24,0.48,0.48}{##1}}}
\@namedef{PYG@tok@cm}{\let\PYG@it=\textit\def\PYG@tc##1{\textcolor[rgb]{0.24,0.48,0.48}{##1}}}
\@namedef{PYG@tok@cpf}{\let\PYG@it=\textit\def\PYG@tc##1{\textcolor[rgb]{0.24,0.48,0.48}{##1}}}
\@namedef{PYG@tok@c1}{\let\PYG@it=\textit\def\PYG@tc##1{\textcolor[rgb]{0.24,0.48,0.48}{##1}}}
\@namedef{PYG@tok@cs}{\let\PYG@it=\textit\def\PYG@tc##1{\textcolor[rgb]{0.24,0.48,0.48}{##1}}}

\def\PYGZus{\char`\_}

\def\PYGZhy{\char`\-}

\def\PYGZdq{\char`\"}

\makeatother

\usepackage[american]{babel}
\usepackage{mathtools} % amsmath with fixes and additions
\usepackage{booktabs} % commands to create good-looking tables
\usepackage{tikz} % nice language for creating drawings and diagrams

\title{\projectname{}: An Extensive Benchmark for Scientific Machine Learning}

\author{%
  Makoto~Takamoto\thanks{E-mail:{\texttt{Makoto.Takamoto@neclab.eu}}}\\
  NEC Labs Europe \\
  \And
  Timothy~Praditia\thanks{E-mail:\texttt{timothy.praditia@iws.uni-stuttgart.de}}\\
  University of Stuttgart\\ %\\
  \And
  Raphael~Leiteritz\\
  University of Stuttgart %\\
  \AND
    Dan~MacKinlay \\
  CSIRO's Data61 
  \And
  Francesco~Alesiani\\
  NEC Labs Europe \\
  \AND
  Dirk~Pflüger \\
  University of Stuttgart %\\
  \And
  Mathias~Niepert \\
  University of Stuttgart %\\
}

\begin{document}

\maketitle

\begin{abstract}
Machine learning-based modeling of physical systems has experienced increased interest in recent years. 
Despite some impressive progress, there is still a lack of  benchmarks for Scientific ML that are easy to use but still challenging and representative of a wide range of problems.
We introduce \projectname{}, a benchmark suite of time-dependent simulation tasks based on Partial Differential Equations (PDEs). 
\projectname{} comprises both code and data to benchmark the performance of novel machine learning models against both classical numerical simulations and machine learning baselines.
Our proposed set of benchmark problems contribute the following unique features: 
(1) A much wider range of PDEs compared to existing benchmarks, ranging from relatively common examples to more realistic and difficult problems; 
(2) much larger ready-to-use datasets compared to prior work, comprising multiple simulation runs across a larger number of initial and boundary conditions and PDE parameters; 
(3) more extensible source codes with user-friendly APIs for data generation and baseline results with popular machine learning models (FNO, U-Net, PINN, 
Gradient-Based Inverse Method). 
\projectname{} allows researchers to extend the benchmark freely for their own purposes using a standardized API and to compare the performance of new models to existing baseline methods. 
We also propose new evaluation metrics with the aim to provide a more holistic understanding of learning methods in the context of Scientific ML. 
With those metrics we identify tasks which are challenging for recent ML methods and propose these tasks as future challenges for the community.
The code is available at \url{https://github.com/pdebench/PDEBench}.

\end{abstract}

\section{Motivation}
 \label{sec:motivation}

\begin{figure}[t!]
    \centering
        \begin{subfigure}{0.2\textwidth}
         \centering
         \includegraphics[width=\textwidth]{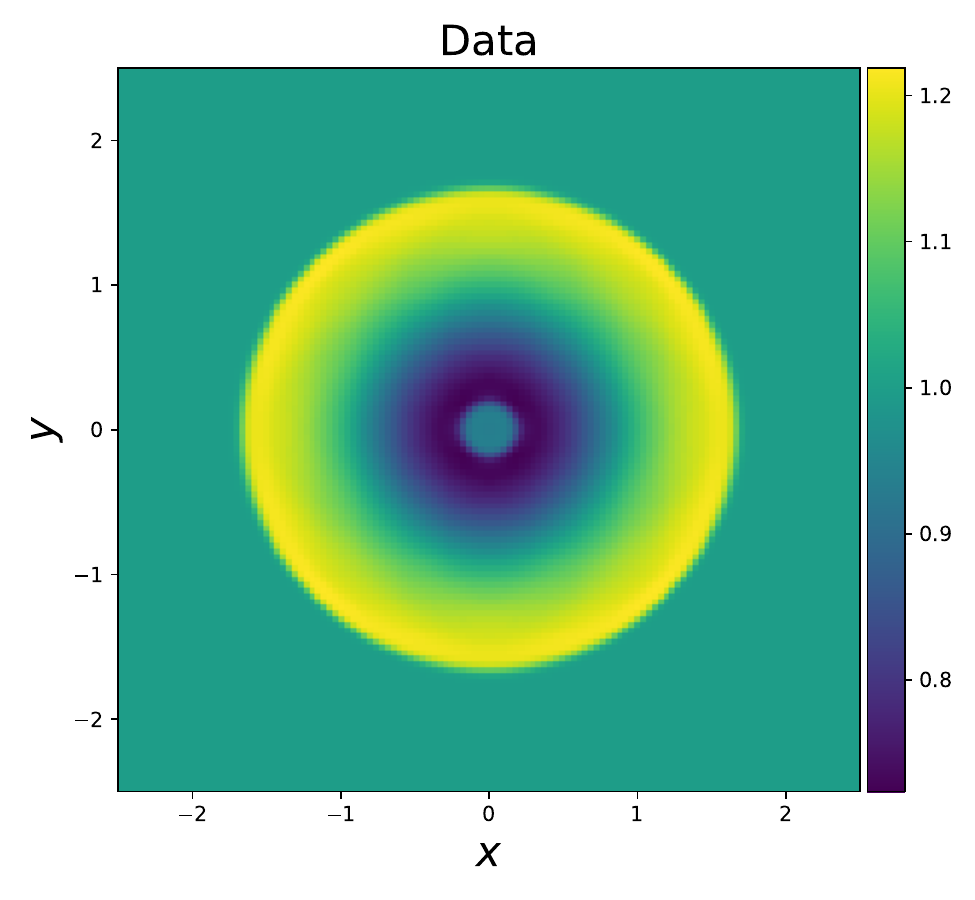}
         \caption{}
     \end{subfigure}
     \hfill
         \begin{subfigure}{0.21\textwidth}
         \centering
         \includegraphics[width=\textwidth]{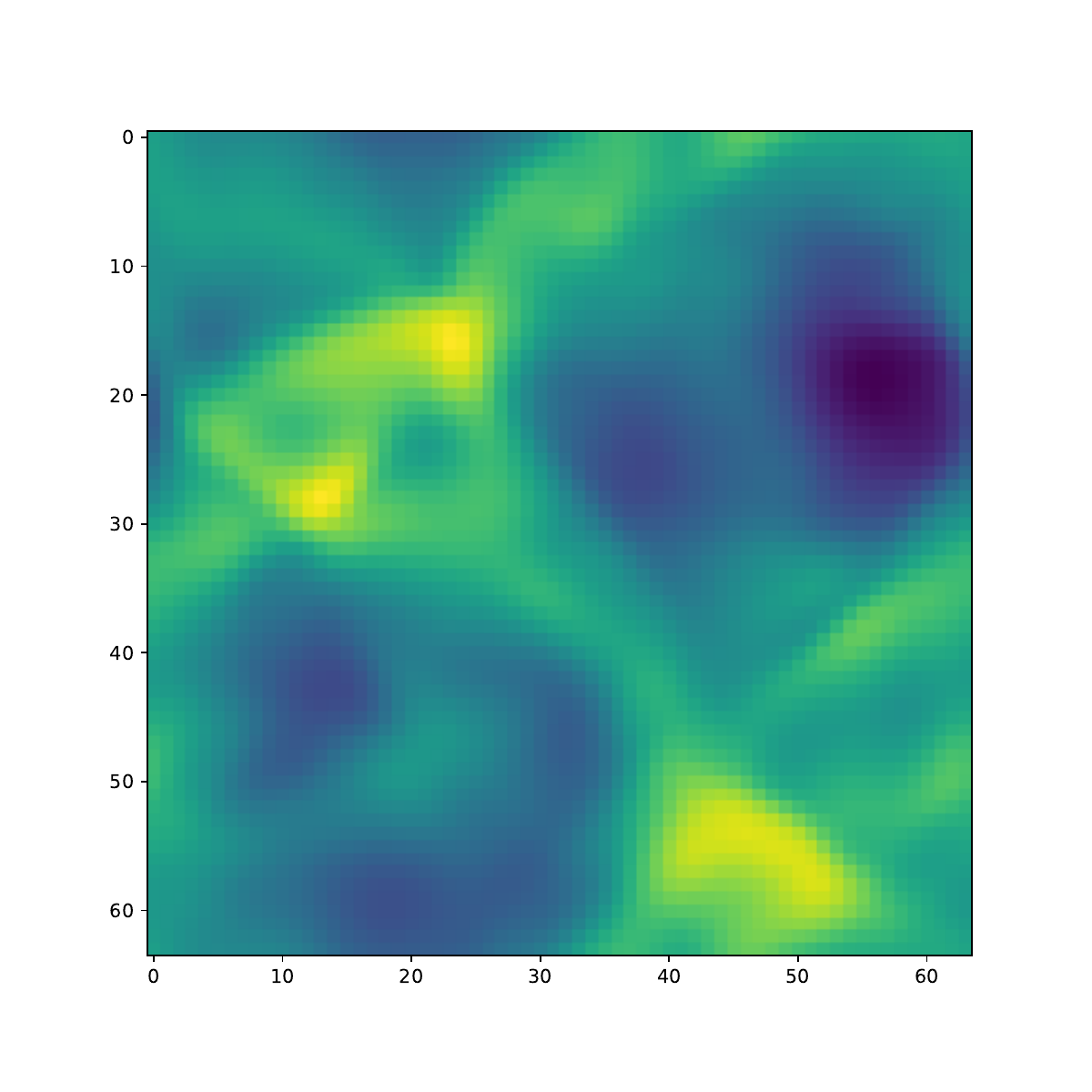}
         \caption{}
     \end{subfigure}
     \hfill
    \begin{subfigure}{0.17\textwidth}
         \centering
         \includegraphics[width=\textwidth]{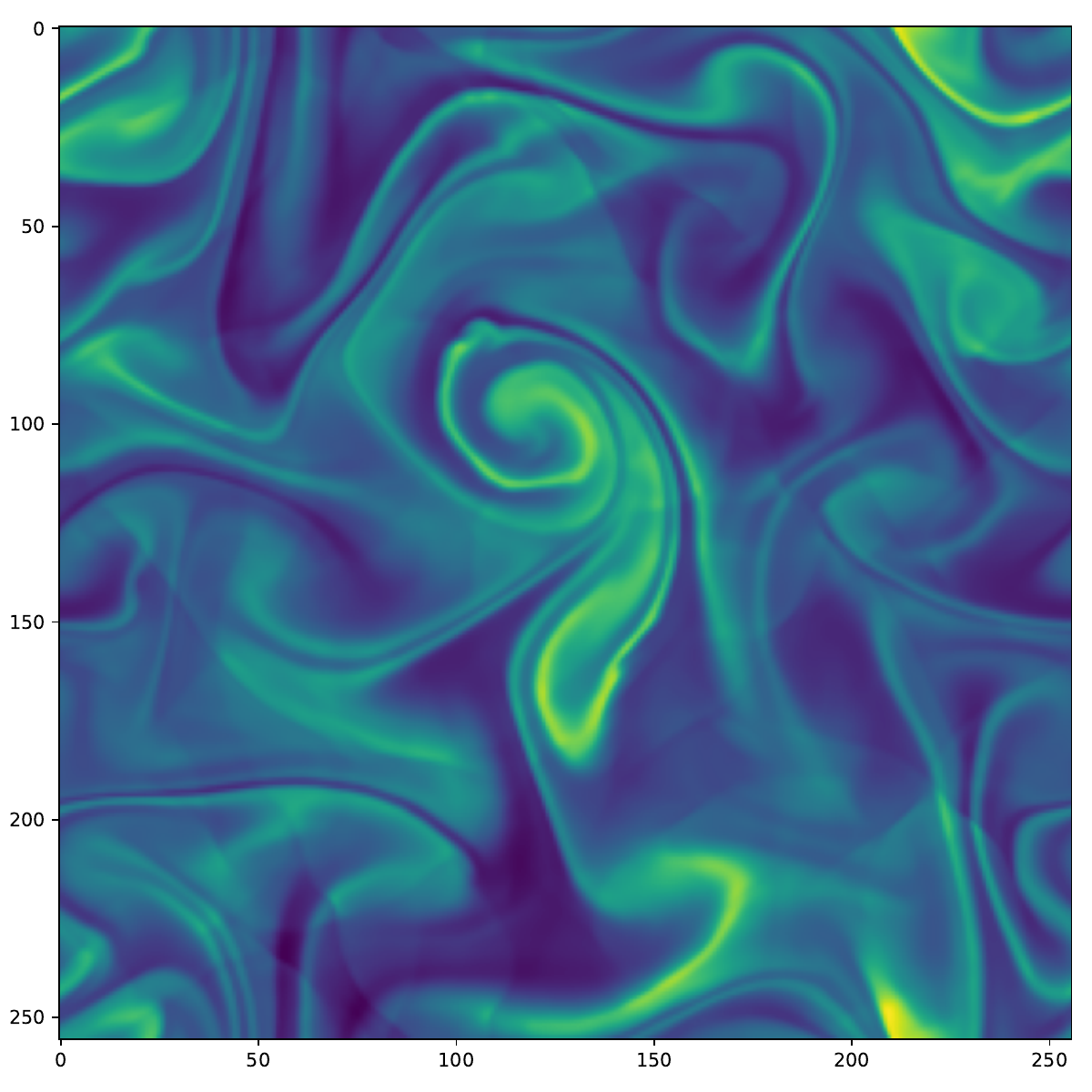}
         \caption{}
     \end{subfigure}
     \hfill
    \begin{subfigure}{0.2\textwidth}
         \centering
         \includegraphics[width=\textwidth]{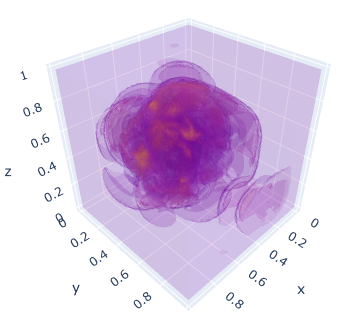}
         \caption{}
     \end{subfigure}
     \vspace{-2em} % REMOVING SUBFIGURE LABELS
    \caption{\projectname{} provides multiple non-trivial challenges from the Sciences to benchmark current and future ML methods, including wave propagation and turbulent flow in 2D and 3D}
    \label{fig:gallery}
\end{figure}

In the emergent area of \emph{Scientific Machine Learning} (or \emph{machine learning for physical sciences} or \emph{data-driven science}), recent progress has broadened the scope of traditional machine learning (ML) methods to include the time-evolution of physical systems.
Within this field, rapid progress has been made in the use of neural networks to make predictions using functional observations over continuous domains~\citep{ChenNeural2018,RackauckasEssential2019} or with challenging constraints and with physically-motivated conservation laws~\citep{lu2021deepxde,WangPhysicsinformed2020,RaissiPhysicsinformed2019}.
These neural networks provide an approach to solving PDEs complementing traditional numerical solvers. For instance, data-driven ML methods are useful when observations are noisy or the underlying physical model is not fully known or defined~\cite{Eivazi_Tahani_Schlatter_Vinuesa_2021}.
Moreover, neural models have the advantage of being continuously differentiable in their inputs, a useful property in several applications. In physical system design~\cite{allen2022physical}, for instance, the models are themselves physical objects and thus not analytically differentiable.
Similarly, in many fields such as  hydrology~\cite{GladishEmulation2018}, benchmark physical simulation models exist but the forward simulation models non-differentiable black boxes. 
This complicates optimisation, control, sensitivity analysis, and solving inverse inference problems.
While complex methods such as Bayesian optimisation~\cite{MockusBayesian1975,SnoekPractical2012,OHaganCurve1978} or reduced order modelling~\cite{guo201975} are in part an attempt to circumvent this lack of differentiability, gradients for neural networks are readily available and efficient.

For classical ML applications such as image classification, time series prediction or text mining, various popular benchmarks exist, and evaluations using these benchmarks provides a standardised means of testing the effectiveness and efficiency of ML models.
As yet, a widely accessible, practically simple, and statistically challenging benchmark with ready-to-use datasets to compare methods in Scientific ML is missing.
While some progress towards reference benchmarks has been made in recent years (see section \ref{related-work}), we aim to provide a benchmark  that is more comprehensive with respect to the PDEs covered and which enables more diverse methods for evaluating the efficiency and accuracy of the ML method.
The problems span a range of governing equations as well as different assumptions and conditions; see \autoref{fig:gallery} for a visual teaser. Data may be generated by executing code through a common interface, or by downloading high-fidelity datasets of simulations. 
All code is released under a permissive open source license, facilitating re-use and extension. We also propose an API to ease the implementation and evaluation of new methods, provide recent competitive baseline methods such as FNOs and autoregressive models based on the U-Net, and a set of pre-computed performance metrics for these algorithms.
We may thus compare their predictions against the ``ground truth'' provided by baseline simulators used to generate the data.

As in other machine learning application domains, benchmarks in Scientific ML may serve as a source of readily-available training data for algorithm development and testing without the overhead of generating data \emph{de novo}.
In these emulation tasks, the training/test data is notionally unlimited, since more data may always be generated by running a simulator.
However, in practice, producing such datasets can have an extremely high burden in compute time, storage and in access to the specialised skills needed to produce them.
\projectname{} also addresses the need for quick, off-the-shelf training data, bypassing these barriers while providing an easy on-ramp to be extended.

In this work, we propose a versatile benchmark suite for Scientific ML (a) providing diverse data sets with distinct properties based on 11 well-known time-dependent and time-independent PDEs, (b) covering both ``classical'' forward learning problems and inverse settings, (c) all accessible via a uniform interface to read/store data across several applications, (d) extensible, (e) with results for popular state-of-the-art ML models (FNO, U-Net, PINN) for (f) a set of metrics that are better-suited for Scientific ML, (g) with both data to download and code to generate more data, and (h) pre-trained models to compare against. The inverse problem scenarios comprise initial and boundary conditions and PDE parameters (e.g.\ viscosity). Each data set has a sufficiently large number of samples for training and test, for a variety of parameter values, with a resolution high enough to capture local dynamics.  As an additional note, our goal is not to provide a complete benchmark that includes all possible combinations of inference tasks on all known experiments, but rather to ease the task for subsequent researchers to benchmark their favoured methods. Part of our goal here is to invite other researchers to fill in the gaps for themselves by leveraging our ready-to-run models.

To evaluate ML methods for scientific problems, we consider several metrics that go beyond the standard RMSE and include properties of the underlying physics.
The initial experimental results we obtained using \projectname{} confirm the need for comprehensive Scientific ML benchmarks: There is no one-size-fits-all model and there is plenty of room for new ML developments. The results show that the standard error measure in ML, the RMSE on test data, is not a good proxy for evaluating ML models, in particular in turbulent and non-smooth regimes where it fails to capture small spatial scale changes. We furthermore cover an application where a parameter of the underlying PDE heavily influences the difficulty of the problem for ML baselines. We also observe unexpected experimental results for the generalization behavior of auto-regressive training, which seems to be more challenging in the scientific realm. In the remainder of the paper, we first address related work, introduce \projectname{} and the underlying design choices, and discuss results for a few selected experiments.

\section{Related Work}
\label{related-work}

PDE benchmarking has particular challenges.
Unlike many classic datasets, PDE datasets can be large on a gigabyte or terabyte scale and still contain only few data points.
And unlike monolithic benchmark datasets such as ImageNet, the datasets for each PDE approximation task are specific to that task.
Each set of governing equations or experiment design assumptions leads to a distinct dataset of PDE samples.
Recent works in PDEs have attempted to produce standardised datasets covering well-known challenges \citep{otness2021extensible,huang2021largescale,stachenfeld2022coarse}.
\cite{huang2021largescale} targets non-ML uses.
\cite{stachenfeld2022coarse} is specialised for particular classes of equations.
Of these, the excellent work of \cite{otness2021extensible} is most closely related, but with only four physical systems, it still lacks sufficient scale and diversity of data to challenge emerging ML algorithms.
We expand the range of benchmarks in this domain by providing a larger, more diverse problem selection and scale than these previous attempts (11 PDEs with different parametrizations leading to 35 datasets). 
We additionally consider inverse problems for PDEs \citep{StuartInverse2010,TarantolaInverse2005}, with the goal to identify unobserved latent parameters using ML. This has not been covered by benchmarks so far, despite its increasing importance in the community. 
Furthermore, most work in this scope considers classical statistical error measures such as the RMSE over the whole domain and at most PDE-motivated variants such as the RMSE of the gradient~\cite{otness2021extensible}. Measures based on properties of the underlying physical systems, as studied in this work, are lacking.

An overview and taxonomy of Scientific ML developments can be found in \cite{LavinSimulation2021,BruntonDatadriven2019}.
For developing our baselines, we focus on using neural network models to approximate the outputs of some \emph{ground truth} PDE solver, given data generated by that solver, which itself aims to directly implement the numerical solution of a given partial differential equation.
A range of methods aim to solve problems fitting this description, reviewed in \cite{KashinathPhysicsinformed2021}.
Methods include Physics-informed neural networks (PINNs) \citep{RaissiPhysicsinformed2019}, Neural operators (NOs) \citep{li2020fourier,KovachkiNeural2021}, treating ResNet as a PDE approximant \citep{RuthottoDeep2018}, custom architectures for specific problems such as TFNet for turbulent fluid flows \citep{WangPhysicsinformed2020}, and generic image-to-image regression models such as the U-Net \citep{RonnebergerUNet2015}.
These approaches each have different assumptions, domains of applicability, and data processing requirements. For a more comprehensive discussion of prior work in Scientific ML we refer the reader to Appendix~\ref{sec:ap:related}.

\section{\projectname{}: A Benchmark for Scientific Machine Learning}
\label{sec:pdebench}

In the following we describe the general learning problem addressed with the benchmark, the currently covered PDEs, existing implemented baselines (all developed using PyTorch~\citep{pytorch}, and PINN specifically using DeepXDE~\citep{lu2021deepxde}), and the ways in which the benchmark follows FAIR data principles \cite{wilkinson2016fair}.

\subsection{General Problem Definition}

A solution to a PDE is a vector-valued function \(\textbf{v}:\mathcal{T}\times\mathcal{S}\times\Theta \rightarrow \mathbb{R}^{d} \) \mathias{isn't it better to actually have the arrow to some vector space? it has confused a reviewer} on some spatial domain \(\mathcal{S}\), with temporal index \(\mathcal{T}\), and some possibly function-valued parameter-space \(\Theta\).
For example in a heat diffusion equation, \(\textbf{v}\) might represent the local temperature $\tau \in \mathbb{R}^{1}$ of some substrate at some given point \(\mathbf{s}\in\mathcal{S}\), at a given moment \(t\in\mathcal{T}\) , and conditional upon a spatially-varying scalar conductivity field representing an inhomogeneous substrate \(\theta: \mathcal{S}\to \mathbb{R}^+\).
The operator mapping from the state of the solution at one timestep to the solution one time step later,  \(\mathfrak{F}_\theta:\mathbf{v}_{\theta}(t,\cdot) \to \mathbf{v}_{\theta}(t+1,\cdot)\) is referred to as the \emph{forward propagator}.

The objective of Scientific ML is to find some ML-based surrogate, sometimes referred to as an emulator, of this forward propagator by learning an approximation \(\widehat{\mathfrak{F}}_\theta\simeq \mathfrak{F}_\theta\).
The forward propagator of a PDE is dependent not only on the current state, but also upon both spatial and temporal derivatives of the state field.
In practice, temporal derivatives of solutions are often not conveniently encoded by system states at one single time step. Hence, the forward propagator may also depend on multiple previous timesteps of the solution, enabling finite-difference approximations of the temporal derivatives.
The discretised forward propagator \(\mathring{\mathfrak{F}}_\theta\) then operates on \(\ell\geq 1\) consecutive timesteps so that  \(\mathring{\mathfrak{F}}_\theta:\mathbf{v}_{\theta}(t-\ell,\cdot), \dots,\mathbf{v}_{\theta}(t-1,\cdot) \mapsto \mathbf{v}_{\theta}(t,\cdot)\), which is abbreviated as \(\mathbf{v}_{\theta}([t{-}\ell{:}t{-}1],\cdot):=\mathbf{v}_{\theta}(t-\ell,\cdot), \dots,\mathbf{v}_{\theta}(t-1,\cdot)\).

We seek to approximate this discretised operator with an emulator
\(\widehat{\mathfrak{F}}_\theta\simeq \mathring{\mathfrak{F}}_\theta\) in the sense that predictions the emulator makes should be close to the ground truth simulation given the same inputs, with respect to some measure of cost.
We fix a parametric class of models \(\{\mathfrak{F}_{\theta,\phi}\}_{\phi}.\)
From this class we \emph{learn} a surrogate 
\(\widehat{\mathfrak{F}}_{\theta,\phi}\) from data.
In learning, we take a dataset \(\mathcal{D}\) comprising discretized PDE solutions conditional on selected parameter values ($\theta_k$),
\(\mathcal{D}:=\{\mathbf{v}^{(k)}_{\theta_k}([0{:}t_{\text{max}}],\cdot) \mid k=1,\dots,K\}\).
Fixing a loss functional \(L\), we aim to find some \(\phi\) achieving a minimal total loss on the training dataset
\begin{equation}\widehat{\phi}=\operatorname{argmin}_{\phi} \sum_{t=1}^{t_\text{max}}\sum_{k=1}^{K} L\left(\mathfrak{F}_{\theta_k,\phi}\{\mathbf{v}^{(k)}_{\theta_k}([t{-}\ell{:}t{-}1],\cdot)\},\mathbf{v}^{(k)}_{\theta_k}(t,\cdot)\right).
\end{equation}
Due to the use of iterative optimization algorithms such as stochastic gradient descent and the non-convex nature of the above optimization problem, we typically obtain local optima.
\(\mathcal{D}\) is generated by a ground-truth solver designed to simulate the desired dynamics with high precision.
In this data we may vary initial conditions, that is, varying \(\mathbf{v}_{\theta}(0,\cdot)\), varying \(\theta\), or both.

In addition to the forward problem, we also consider the use of learned surrogate models to approximately solve \emph{inverse problems} \citep{StuartInverse2010,TarantolaInverse2005}, where an unknown initial condition \(\mathbf{v}_{\theta}(0,\cdot)\) or unknown parameter \(\theta\) is chosen to be congruent with some observed outputs
\(\mathbf{v}_{\theta}([t{:}t+\ell],\cdot)\).
We follow \cite{li2020fourier,MacKinlayModel2021} in using an approximate surrogate approach, taking the forward surrogates as mean predictors for the model.
We assume \(\mathbf{v}_{\theta}(t,\cdot)=\mathfrak{F}_{\theta,\phi}\{\mathbf{v}_{\theta}([t{-}\ell{:}t{-}1],\cdot)\}+\epsilon\) for some mean-zero observation noise \(\epsilon\), and assuming a prior distribution for the unknown of interest.
Other inversion methods can  be used in this domain, such as generative adversarial models, \citep{ChuLearning2021} or variational autoencoders \citep{TaitVariational2020}.

\begin{table*}%[h!]
    \caption{Summary of \projectname{}'s datasets with their respective number of spatial dimensions $N_d$, time dependency, spatial resolution $N_s$, temporal resolution $N_t$, and number of samples generated.}
    \label{tab:data_overview}
    \centering
    \begin{tabularx}{\textwidth}{Xrlrrr}
        \toprule
        PDE & $N_d$ & Time & $N_s$ & $N_t$ & Number of samples \\
        \midrule
        advection & $1$ & yes & $1\,024$ & $200$ & $10\,000$ \\
        Burgers' & $1$ & yes & $1\,024$ & 200 & $10\,000$ \\
        diffusion-reaction & $1$ & yes & $1\,024$ & $200$ & $10\,000$ \\
        diffusion-reaction & $2$ & yes & $128 \times 128$ & $100$ & $1000$ \\
        diffusion-sorption & $1$ & yes & $1\,024$ & $100$ & $10\,000$ \\
        compressible Navier-Stokes & $1$ & yes & $1\,024$ & $100$ & $10\,000$ \\
        compressible Navier-Stokes & $2$ & yes & $512 \times 512$ & $21$ & $1000$ \\
        compressible Navier-Stokes & $3$ & yes & $128 \times 128 \times 128$ & $21$ & $100$ \\
        incompressible Navier-Stokes & $2$ & yes & $256 \times 256$ & $1000$ & $1000$ \\
        Darcy flow & $2$ & no & $128 \times 128$ & -- & $10\,000$ \\
        shallow-water & $2$ & yes & $128 \times 128$ & $100$ & $1000$ \\
        \bottomrule
    \end{tabularx}
\end{table*}

\subsection{Overview of Datasets and PDEs}
\label{sec:data_overview}

The current version of the benchmark provides datasets generated for various PDEs ranging from $1$ to $3$ dimensional spatial domains.
There are both time-dependent and time-independent PDEs. The current datasets are summarized in \autoref{tab:data_overview}; see \autoref{fig:gallery} for a visual teaser. 
Each sample is generated with different parameters, initial conditions, and boundary conditions.
Generalization to different parameters, varying initial conditions, and proper treatment of complex boundary conditions are still open challenges in Scientific ML \cite{karlbauer2022composing, Brandstetter2022MPNPDE, BelbutePeres2021HyperPINN, leiter2021}.
The parameters which are varied to provide several datasets include the advection speed in the advection equation, the forcing term in the Darcy flow, as well as the viscosity in the Burgers' and compressible Navier-Stokes equations all of which can lead to significantly different behaviors of the simulated systems. 
Additionally, besides the periodic boundary condition that is most commonly used in Scientific ML studies, we also provide datasets generated with the Neumann boundary condition in the 2D diffusion-reaction and shallow-water equations, the Cauchy boundary condition in the diffusion-sorption equation, and the Dirichlet condition in the incompressible Navier-Stokes equation. 

We designed this benchmark to represent a diverse set of challenges for emulation algorithms.
In particular, we focus on hydromechanical field equations.
Following this philosophy, we selected 6 basic and 3 advanced real-world problems.
The basic PDEs are stylized, simple models: 1D advection/Burgers/Diffusion-Reaction/Diffusion-Sorption equations, 2D Diffusion-Reaction equation, and 2D DarcyFlow; the advanced and real-world PDEs incorporate features of real-world modeling tasks:
Compressible and incompressible Navier-Stokes equations, and shallow-water equations. 
The PDEs exhibit a variety of behaviors of real-world significance which are known to challenge emulators, such as sharp shock formation dynamics, sensitive dependence on initial conditions, diverse boundary conditions, and spatial heterogeneity.
Finding a surrogate model which can approximate these challenging dynamics with high fidelity we argue is a necessary precondition to applying such models in the real world.
While some of these have been used in prior work, a publicly available benchmark dataset is, to the best of our knowledge, not available. 

In the following we provide a brief introduction and important features of the advanced PDEs.
More detailed explanations for all the PDEs are provided in \autoref{sec:ap:prob_desc}. 

\label{sec:pde_desc}

\paragraph{Compressible Navier-Stokes equations}\label{sec:nast-intro}\dan{We can possibly move this entirely to the appendix, if we think the section above is good enough?}
The compressible fluid dynamics equations describe a fluid flow
whose expression is given as: 
\begin{subequations}\label{eq:nast}
\begin{align} 
    \partial_t \rho + \nabla \cdot (\rho \textbf{v}) &= 0, %\label{eq:nast-1}
    ~~~ \rho (\partial_t \textbf{v} + \textbf{v} \cdot \nabla \textbf{v}) = - \nabla p + \eta \triangle \textbf{v} + (\zeta + \eta/3) \nabla (\nabla \cdot \textbf{v}),
    \label{eq:nast-2}\\
    \partial_t (\epsilon + \rho v^2/2) &+ \nabla \cdot [( p + \epsilon + \rho v^2/2 ) \bf{v} - \bf{v} \cdot \sigma' ] = 0,\label{eq:nast-3}
\end{align}
\end{subequations}
where $\rho$ is the mass density, $\bf{v}$ is the fluid velocity, $p$ is the gas pressure, $\epsilon$ is an internal energy described by the equation of state, $\sigma'$ is the viscous stress tensor, and $\eta$ and $\zeta$ are shear and bulk viscosity, respectively. 
This equation can describe more complex phenomena, such as shock wave formation and propagation. 
It is applied to many real-world problems, such as the aerodynamics around airplane wings and interstellar gas dynamics. 

\paragraph{Incompressible Navier-Stokes equations}

The Navier-Stokes equation is the incompressible version of the compressible fluid dynamics equation, applicable to sub-sonic flows.
This equation can model a variety of systems, from hydromechanical systems to weather forecasting or investigating turbulent dynamics.

\paragraph{Shallow-Water Equations}

The shallow-water equations, derived from the compressible Navier-Stokes equations, present a suitable framework for modeling free-surface flow problems.
In 2D, these come in the form of the following system of hyperbolic PDEs,
\begin{subequations} \label{eq:swe}
\begin{align}
     \partial_t h + \nabla h {\bf u} = 0, \quad
     \partial_t h {\bf u} + \nabla \left( {\bf u}^2 h + \frac{1}{2} g_r h^2 \right) = - g_r h \nabla b \, , 
\end{align}
\end{subequations}
where ${\bf u} = u, v$ being the velocities in the horizontal and vertical direction, $h$ describing the water depth, and $b$ describing a spatially varying bathymetry.
$h {\bf u}$ can be interpreted as the directional momentum components and $g_r$ describes the gravitational acceleration.
The mass and momentum conservation properties even hold across shocks in the solution and thus challenging datasets can be generated.
Example applications include the simulation of tsunamis or general flooding events.

\subsection{Overview of Metrics} 

The standard approach of computing the RMSE on test data falls short of capturing important optimization criteria in Scientific ML. A good fit to (often sparse) data is not sufficient if the physics of the underlying problem is severely violated. Physics-informed learning that aims to conserve physical quantities must therefore be evaluated with appropriate metrics. A global, averaged metric for instance cannot capture small spatial scale changes critical in turbulent regimes. Moreover, a single evaluation metric is not sufficient to compare different methods with respect to their ability to extrapolate to unseen time steps and parameters which are important but underexplored evaluation criteria for ML surrogates.
Hence, the proposed benchmark includes several novel metrics which we believe provide a deeper and more holistic understanding of the surrogate's behavior and which are designed to reflect both the data and physics perspective. The following table summarizes the metrics used; further details can be found in the \autoref{sec:app:metrics}.
\begin{center}
% \footnotesize
    \begin{tabularx}{\textwidth}{lll}
        \toprule
        Scope & Acronym & Metric \\
        \midrule
        \multirow{3}{*}{\parbox{1cm}{Data view}} & RMSE & root-mean-squared-error \\
        & nRMSE & normalized RMSE (ensuring scale independence) \\
        & max error & maximum error (local worst case; also proxy for stability of time-stepping) \\
        \midrule
        \multirow{5}{*}{\parbox{1cm}{Physics view}} & cRMSE & RMSE of conserved value (deviation from conserved physical~quantity) \\
        & bRMSE & RMSE on boundary (whether boundary condition can be learned)\\
        & fRMSE low & RMSE in Fourier space, low frequency regime (wavelength dependence)\\
        & fRMSE mid & RMSE in Fourier space, medium frequency regime \\
        & fRMSE high & RMSE in Fourier space, high frequency regime \\
        \bottomrule
    \end{tabularx}
    
\end{center}

\subsection{Existing Baseline Surrogate Models}
\paragraph{U-Net} U-Net \citep{RonnebergerUNet2015} is an auto-encoding neural network architecture used for processing images using multi-resolution convolutional networks with skip layers. U-Net is a black-box machine learning model that propagates information efficiently at different scales.
Here, we extended the original implementation, which uses $2$D-CNN, to the spatial dimension of the PDEs (i.e. $1$D,$3$D). 
%\vspace{-1em}
\paragraph{Fourier neural operator (FNO)} FNO \citep{li2020fourier} belongs to the family of Neural Operators (NOs), designed to approximate 
the forward propagator of PDEs.
FNO learns a resolution-invariant NO by working in the Fourier space and has shown success in learning challenging PDEs. 
%\vspace{-1em}
\paragraph{Physics-Informed Neural Networks (PINNs)}  Physics-informed neural networks \citep{RaissiPhysicsinformed2019} are methods for solving differential equations using a neural network $u_{\theta}(t,x)$ to approximate the solution by turning it into a multi-objective optimization problem.
The neural network is trained to minimize the PDE residual as well as the error with regard to the boundary and initial conditions.
PINNs naturally integrate observational data~\citep{leiter2022}, but require retraining for each new condition. 
\vspace{-1em}
\paragraph{Gradient-Based Inverse Method} 
Since the surrogate model is fully differentiable,
we use its gradient to solve inverse inference by minimizing the prediction loss \citep{cao2018inverse,nocedal1999numerical}, where a function surface defining the unknown initial conditions \citep{MacKinlayModel2021}, 
is specified through bilinear interpolation.

\subsection{Data Format, Benchmark Access, Maintenance, and Extensibility}
\label{sec:data_access}

\begin{listing}[t!]
\begin{Verbatim}[commandchars=\\\{\}]
\PYG{k+kn}{from} \PYG{n+nn}{pyDaRUS} \PYG{k+kn}{import} \PYG{n}{Dataset}
\PYG{n}{p\PYGZus{}id} \PYG{o}{=} \PYG{l+s+s2}{\PYGZdq{}doi:10.18419/darus\PYGZhy{}2986\PYGZdq{}}
\PYG{n}{dataset} \PYG{o}{=} \PYG{n}{Dataset}\PYG{o}{.}\PYG{n}{from\PYGZus{}dataverse\PYGZus{}doi}\PYG{p}{(}\PYG{n}{p\PYGZus{}id}\PYG{p}{,} \PYG{n}{filedir}\PYG{o}{=}\PYG{l+s+s2}{\PYGZdq{}data/\PYGZdq{}}\PYG{p}{)}
\end{Verbatim}
\caption{Including a benchmark dataset.}
\label{lst:load}
\end{listing}

% [data format] 
The benchmark consists of different data files, one for each equation, type of initial condition, and PDE parameter, using the HDF5 \cite{HDF5} binary data format.
Each such file contains multiple arrays where each array has the dimensions $N,T,X,Y,Z,V$ with $N$ the number of samples, $T$ the number of time steps, and $X,Y,Z$ the spatial dimensions and $V$ the dimension of the field. Additional information on the data format is provided in the Supplementary Material.

\projectname{}'s datasets are stored and maintained using \textsc{DaRUS}, the University of Stuttgart's data repository based on the OpenSource Software DataVerse\footnote{\url{https://dataverse.org}}. \textsc{DaRUS} follows the Findable, Accessible, Interoperable and Reusable (FAIR) data principles \cite{wilkinson2016fair}. All data uploaded to DaRUS gets a DOI as a persistent identifier, a license, and can be described with an extensive set of metadata, organized in metadata blocks. A dedicated team ensures that \textsc{DaRUS} is continuously maintained. 
Through \textsc{DaRUS} we provide a permanent DOI (\href{https://doi.org/10.18419/darus-2986}{doi:10.18419/darus-2986}) \cite{PDEBenchDataset}
for the benchmark data. We also support a straightforward inclusion of the benchmark with a few lines of code.  
In Listing~\ref{lst:load} we demonstrate the way in which the Dataverse~\citep{dataverse} platform\footnote{\url{https://darus.uni-stuttgart.de/dataverse/sciml_benchmark}} supports the integration of pre-generated datasets using a few lines of code. Specifically, we utilize the easy-to-use pyDaRUS Python package to access the data.
It provides a simple API for both downloading and uploading data as well as providing metadata to the Dataverse platform.

In Listing~\ref{lst:dataset} we show an example leveraging pre-defined classes included in our benchmark code to load specific datasets as PyTorch \citep{pytorch} \lstinline{Dataset} classes.
Subsequently, these can be used to construct common \lstinline{DataLoader} instances for training custom ML models. We utilize the Hydra \citep{hydra} library simplifying the configuration of both surrogate model training as well as the generation of additional datasets.
For the latter, we provide and expose various parameters of the underlying simulations for the end user to tweak.
This provides a low barrier of entry for users to try out benchmarking with new experiments or baseline configurations.
%\vspace{-0.3cm}

\begin{listing}[t!]
\begin{Verbatim}[commandchars=\\\{\}]
\PYG{k+kn}{from} \PYG{n+nn}{pdebench.models.fno.utils} \PYG{k+kn}{import} \PYG{n}{FNODatasetSingle}
\PYG{n}{filename} \PYG{o}{=} \PYG{l+s+s2}{\PYGZdq{}data/2D\PYGZus{}diff\PYGZhy{}react\PYGZus{}NA\PYGZus{}NA\PYGZdq{}}
\PYG{n}{train\PYGZus{}data} \PYG{o}{=} \PYG{n}{FNODatasetSingle}\PYG{p}{(}\PYG{n}{filename}\PYG{p}{)}
\PYG{n}{train\PYGZus{}loader} \PYG{o}{=} \PYG{n}{torch}\PYG{o}{.}\PYG{n}{utils}\PYG{o}{.}\PYG{n}{data}\PYG{o}{.}\PYG{n}{DataLoader}\PYG{p}{(}\PYG{n}{train\PYGZus{}data}\PYG{p}{)}
\end{Verbatim}
%\vspace{-2em}
\caption{Using the PyTorch data loader.}
\label{lst:dataset}
\end{listing}

\section{A Selection of Experiments}

In this section, we present a selection of experiments for the \projectname{} datasets. An exhaustive discussion of all results is beyond the scope of this paper. An extensive set of additional results, tables, and plots can be found in the Appendix.

\subsection{Baseline Setups}
\label{sec:baseline_setup}

\begin{figure}[!]
    \centering
    \begin{subfigure}{0.65\textwidth}
         \centering    
        \includegraphics[width=\textwidth]{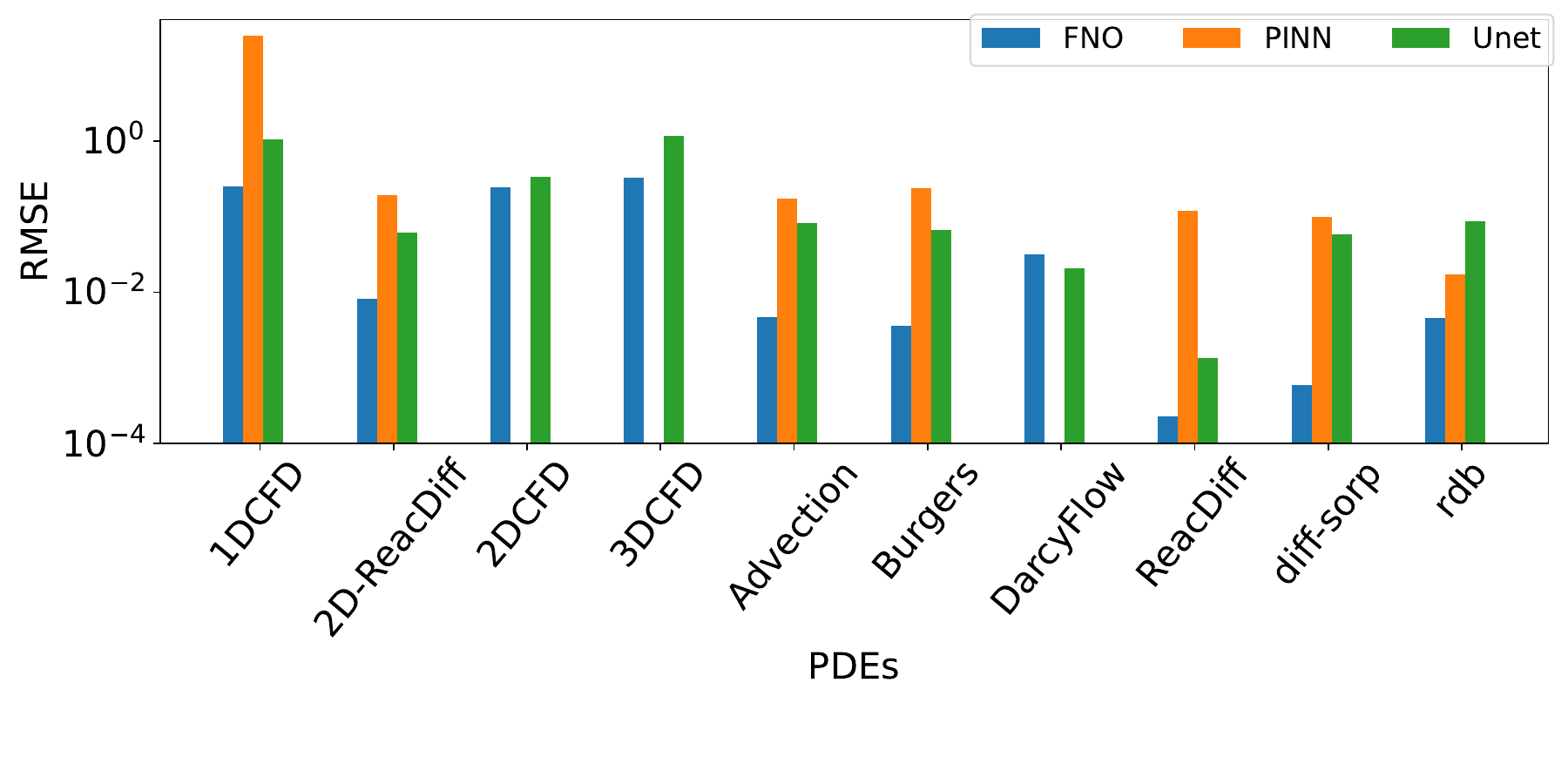}
        \vspace{-2em}
        \caption{}
        \label{fig:baseline_comp_forward}
    \end{subfigure}    
    \begin{subfigure}{0.33\textwidth}
         \centering    
        \includegraphics[width=\textwidth]{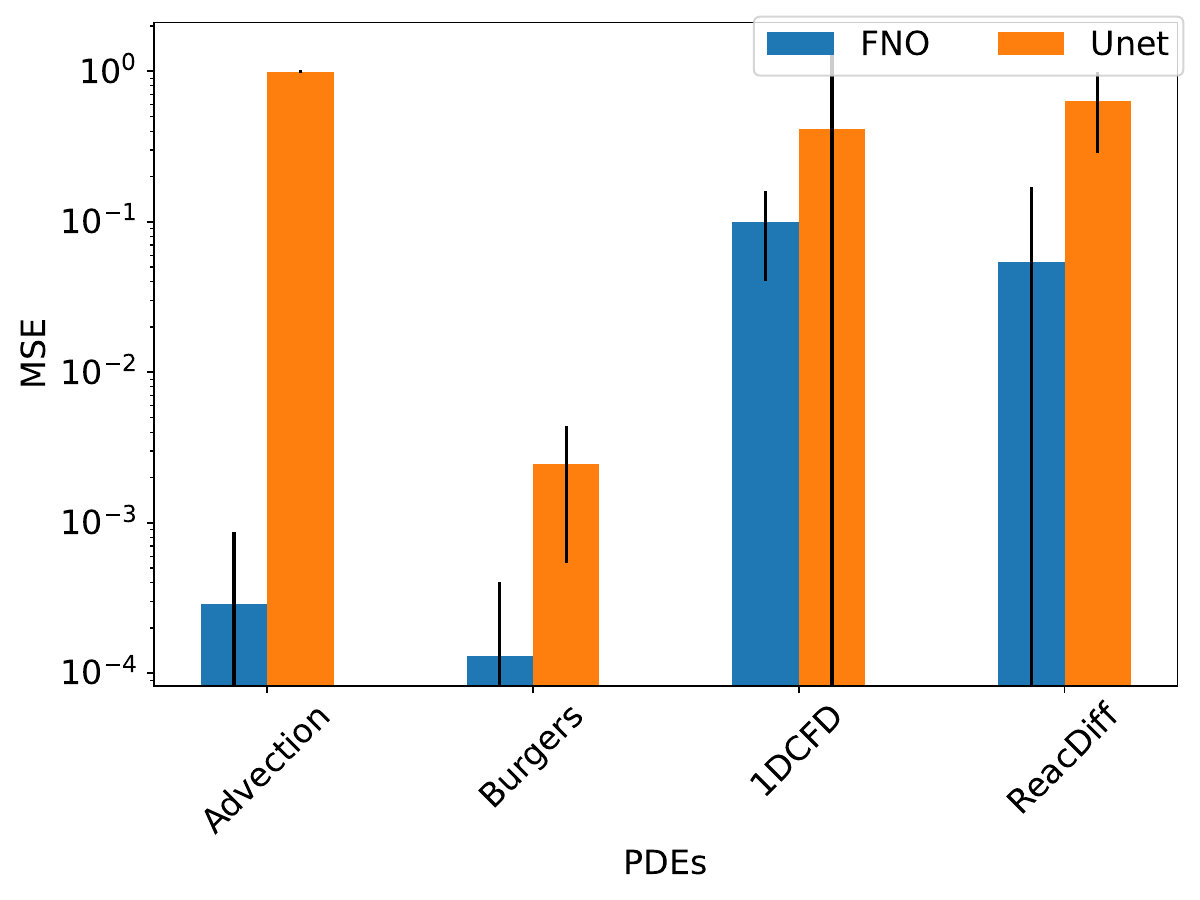}
        \vspace{1em}
        \caption{}
        \label{fig:baseline_comp_inverse}
    \end{subfigure}
     \vspace{-2em} % REMOVING SUBFIGURE LABELS
    \caption{Comparisons of baseline models' performance for different problems for (a) the forward problem and (b) the inverse problem. }%The shown results were averaged over different parameters of each PDE.}
    \label{fig:baseline_comp}
\end{figure}

We trained and tested the baseline emulator models, namely U-Net, FNO, and PINN with the datasets generated with the PDEs described in \autoref{sec:pde_desc}. 
The data was split into training and test data: 90\% was for training and 10\% for test data. 
For FNO, we followed the original implementation, hyperparameters, and training protocols. 
We trained U-Net similarly to FNO, but with the autoregressive methods with the pushforward trick with slight modification to the original implementation \citep{Brandstetter2022MPNPDE}. A more comprehensive comparison between different U-Net training methods is presented in \autoref{sec:U-Net_ar}.
The PINN baseline is implemented using the open-source DeepXDE~\citep{lu2021deepxde} library.
\textcolor{black}{The training was performed on GeForce RTX 2080 GPUs for 1D/2D cases, and GeForce GTX 3090 for 3D cases.} The detailed training protocol and hyper-parameters are provided in \autoref{sec:hyperparameters}. 
Training code and configurations are open and well documented, allowing researchers to easily reproduce or extend these methods. 

\subsection{Baseline Performance}
\label{sec:baseline}

\begin{figure}[!]
    \centering
        \begin{subfigure}{0.3\textwidth}
         \centering
         \includegraphics[width=\textwidth]{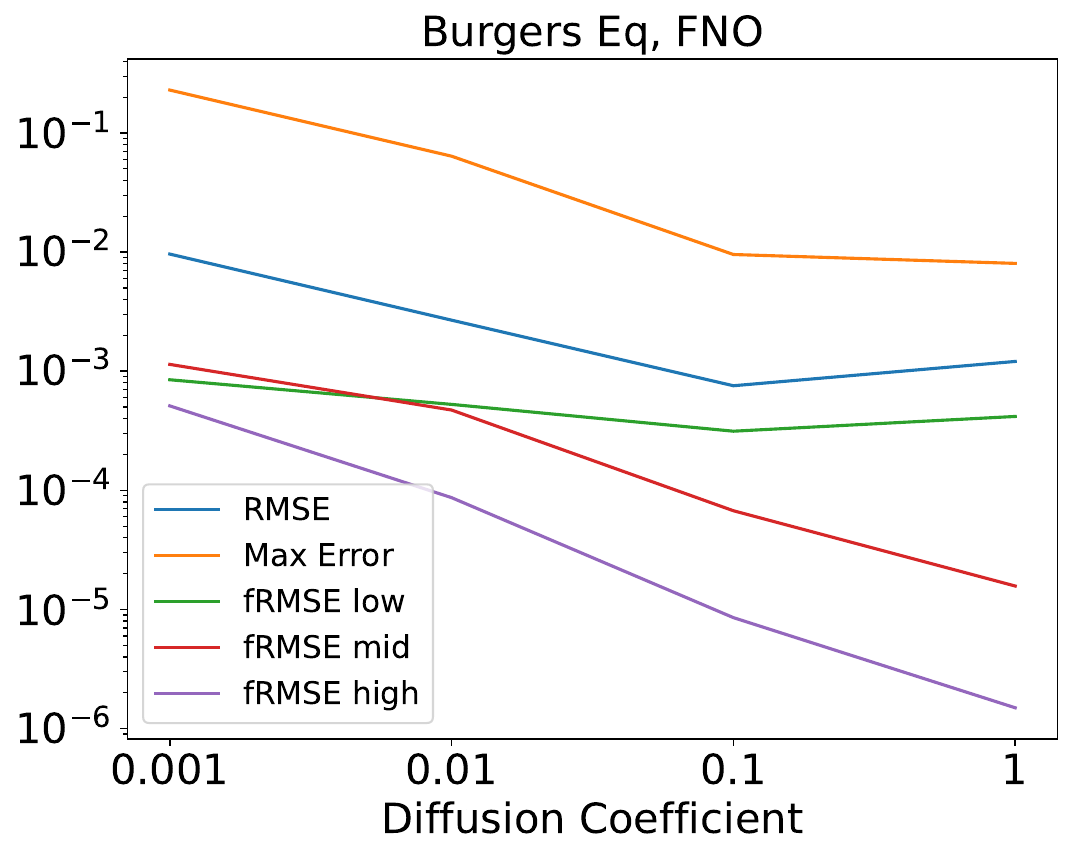}
         \caption{}
         \label{fig:pde_burgers}
     \end{subfigure}
     \hfill
         \begin{subfigure}{0.3\textwidth}
         \centering
         \includegraphics[width=\textwidth]{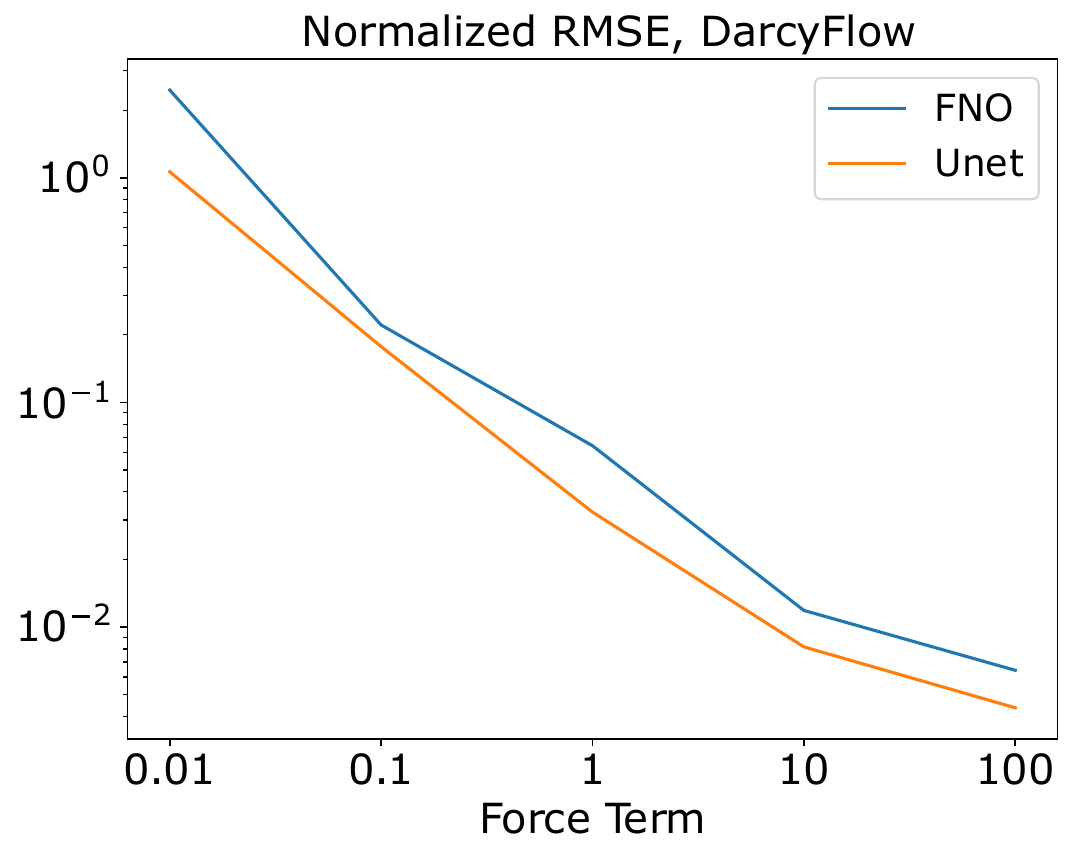}
         \caption{}
         \label{fig:pde_darcy}
     \end{subfigure}
     \hfill
    \begin{subfigure}{0.3\textwidth}
         \centering
         \includegraphics[width=\textwidth]{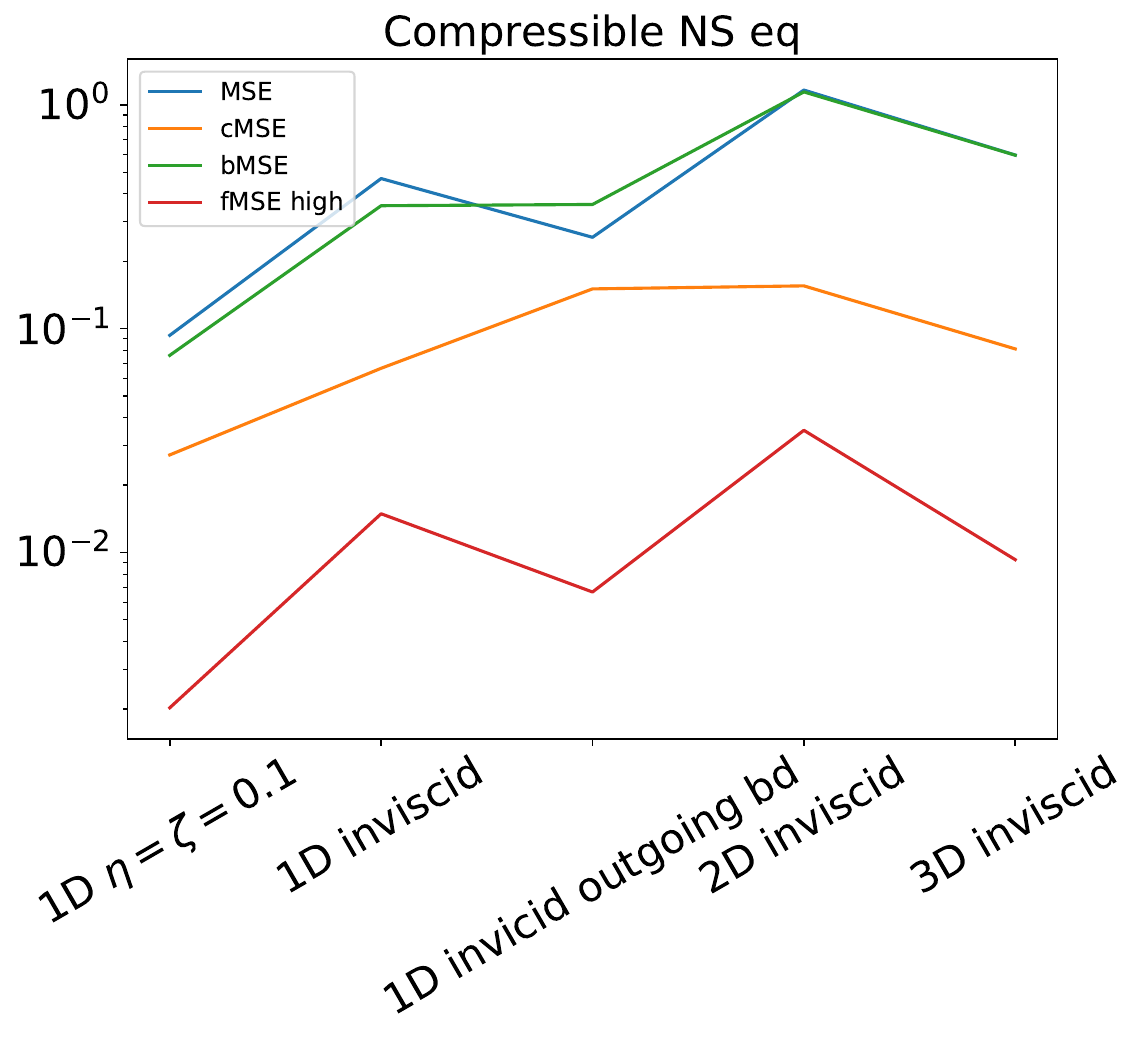}
         \caption{}
         \label{fig:pde_CFD}
     \end{subfigure}
     \vspace{-2em} % REMOVING SUBFIGURE LABELS
     \caption{Detailed visualization of (a) Burgers', (b) DarcyFlow, and (c) Compressible NS eqs.}
    \label{fig:result_analysis}
\end{figure}

\begin{figure}[!]
    \centering
        \begin{subfigure}{0.3\textwidth}
         \centering
         \includegraphics[width=\textwidth]{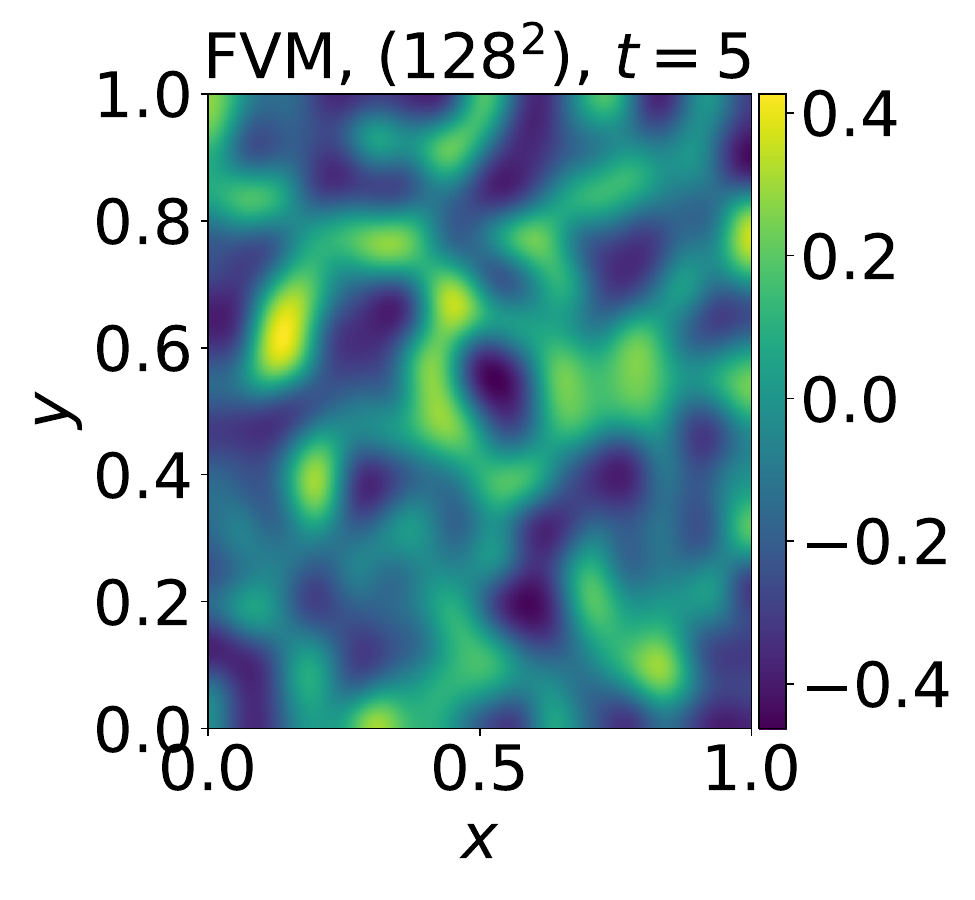}
         \caption{}%Plots of the RMSE calculated at different unrolled time steps.}
    \label{fig:vis_diff-react_data}
    \end{subfigure}
    \hfill
    \begin{subfigure}{0.3\textwidth}
         \centering
         \includegraphics[width=\textwidth]{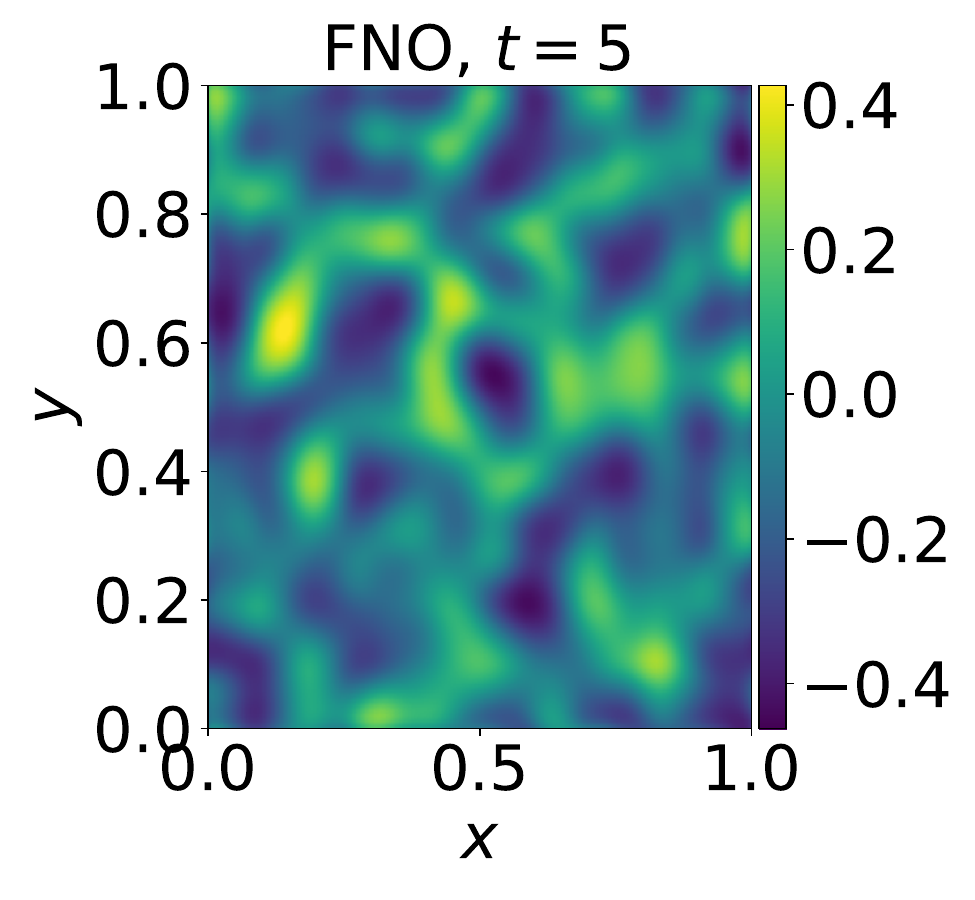}
         \caption{}
         \label{fig:vis_diff-react_fno}
     \end{subfigure}
     \hfill
         \begin{subfigure}{0.3\textwidth}
         \centering
         \includegraphics[width=\textwidth]{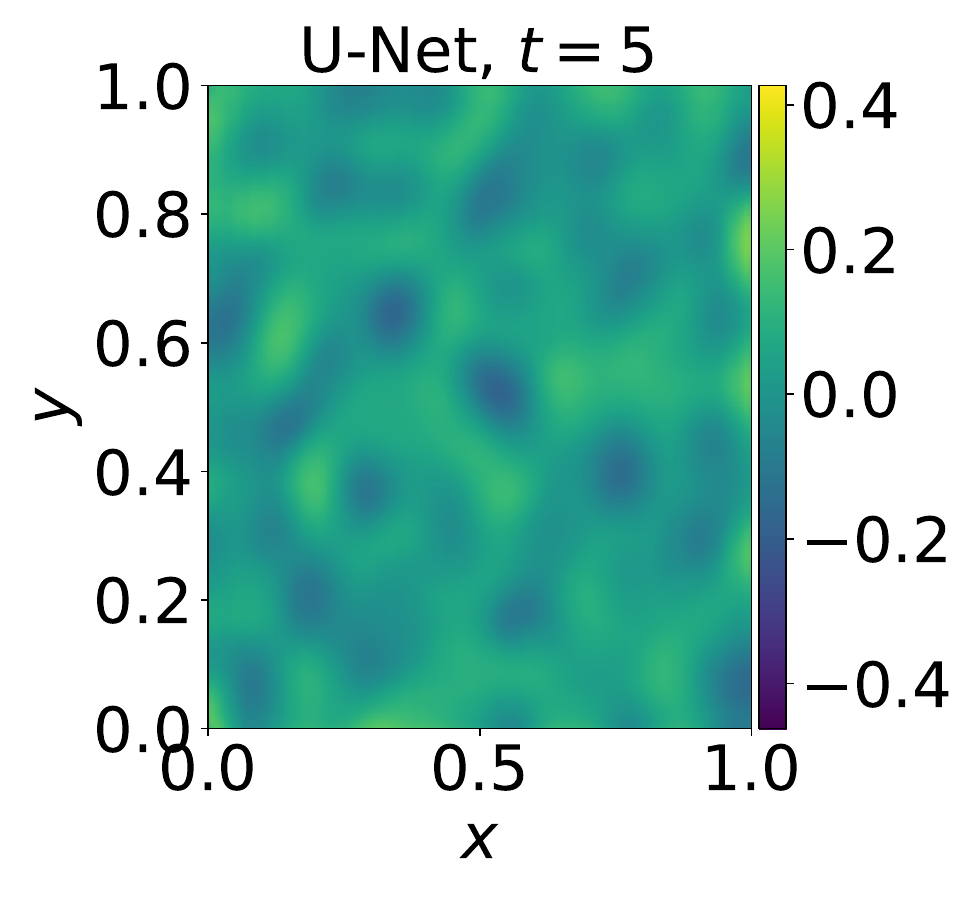}
         \caption{}
         \label{fig:vis_diff-react_unet}
     \end{subfigure}
     \vspace{-2em} % REMOVING SUBFIGURE LABELS
     \caption{(a) Visualization of the 2D diffusion-reaction data generated with a standard finite volume (FVM) solver and a resolution of $128^2$, (b) FNO prediction, and (c) U-Net prediction.}
    \label{fig:vis_diff-react}
\end{figure}

 \autoref{fig:baseline_comp}
 \footnote{%
 In \autoref{fig:baseline_comp} CFD means compressible fluid dynamics or, equivalently, the compressible Navier-Stokes equations. 
 }%
 visualizes the RMSE performance of the surrogate models, averaged for each trained model over different PDE parameters. A more detailed comparison is shown %in \autoref{tab:baseline} 
 in \autoref{sec:app:baseline}.
 \footnote{Note that we omitted the PINN baseline score for the 2D/3D Compressible Navier-Stokes equations and the DarcyFlow. The reason for the former case is limited GPU memory, and the reason for the latter is that the DarcyFlow's problem setup is to learn the mapping from diffusion coefficients $a(x)$ to a steady state solution, which cannot be treated by PINNs. 
 }%
Among the baseline surrogate models, FNO provides the best prediction for most metrics.
It learns the differential operators well, leading to low errors even for the conserved quantities and on the boundaries.
Additionally, FNO has a consistent error of about $4\times 10^{-4}$ across the frequency spectrum for many problems highlighting its ability to learn in Fourier space.
As an example,  \autoref{fig:vis_diff-react} depicts the FNO and U-Net predictions for the 2D diffusion-reaction equation and the training data obtained from a numerical simulator.
Our baseline results further indicate that the PINNs might deal better than expected with high-frequency features, despite prior observations~\citep{wang2022}.
As an example for an inverse problem setup, we identify the initial condition to minimize the prediction error of the ML surrogate over a  $15$ time steps horizon.
In \autoref{fig:baseline_comp_inverse}, we present the MSE of the prediction of estimated initial condition (with error bars) for $4$ of the $11$ datasets (1D).
The results show that FNO outperforms U-Net also for the inverse problem. 
However, our benchmark also reveals several tasks which these methods cannot treat properly. 
First, \autoref{fig:pde_burgers} shows that the FNO's error increases with decreasing diffusion coefficient where a strong discontinuity appears.
This can be attributed to Gibb's phenomenon for FNO's limited maximum wave frequency in Fourier space, as shown by an increase of two orders of magnitude for high-frequency fRMSE.
Second, \autoref{fig:pde_darcy} shows that the normalized RMSE increases with decreasing force term, which is equivalent to decreasing the scale-value of the solution (in our case, force term 0.01 means $\mathrm{mean}(|u|) \approx 0.01$).\footnote{%
Note that U-Net is consistently better than FNO in this case. We postulate that this is due to the strong similarity between this task and the diffusion-like regression task that the original U-Net targets.
}%
Third, \autoref{fig:pde_CFD} shows several metrics for the compressible Navier-Stokes equations.
It shows the overall RMSE is very bad in comparison to the basic PDEs, such as the Burgers equation. 
Interestingly, the 3D inviscid case shows lower error than 2D inviscid case.
We posit this is due to lower resolution resulting in smooth train/validation samples which FNO can learn very efficiently.
This also indicates that high-resolution training samples should be used to create a surrogate  model for real-world problems with a Reynolds number of more than $10^6$.

\begin{figure}[!]
    \centering
        \begin{subfigure}{0.3\textwidth}
         \centering
         \includegraphics[width=\textwidth]{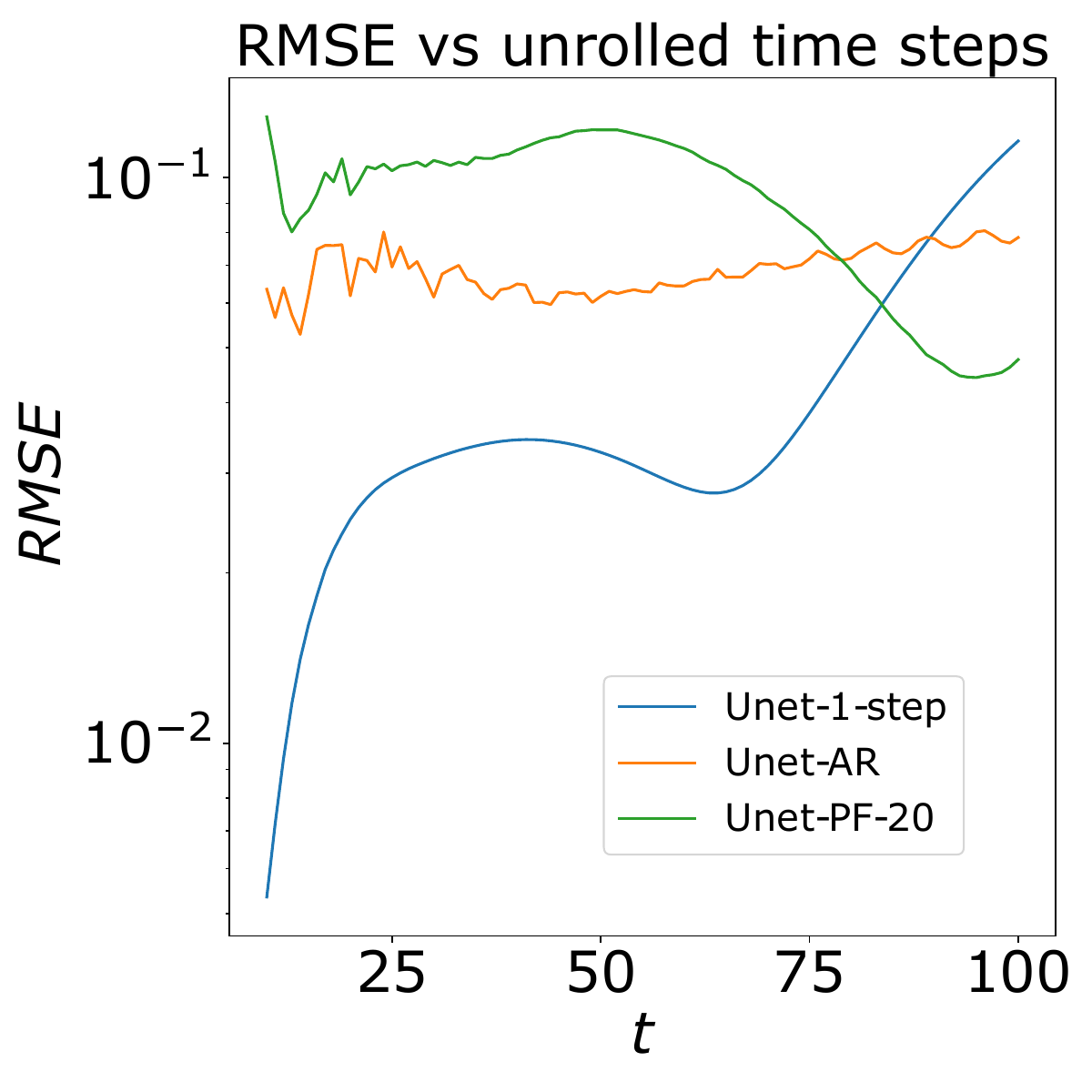}
         \caption{}%Plots of the RMSE calculated at different unrolled time steps.}
    \label{fig:rdb_U-Net_mse_time}
    \end{subfigure}
    \hfill
    \begin{subfigure}{0.3\textwidth}
         \centering
         \includegraphics[width=\textwidth]{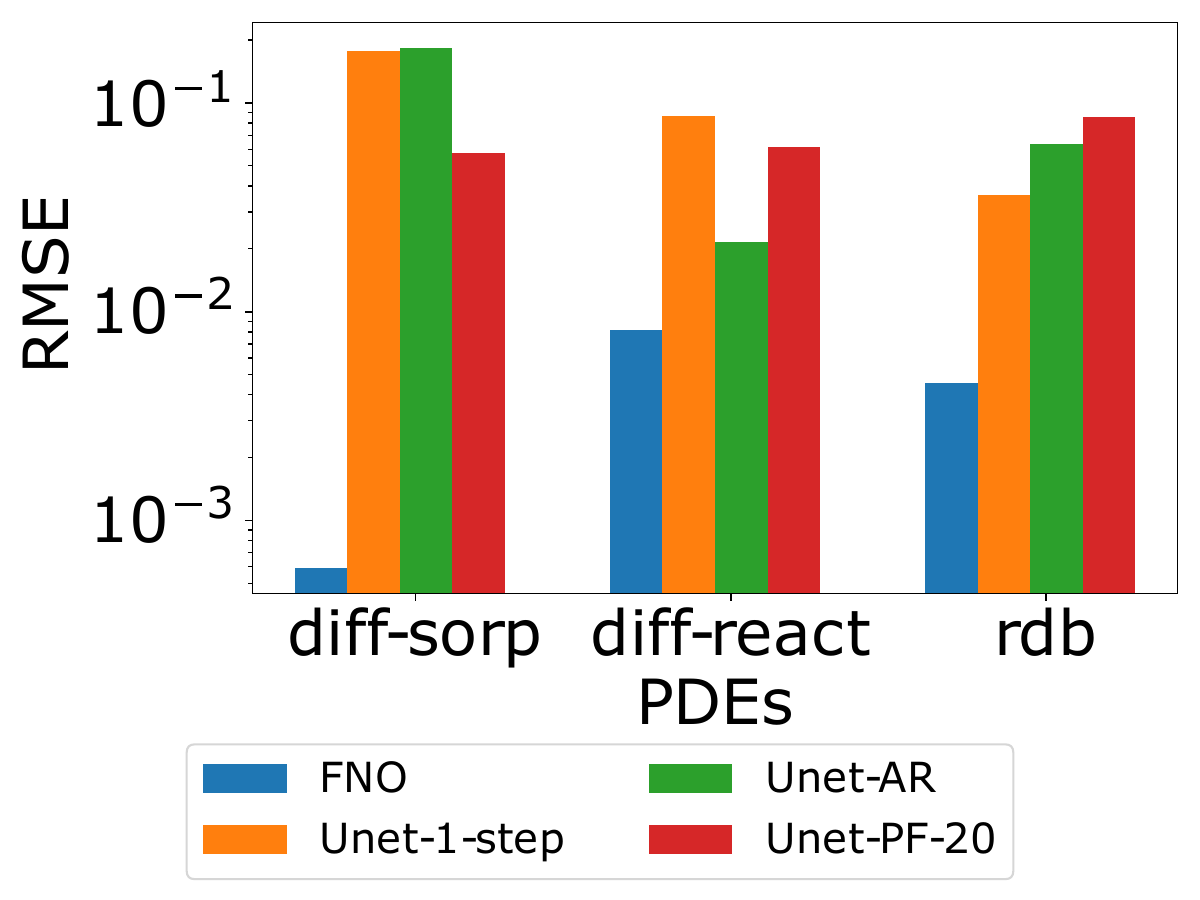}
         \caption{}
         \label{fig:AR_comparison}
     \end{subfigure}
     \hfill
         \begin{subfigure}{0.3\textwidth}
         \centering
         \includegraphics[width=\textwidth]{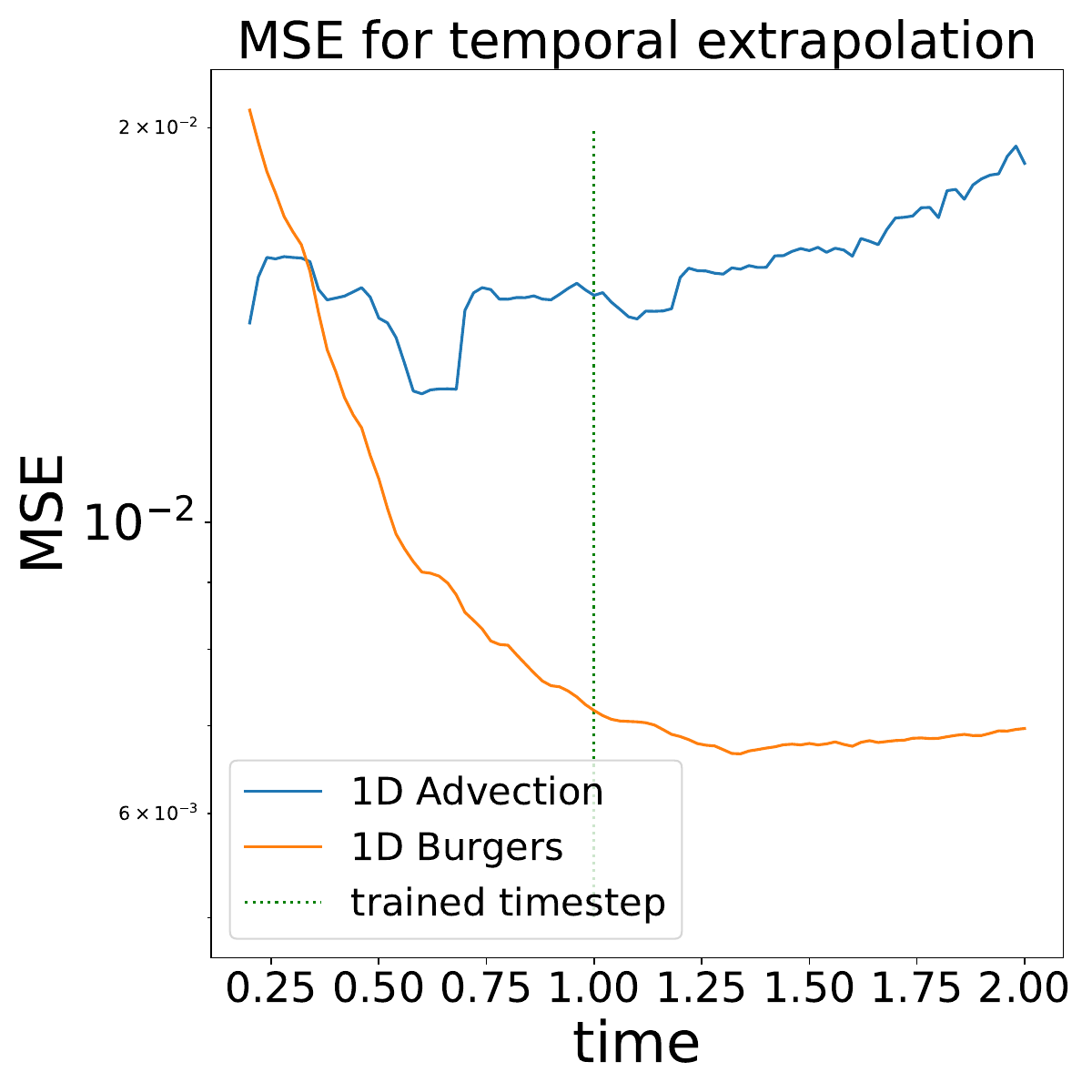}
         \caption{}
         \label{fig:tmp_comparison}
     \end{subfigure}
     \vspace{-2em} % REMOVING SUBFIGURE LABELS
     \caption{(a) Plots of the RMSE calculated at different unrolled time steps, (b) comparison of each autoregressive method, and (c) RMSE for temporal extrapolation.}
    \label{fig:result_analysis_auto}
\end{figure}

\subsection{Temporal Error Analysis}
% [Makoto]
\label{sec:U-Net_ar}

\paragraph{Autoregressive Behaviour of U-Net}
We observed instabilities when training U-Net with fully autoregressive mode.
When trained in an open loop, i.e. only 1 time step ahead without feedback of prediction, the error during testing quickly accumulates with more unrolled time steps.
Therefore, we tried three different training methods as described in the previous section. 
\autoref{fig:rdb_U-Net_mse_time} shows the RMSE behaviour calculated at different unrolled time steps. 
It shows that for different U-Net training strategies, the RMSE behaviors are different.
We observed that U-Net with the pushforward trick provides better stability during training for all problems, and better accuracy for a longer prediction horizon. For this reason, we used the U-Net with pushforward trick as our baseline score for U-Net. 

\paragraph{Temporal Extrapolation}
\autoref{fig:tmp_comparison} plots the RMSE temporal evolution of 1D Advection and Burgers equations predicted using the FNO model. 
Different from the other scenarios, we trained only until half of the time steps used in the other cases (the green dotted line in the figure). The main purpose of this experiment is to test how well the ML model (FNO) learns the temporal dependency of the PDEs with limited information. 
We observe monotonically increasing errors after the time steps used in training ($t>1$).
This indicates that the present ML models cannot properly capture the dynamic behavior of the PDEs, and it is a challenging task to provide reliable predictions beyond the time experienced during training. 

\subsection{Inference Time Comparison}

In this subsection, we provide runtime comparisons between the numerical PDE solvers that we use to generate a single data sample and the FNO, the most accurate ML model according to our experiments. 
A fair runtime comparison between classical solvers and ML methods is challenging because these method classes are usually optimized for different hardware setups.
In this paper, we provide the information on the hardware and system configuration in \autoref{sec:ap:runtime}
\footnote{
  We used the same hardware resources and system configuration for the 2D/3D CFD simulations and the ML methods. Whether or not this consitutes a fair comparison is surely debatable. 
}
.
In addition, we list the ML models' total training time. All the runtime numbers are given in seconds. The highest computational demand originates from the ML training time. However, once the ML models were trained,  the ML models' predictions can be computed multiple orders of magnitudes more efficiently. A more complete overview of these experiments can be found in \autoref{sec:ap:runtime}. 
The ML model allows us to predict a solution even in the strong-viscous regime ($\eta=\zeta=0.1$) efficiently since it eliminates the stability restriction defined by the Courant-Friedrichs-Lewy condition \citep{lewy1928partiellen} 
\footnote{
Note that in the table the ML inference and training time does not monotonically increase with spatial-dimension. This is because the 2D/3D cases' time-step numbers and training sample numbers is much smaller than Diffusion-sorption case.
}. 
A more detailed analysis of the resolution sensitivity of inference time is provided in \autoref{sec:time_sensitivity}. 

\begin{center}
    \begin{tabularx}{\textwidth}{lrrrr}
        \toprule
        PDE & Resolution  & Simulation time  &  ML inference time & ML training time  \\
        \midrule
        Diffusion-sorption & $1\,024^1$ & $59.83$ & $0.32$ & $48\,760$ \\
        2D CFD ($\eta=\zeta=0.1$)& $512^2$  & $582.61$  & $0.14$ & $107\,567$  \\
        3D CFD (inviscid)& $64^3$ & $60.06$ & $0.14$ & $12\,387$  \\
        \bottomrule
        \label{tab:inf_time}

    \end{tabularx}
\end{center}    

\section{Conclusions and Limitations}
\label{sec:conclusions}
With \projectname{} we contribute an extensive benchmark suite for comparing and evaluating methods on the realm of Scientific ML.
We provide both pre-computed datasets for easy access in a dataverse as well as code to generate new data from configurable simulation runs.
The focus is on time-dependent PDE problems ranging from simple 1D equations to challenging 3D coupled systems of equations featuring challenging boundary conditions.
Furthermore, we present a variety of different evaluation metrics in order to better understand strengths and weaknesses of the machine learning methods in a scientific computing context.
We also provide an example application for utilizing Scientific ML methods for inverse modeling with our benchmark data.
We believe this will be an important area in the future for machine learning models to produce competitive results both in accuracy as well as runtime when compared to numerical methods.

\paragraph{Limitations} While \projectname{} provides an easy-to-use, modular and extensible approach to devising and testing ML surrogates for PDE simulations, the scope of our benchmark is naturally limited. Our main focus is on time-dependent PDEs for different types of flow problems which pose a wide variety of challenges to Scientific ML. We currently do not cover other types of physics nor quantum mechanics as this goes beyond the scope of this paper. With respect to flow problems, our main limitations are that we do not yet cover multi-phase flows or non-rectangular domains. This is left for future work. 

\clearpage
% %%%%%%%%%%%%%%%%%%%%%%%%%%%%%%%%%%%%%%%%%%%%%%%%%%%%%%%%%%%%
\section*{Acknowledgements}
We thank MinMae (John) Kim, Ran Zhang, Tianqi Wang, Yizhou Yang, Gefei Shan and Simon Brown of the \hyperlink{https://comp.anu.edu.au/TechLauncher/}{ANU Techlauncher} for consulting on code design and implementation. Partially funded by Deutsche Forschungsgemeinschaft (DFG, German Research Foundation) under Germany's Excellence Strategy - EXC 2075 – 390740016. We acknowledge support by the Stuttgart Center for Simulation Science (SimTech). We further thank the DaRUS-team, in particular Jan Range.

% \bibliographystyle{abbrvnat}
% \bibliography{physics_benchmark}

\section*{Checklist}

%%% BEGIN INSTRUCTIONS %%%
%The checklist follows the references.  Please
%read the checklist guidelines carefully for information on how to answer these
%questions.  For each question, change the default \answerTODO{} to \answerYes{},
%\answerNo{}, or \answerNA{}.  You are strongly encouraged to include a {\bf
%justification to your answer}, either by referencing the appropriate section of
%your paper or providing a brief inline description.  For example:
%\begin{itemize}
%  \item Did you include the license to the code and datasets? \answerYes{See Section~\ref{gen_inst}.}
%  \item Did you include the license to the code and datasets? \answerNo{The code and the data are proprietary.}
%  \item Did you include the license to the code and datasets? \answerNA{}
%\end{itemize}
%Please do not modify the questions and only use the provided macros for your
%answers.  Note that the Checklist section does not count towards the page
%limit.  In your paper, please delete this instructions block and only keep the
%Checklist section heading above along with the questions/answers below.
%%% END INSTRUCTIONS %%%

\begin{enumerate}

\item For all authors...
\begin{enumerate}
  \item Do the main claims made in the abstract and introduction accurately reflect the paper's contributions and scope?
    \answerYes{All the contributions listed in the abstract are elaborated in \autoref{sec:pdebench}.}
  \item Did you describe the limitations of your work?
    \answerYes{See the last paragraph of \autoref{sec:conclusions}.}
  \item Did you discuss any potential negative societal impacts of your work?
    \answerNA{We believe that our work does not have any potential negative societal impacts as it does not contain confidential or private data.}
  \item Have you read the ethics review guidelines and ensured that your paper conforms to them?
    \answerYes{We do not include any confidential or private data, only numerical values related to general physical systems with no potential ethical issues or severe environmental damage.}
\end{enumerate}

\item If you are including theoretical results...
\begin{enumerate}
  \item Did you state the full set of assumptions of all theoretical results?
    \answerNA{We do not include any theoretical results.}
	\item Did you include complete proofs of all theoretical results?
    \answerNA{See the previous point.}
\end{enumerate}

\item If you ran experiments (e.g. for benchmarks)...
\begin{enumerate}
  \item Did you include the code, data, and instructions needed to reproduce the main experimental results (either in the supplemental material or as a URL)?
    \answerYes{We include the parameters used to reproduce the datasets in \autoref{sec:ap:prob_desc} and instructions to run the code in the README file in our code repository.}
  \item Did you specify all the training details (e.g., data splits, hyperparameters, how they were chosen)?
    \answerYes{See \autoref{sec:hyperparameters}.}
	\item Did you report error bars (e.g., with respect to the random seed after running experiments multiple times)?
    \answerYes{We ran several experiments using different parameters. The mean and standard deviation values for the errors are reported in the \autoref{fig:baseline_comp}.}
	\item Did you include the total amount of compute and the type of resources used (e.g., type of GPUs, internal cluster, or cloud provider)?
    \answerYes{The types of GPUs used are provided in \autoref{sec:baseline_setup}.}
\end{enumerate}

\item If you are using existing assets (e.g., code, data, models) or curating/releasing new assets...
\begin{enumerate}
  \item If your work uses existing assets, did you cite the creators?
    \answerYes{We adopted the implementation of the baseline models with some modifications, with proper citation and credits to the authors, as well as existing software packages.}
  \item Did you mention the license of the assets?
    \answerYes{Appropriate license notices are included in the affected source code files, and license of new assets is included in the supplementary materials.}
  \item Did you include any new assets either in the supplemental material or as a URL?
    \answerYes{All the data generation scripts are included in the code repository.}
  \item Did you discuss whether and how consent was obtained from people whose data you're using/curating?
    \answerNA{We generate all of our data, and the models we use are all openly accessible to the public, so we adopt and modify them, and include citations to the original authors.}
  \item Did you discuss whether the data you are using/curating contains personally identifiable information or offensive content?
    \answerNA{We do not include any personal information or offensive content in our datasets.}
\end{enumerate}

\item If you used crowdsourcing or conducted research with human subjects...
\begin{enumerate}
  \item Did you include the full text of instructions given to participants and screenshots, if applicable?
    \answerNA{We use neither crowdsourcing nor conduct research with human subjects.}
  \item Did you describe any potential participant risks, with links to Institutional Review Board (IRB) approvals, if applicable?
    \answerNA{See the previous point.}
  \item Did you include the estimated hourly wage paid to participants and the total amount spent on participant compensation?
    \answerNA{See the previous point.}
\end{enumerate}

\end{enumerate}

% \end{comment}

\bibliographystyle{abbrvnat}
\bibliography{physics_benchmark}

% %%%%%%%%%%%%%%%%%%%%%%%%%%%%%%%%%%%%%%%%%%%%%%%%%%%%%%%%%%%%
\clearpage
\clearpage
\clearpage
\clearpage
% \cleardoublepage

\setcounter{page}{1}
\resetlinenumber

\appendix
\section*{\projectname{}: An Extensive Benchmark for Scientific Machine Learning. \\ Supplementary Material \footnote{\projectname{} repository \url{https://github.com/pdebench/PDEBench}.}}
\section{Continuation of Related Work}
\label{sec:ap:related}

Benchmarks in machine learning are an ubiquitous feature of the field. In recent years, their design and implementation has become a research area of its own right.
Easily accessible and widely used image classification benchmarks such as MNIST and ImageNet are widely credited with accelerating progress in machine learning.
Various domains in machine learning have widely influential datasets:
In time series forecasting there are the Makridakis competitions \citep{MakridakisM42020}, in reinforcement learning there is the OpenAI Gym \citep{brockman2016openaigym}.
Generic classification problems use, for example, the Penn Machine Learning Benchmark \citep{OlsonPMLB2017}.

Closely related to the chosen Scientific ML baselines is the problem of directly differentiating through the numerical solver, which can itself be used in training an approximating model, or to directly solve some optimization or control problem of interest.
Differentiable direct PDE solvers are increasingly available, e.g. \cite{MituschDolfinadjoint2019} and frequently built upon neural network technology stacks \citep{freeman2021brax,bezgin2020jaxfluids,HollPhiflow2020}.

Recent efforts have attempted to unify Scientific ML surrogates for PDEs under a single interface.
For example, NVIDIA's MODULUS/SimNet \citep{HennighNVIDIA2020}  implements a variety of methods in a single framework, although unfortunately under onerous intellectual property restrictions and an opaque contribution process.
The DeepXDE project \citep{lu2021deepxde} is available under an open license and provides an impressive range of capabilities, but is largely restricted to PINN and DeepONet methods \citep{RaissiPhysicsinformed2019}.

\section{Detailed metrics description}
\label{sec:app:metrics}

The classic loss metrics we use are (1) root-mean-squared-error (RMSE), (2) normalized RMSE (nRMSE), (3) maximum error.
These measure the emulating model's global performance but neglect local performance.
Thus we include extra metrics to measure specific failure modes: (4) RMSE of the conserved value (cRMSE), (5) RMSE at boundaries (bRMSE), (6) RMSE in Fourier space (fRMSE) constrained to low, middle, and high-frequency regions.\dan{I modified this a little; is it correct?}

The normalized RMSE is a variant of the RMSE to provide scale-independent information defined as:
\begin{equation}
  \mathrm{nRMSE} \equiv \frac{||u_{\rm pred} - u_{\rm true}||_2}{||u_{\rm true}||_2},
\end{equation}
where $||u||_2$ is the $L_2$-norm of a (vector-valued) variable $u$, and $u_{\rm true}, u_{\rm pred}$ are true and predicted value, respectively. 
The maximum error measures the model's worst prediction, which quantifies both local performance and models' stability of their prediction.
cRMSE is defined as $\mathrm{nRMSE} \equiv ||\sum u_{\rm pred} - \sum u_{\rm true}||_2/N$, which measure the deviation of the prediction from some physically conserved value.\dan{perhaps this needs an example of a conserved value to make it clear? Also, what does the summation run over here?}
bRMSE measures the error at the boundary, indicating if the model understand the boundary condition properly. 
Finally, fRMSE measures the error in low/middle/high-frequency ranges defined as 
\begin{equation}
    \frac{\sqrt{ \sum^{k_{\text{max}}}_{k_{\text{min}}} |\mathcal{F}(u_{\rm pred}) - \mathcal{F}(u_{\rm true}))|^2}}{k_{\rm max} - k_{\rm min} + 1},
\end{equation}
where $\mathcal{F}$ is a discrete Fourier transformation, and $k_{\text{min}}$, $k_{\text{max}}$ are the minimum and maximum indices in Fourier coordinates.
In our paper, we define the low/middle/high-frequency regions as
Low: $k_{\text{min}} = 0, k_{\text{max}}= 4$,
Middle: $k_{\text{min}} =5, k_{\text{max}}  =12$, and
High: $k_{\text{min}}=13, k_{\text{max}}=\infty $.
This allows a quantitative discussion of the model performance's dependence on the wavelength. In the multi-dimensional cases, we first integrate the angular coordinate direction of $|\mathcal{F} [u_{\rm pred} - u_{\rm true}](k)|^2$, and take the sum along the $k$-coordinate. 

\subsection{Inverse Problem Metrics}
For the inverse problem setup, we selected various metrics. The major difference with respect to the forward metrics is that we have two main quantities to measure:
\begin{itemize}
    \item the error of the {\it quantity} we want to estimate, in our case the initial condition $u_0$: 
    $$
    \mathcal{L}(u_0,\hat{u}_0)
    $$
    where $\hat{u}_0$ is the estimated value;
    \item the error of the {\it prediction} based on the estimated initial condition $u(t,x|u_0)$,
    $$
    \mathcal{L}\left(u(t,x|u_0),u(t,x|\hat{u}_0)\right)
    $$    
\end{itemize}
In general, we expect a larger error when we measure the error in the estimated quantity w.r.t. the predicted quantity. This is mainly due to the early decay of high frequencies of the PDE. 
We evaluated the error of the prediction at a specific instant in time $t=T$, that has been selected as $T=15$ for all the tested datasets, expect $T=5$ for the CFD dataset. 

The metrics that we used for the inverse problem are: 1) MSE 2) the normalized $\ell_2$ norm (L2), 3) the normalized $\ell_3$ norm (L3); 4) the FFT MSE, the FFT L2 and 5) the FFT L3. For the frequency metrics we investigated the low frequency (between $0$ and $1/4$ of the max frequency), the middle frequency (between $1/4$ and $3/4$) and high frequency (between $3/4$ and the maximum frequency) ranges. In Fig.\ref{fig:ic11d}, the right figure shows the frequency power density, where we see that the largest error is found in the middle frequency range. 

\section{Training Protocol and Hyperparameters}
\label{sec:hyperparameters}

The model was trained for 500 epochs with the Adam optimizer \citep{kingma2014adam} as per the protocol of the original FNO.
The initial learning rate was set as $10^{-3}$ and reduced by half after each 100 epochs.
The datasets are split into 90\%  training and 10\%  validation and testing.
For the PINNs, we use  DeepXDE~\citep{lu2021deepxde} implementation.
The training was performed for 15,000 epochs with the Adam optimizer, with the learning rate set to $10^{-3}$.
As with the example problems from that library we use a fully-connected network of depth $6$ with $40$ neurons each.
In contrast to the other surrogate models, the PINN baseline can be trained and tested only on a single sample, and is valid only for a specific initial and boundary condition.
To get more reliable error bounds, we thus chose to train the PINN baseline for $10$ different samples per dataset and average the resulting error metrics. 

\subsection{Inverse problem} 
For testing the power of surrogate models to solve inverse problems, we consider a simplified scenario where the machine learning model directly predicts a specific time in the future $t=T$.
When training to predict a specific time in the future, we reduce the training time and avoid to consider the effect of training approaches (as discussed in the temporal analysis section \autoref{sec:U-Net_ar}) in evaluating the surrogate models.
We trained over $N_\text{epoch}=20$ epochs and we selected as final time step $T=15$ for all tested datasets, expect for the CFD dataset where we selected $T=5$.
We used similar parameters used in the forward training, while we selected  $64$ hidden values to be estimated for the initial condition and $100$ samples to test and $0.2$ as learning rate for the gradient method. The loss function for the gradient computation is the MSE. 

\section{Detailed Problem Description}
\label{sec:ap:prob_desc}

In this section, we provide more detailed descriptions of each PDE and its applications. Note that PDE is the basic  mathematical tool to describe the evolution of the system in physics. Interested readers are referred to representative textbooks of physics, for example, \citep{1963flp..book.....F}.

\subsection{1D Advection Equation}

\begin{figure}[h!]
  \centering
  \begin{subfigure}{0.48\textwidth}
    \centering
    \includegraphics[width=\textwidth]{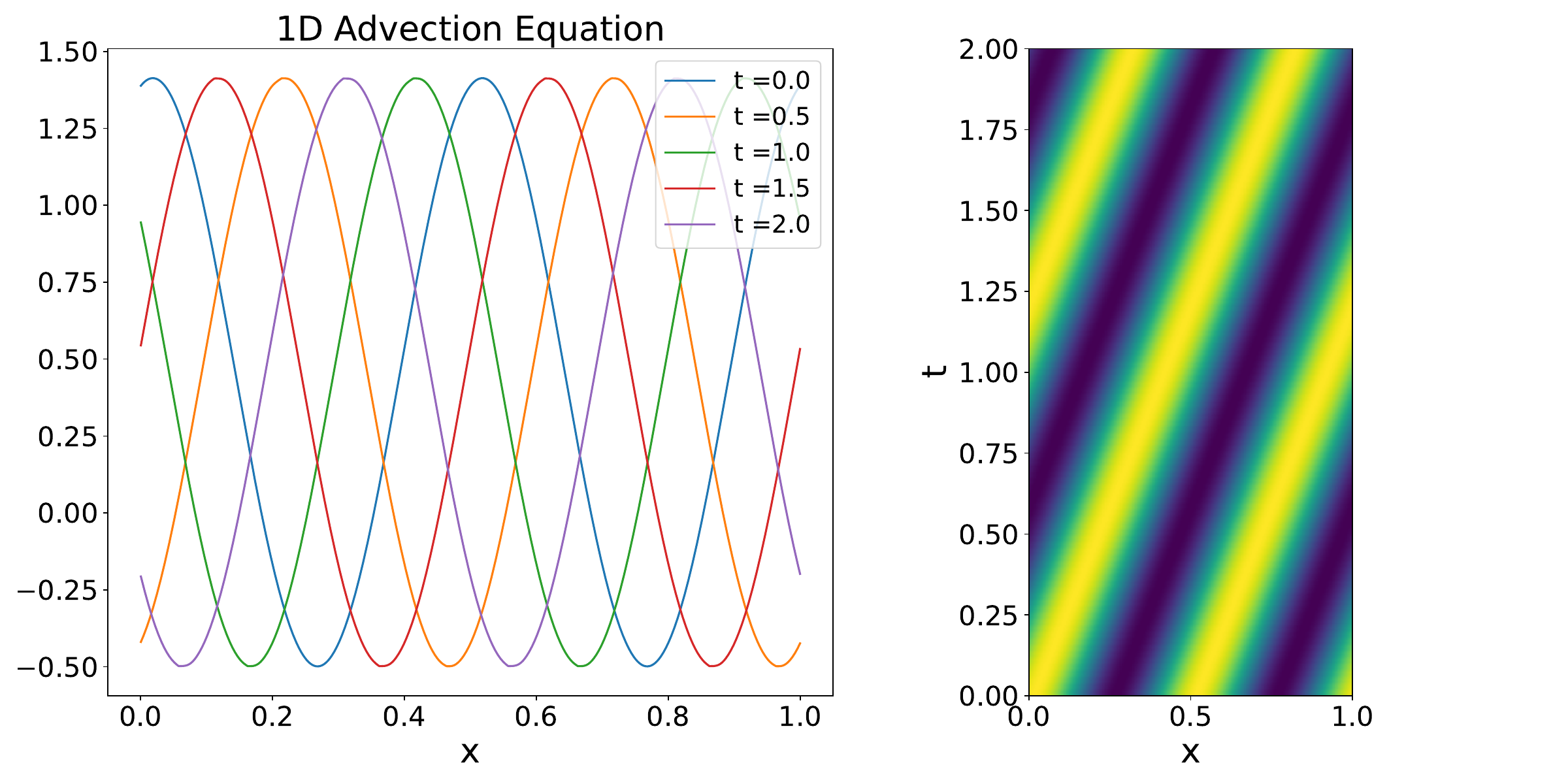}
    \caption{1D Advection ($\beta=0.4$)}
    \label{fig:adv_tmp}
  \end{subfigure}
  \hfill
  \begin{subfigure}{0.48\textwidth}
      \centering
      \includegraphics[width=\textwidth]{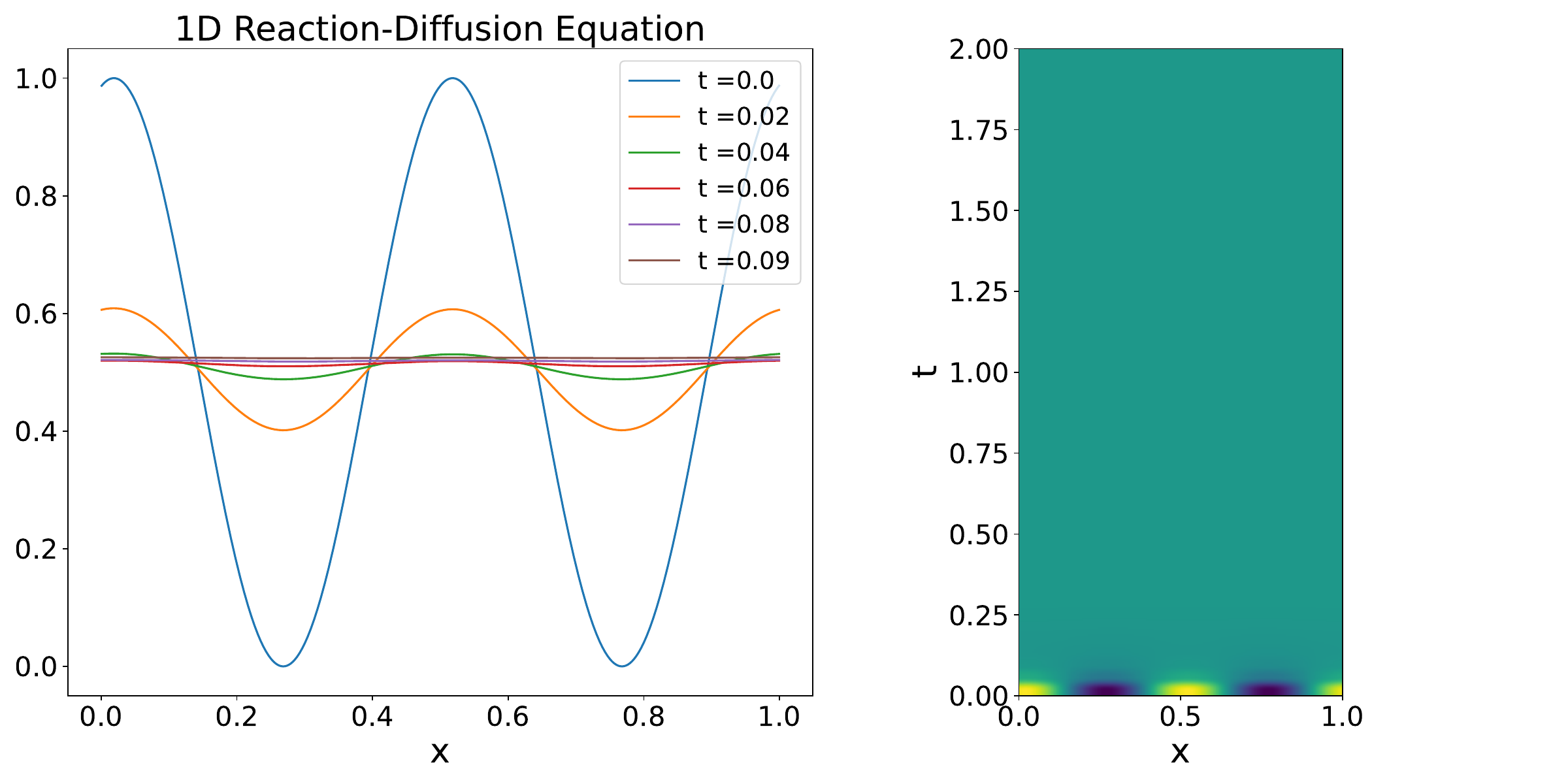}
      \caption{1D Reaction-Diffusion ($\nu=0.5, \rho=1$)}
      \label{fig:reac_tmp}
  \end{subfigure}
  \caption{Visualization of the time evolution of 1D Advection equation and Reaction-Diffusion equation.
  }
 \label{fig:Adv_Reac_tmp}
 \end{figure}

The advection equation models pure advection behavior without non-linearity whose expression is given as:
\begin{align}
     \partial_t u(t,x) + \beta \partial_x  u(t,x) &= 0, 
     ~~~ x \in (0,1), t \in (0,2], \\
     u(0,x) &= u_0(x), ~ x \in (0,1),
\end{align}
where $\beta$ is a constant advection speed. 
Note that the exact solution of the system is given as: $u(t, x) = u_0 (x - \beta t)$.

In our dataset, we only considered the periodic boundary condition. 
As an initial condition, we use a super-position of sinusoidal waves as: 
\begin{equation}
    u_0(x) = \sum_{k_i = k_1,\dots,k_N} A_{i} \sin(k_i x + \phi_i),
    \label{eq:init}
\end{equation}
where $k_i = 2 \pi \{n_i \}/L_x$ are wave numbers whose $\{ n_i \}$ are integer numbers selected randomly in $[1, n_{\rm max}]$, 
$N$ is the integer determining how many waves to be added, 
$L_x$ is the calculation domain size, 
$A_i$ is a random float number uniformly chosen in $[0, 1]$, 
and $\phi_i$ is the randomly chosen phase in $(0, 2\pi)$. 
In 1D-advection case, we set $k_{\rm max} = 8$ and $N = 2$. 
After calculating \autoref{eq:init}, 
we randomly operate the absolute value function with random signature and the window-function 
with 10\% probability, respectively. 

The numerical solution was calculated with the temporally and spatially 2nd-order upwind finite difference scheme. 

\subsection{1D Diffusion-Reaction Equation}

Here, we consider a one-dimensional diffusion-reaction type PDE, 
that combines a diffusion process and a rapid evolution from a source term \cite{krishnapriyan2021characterizing}. 
The equation is expressed as:
 \begin{align}
     \partial_t u(t,x) & - \nu \partial_{xx} u(t,x) - \rho u (1 - u) = 0,  ~~~ x  \in (0,1), t \in (0,1], \\
     u(0,x) &= u_0(x), ~~~ x \in (0,1).
 \end{align}
Note that the variable $u$ develops at potentially exponential rate because of the force term which depends on $u$. 
measure the ability to capture very rapid dynamics. 

Similar to the 1D advection equation case, 
we use the periodic boundary condition and \autoref{eq:init} as the initial condition. 
To avoid an ill-defined initial condition, 
we also applied the absolute value function and a normalization operation, dividing the initial condition by the maximum value. 
The numerical solution was calculated with the temporally and spatially 2nd-order central difference scheme.
For the source term part, we use the piecewise-exact solution (PES) method \citep{2007ApJ...658L..99I}. 

\subsection{Burgers equation}

\begin{figure}[h!]
  \centering
  \begin{subfigure}{0.48\textwidth}
    \centering
    \includegraphics[width=\textwidth]{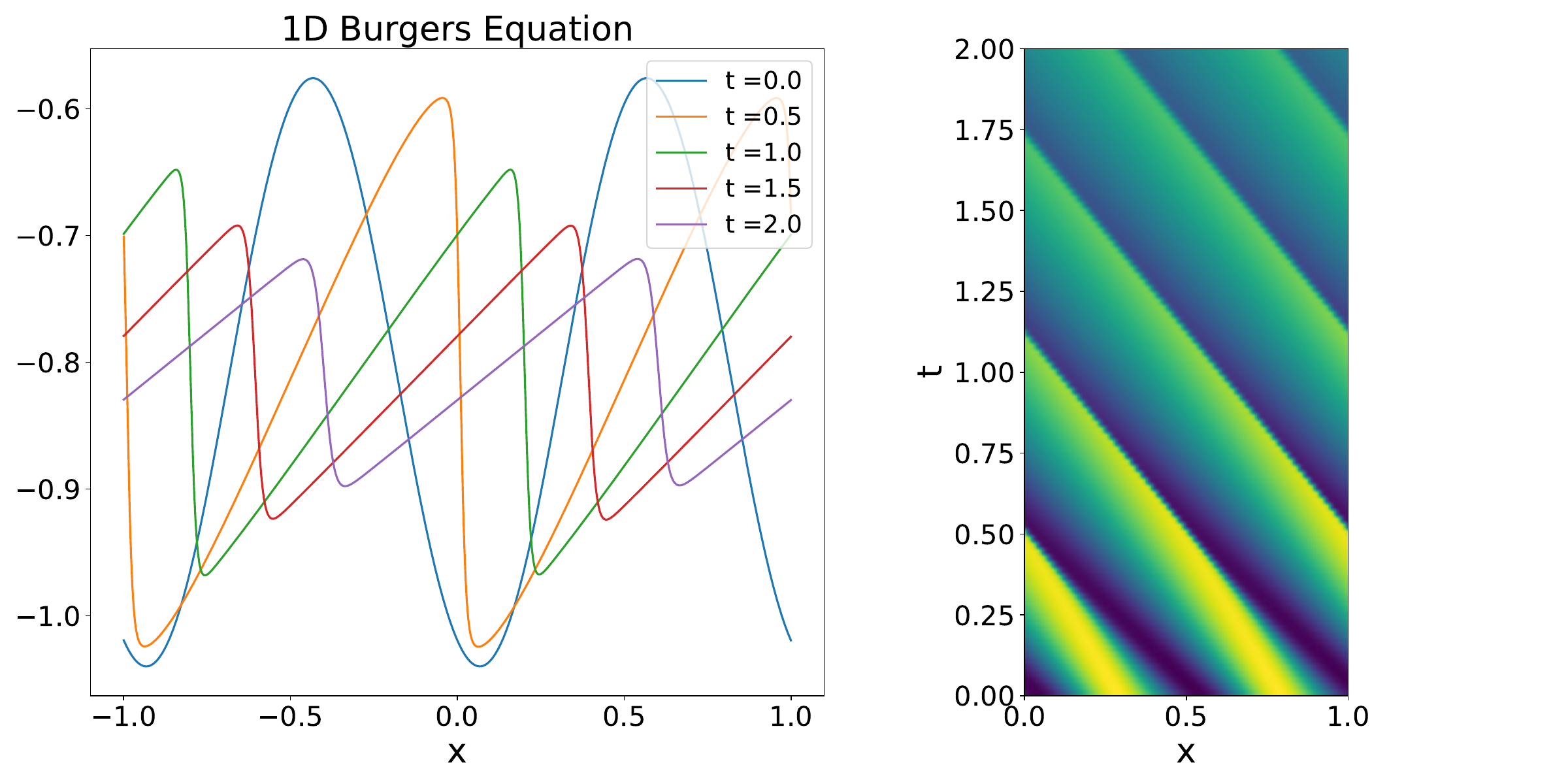}
    \caption{1D Burgers ($\nu=0.01$)}
    \label{fig:bgs_tmp}
  \end{subfigure}
  \hfill
  \begin{subfigure}{0.48\textwidth}
      \centering
      \includegraphics[width=\textwidth]{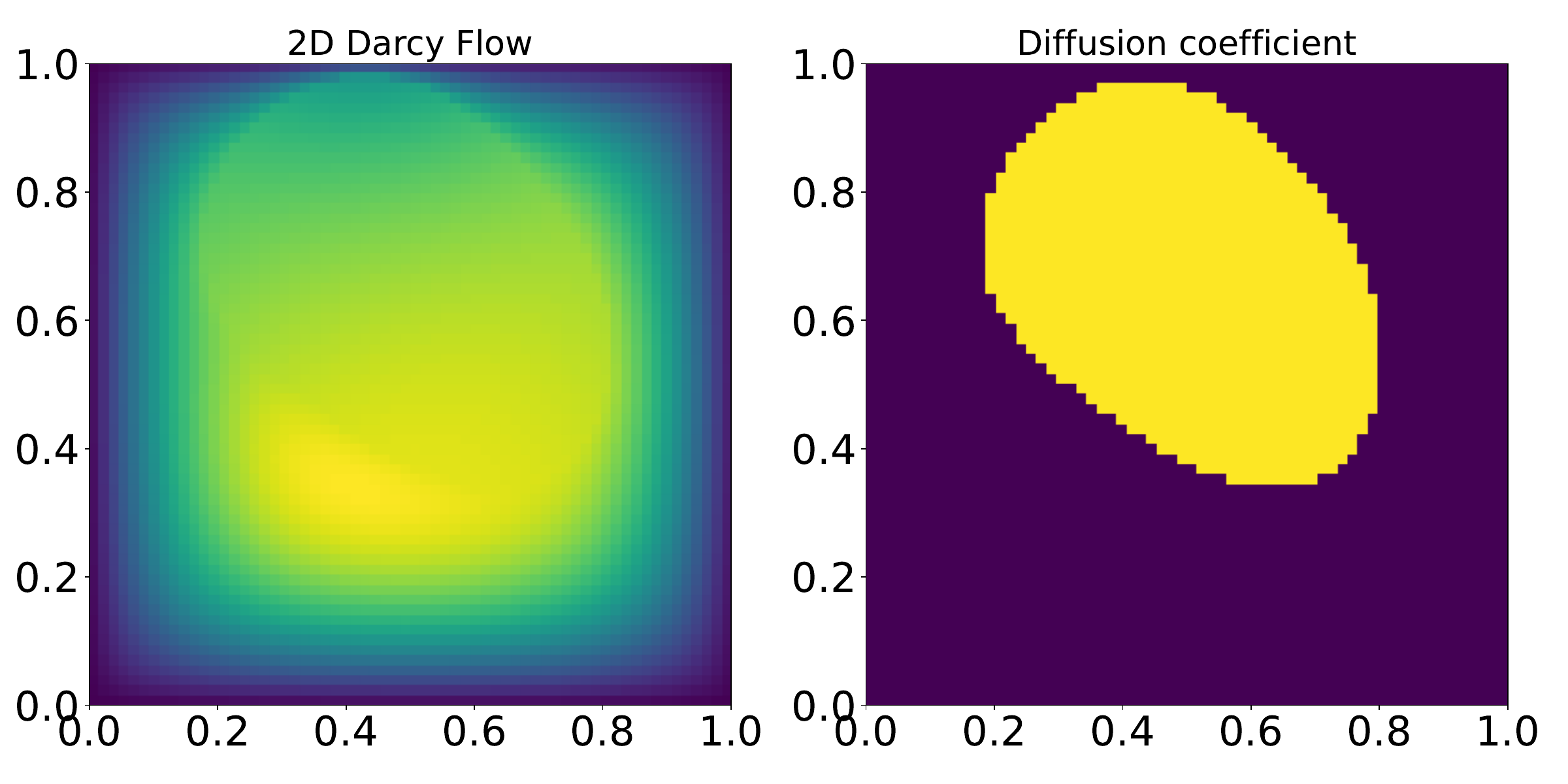}
      \caption{2D Darcy Flow ($\beta=1.0$)}
      \label{fig:dar_tmp}
  \end{subfigure}
  \caption{Visualization of the time evolution of 1D Burgers equation and 2D Darcy Flow.
  }
 \label{fig:bgs_dar_tmp}
 \end{figure}

The Burgers' equation is a PDE modeling the non-linear behavior and diffusion process in fluid dynamics as 
\begin{align}
    \partial_t u(t,x) + \partial_x  (u^2(t,x)/2) &= \nu/\pi \partial_{xx} u(t,x), ~~~ x \in (0,1), t \in (0,2], \\
    u(0,x) &= u_0(x), ~ x \in (0,1),
\end{align}
where $\nu$ is the diffusion coefficient, which is assumed constant in our dataset. 

Note that setting $R \equiv \pi u L / \nu$ describes the system's evolution as the Reynolds number of the Navier-Stokes equation \eqref{eq:nast}; $R > 1$ means the strong non-linear case support forming shock phenomena, and $R < 1$ means the diffusive case.

Similar to the 1D advection equation case, 
we use the periodic boundary condition and \autoref{eq:init} as the initial condition. 
The numerical solution was calculated with the temporally and spatially 2nd-order upwind difference scheme for the advection term, and the central difference scheme for the diffusion term.

\subsection{Darcy Flow}
We experiment with the steady-state solution of 2D Darcy Flow over the unit square, whose viscosity term $a(x)$ is an input of the system. The solution of the steady-state is defined by the following equation
\begin{align}
    - \nabla (a(x) \nabla u(x))  &= f(x), ~~~~ x \in (0,1)^2, 
    \label{eq:darcy}\\
    u(x) & =0, ~~~~ x \in \partial (0,1)^2. 
\end{align}
In this paper, the force term $f$ is set as a constant value $\beta$, 
changing the scale of the solution $u(x)$. 
Instead of directly solving \autoref{eq:darcy}, we obtained the solution by solving a temporal evolution equation: 
\begin{align}
    \partial_t u(x,t) - \nabla (a(x) \nabla u(x, t))  &= f(x), ~~~~ x \in (0,1)^2,  
\end{align}
with random field initial condition, until reaching a steady state. 
The numerical calculation was performed the same as the case of the 1D Diffusion-Reaction equation. 

\subsection{Compressible Navier-Stokes equation}
\label{sec:cfd-intro}

\begin{figure}[h!]
  \centering
  \includegraphics[width=\textwidth]{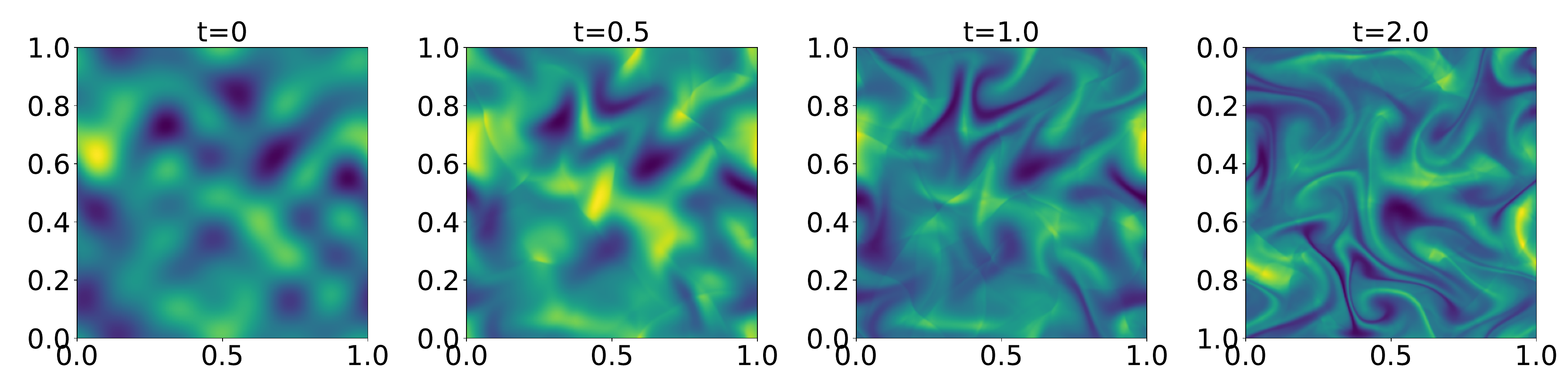}
  \caption{Visualization of the time evolution of the density in the case of 2D Compressible Navier-Stokes equations (inviscid, $M=0.1$).}
  \label{fig:2D_CFD_tmp}
 \end{figure}

The compressible fluid dynamic equations describe a fluid flow, 
 \begin{subequations}\label{eq:cnast}
 \begin{align}
     \partial_t \rho + \nabla \cdot (\rho \textbf{v}) &= 0, \label{eq:cnast-1}
     \\
     \rho (\partial_t \textbf{v} + \textbf{v} \cdot \nabla \textbf{v}) &= - \nabla p + \eta \triangle \textbf{v} + (\zeta + \eta/3) \nabla (\nabla \cdot \textbf{v}),
     \label{eq:cnast-2}\\
     \partial_t \left[ \epsilon + \frac{\rho v^2}{2} \right] &+ \nabla \cdot \left[ \left(\epsilon + p + \frac{\rho v^2}{2} \right) \bf{v} - \bf{v} \cdot \sigma' \right] = 0,\label{eq:cnast-3}
 \end{align}
\end{subequations}
where $\rho$ is the mass density, 
${\bf v}$ is the velocity, 
$p$ is the gas pressure, 
$\epsilon = p/(\Gamma - 1)$ is the internal energy, 
$\Gamma = 5/3$, 
$\sigma'$ is the viscous stress tensor, 
and $\eta, \zeta$ are the shear and bulk viscosity, respectively. 

\projectname{} provides the following training datasets for the compressible Navier-Stokes equations: 
\begin{center}
    \begin{tabularx}{9.5cm}{cccc}
        \toprule
        $N_d$ & initial field &  boundary condition & $(\eta, \zeta, M)$  \\
        \midrule
        1D &  random field & periodic & $(10^{-8}, \ 10^{-8}, \ -)$  \\
        1D &  random field & periodic & $(10^{-2}, \ 10^{-2}, \ -)$  \\
        1D &  random field & periodic & $(10^{-1}, \ 10^{-1}, \ -)$  \\
        1D &  random field & out-going & $(10^{-8}, \ 10^{-8}, \ -)$  \\
        1D &  shock-tube & out-going & $(10^{-8}, \ 10^{-8}, \ -)$  \\
        \midrule
        2D &  random field & periodic & $(10^{-8}, \ 10^{-8}, \ 0.1)$  \\
        2D &  random field & periodic & $(10^{-2}, \ 10^{-2}, \ 0.1)$  \\
        2D &  random field & periodic & $(10^{-1}, \ 10^{-1}, \ 0.1)$  \\
        2D &  random field & periodic & $(10^{-8}, \ 10^{-8}, \ 1.0)$  \\
        2D &  random field & periodic & $(10^{-2}, \ 10^{-2}, \ 1.0)$  \\
        2D &  random field & periodic & $(10^{-1}, \ 10^{-1}, \ 1.0)$  \\
        2D &  turbulence & periodic & $(10^{-8}, \ 10^{-8}, \ 0.1)$  \\
        2D &  turbulence& periodic & $(10^{-8}, \ 10^{-8}, \ 1.0)$  \\
        \midrule
        3D &  random field & periodic & $(10^{-8}, \ 10^{-8}, \ 1.0)$  \\
        3D &  random field & periodic & $(10^{-2}, \ 10^{-2}, \ 1.0)$  \\
        \bottomrule
        %\label{tab:CFD_params}
    \end{tabularx}
    
\end{center}
where $N_d$ is the number of spatial dimensions, 
$M = |v|/c_s$ is the Mach number, 
$c_s = \sqrt{\Gamma p / \rho}$ is the sound velocity. 
The outgoing  boundary condition is copying the neighbor cell to the boundary region which allows waves and fluid to escape from the computational domain, and is popular for astrohydrodynamics simulations \citep{1992ApJS...80..753S}.
The random field initial condition is applying \autoref{eq:init} which is extended to higher dimensions for the 2D and 3D cases. 
Note that density and pressure are prepared by adding a uniform background to the perturbation field \autoref{eq:init}. 
The turbulence initial condition considers turbulent velocity with uniform mass density and pressure. 
The velocity is calculated similarly to \autoref{eq:init} as 
\begin{equation}
    {\bf v}(x,t=0) = \sum_{i=1}^{n} {\bf A}_{i} \sin(k_i x + \phi_i),
    \label{eq:init_turb}
\end{equation}
where $n=4$ and $A_i = \bar{v}/|k|^d$, and $d = 1, 2$ when considering 2D and 3D, respectively. 
$\bar{v}$ is determined by the initial Mach number as $\bar{v} = c_s M$. 
To reduce the compressibility effect, we subtracted the compressible field from \autoref{eq:init_turb} by the Helmholtz-decomposition in the Fourier space. 

The shock-tube initial field is composed as $Q(x,t=0) = (Q_L, Q_R)$, where $Q = (\rho, {\bf v}, p)$ and $Q_L, Q_R$ are randomly determined constant values. The location of the initial discontinuity is also randomly determined. 
This problem is called the "Riemann problem", and the initial discontinuity generates shocks and rarefaction depending on the values of $Q_L, Q_R$, which are very difficult to obtain without solving the PDEs. This scenario can be used for a rigorous test if ML models fully understand \autoref{eq:cnast-1} - \autoref{eq:cnast-3}. 
The numerical solution was calculated with the temporally and spatially 2nd-order HLLC scheme \citep{1994ShWav...4...25T} with the MUSCL method \citep{1979JCoPh..32..101V} for the inviscid part, and the central difference scheme for the viscous part.

\subsection{Inhomogenous, incompressible Navier-Stokes}\label{sec:nast-incomp-intro}
A popular simplification of the Navier-Stokes equation is the incompressible version, commonly used to model dynamics supposed to be  far lower than the speed of propagation of waves in the medium,
\begin{align} 
    \nabla \cdot \textbf{v} &= 0, 
    ~~~ \rho (\partial_t \textbf{v} + \textbf{v} \cdot \nabla \textbf{v}) = - \nabla p + \eta \triangle \textbf{v} \label{eq:nast-incomp-1}.
\end{align}
These simplify  the compressible Navier-Stokes equations Eq.~\eqref{eq:nast}%
, by substituting the first term in Eq.~\eqref{eq:nast-incomp-1} 
instead of the first term in \eqref{eq:cnast}, 
from which we can eliminate several elements in the second terms of Eq.~\eqref{eq:nast-incomp-1}.
Additionally, we have introduced the assumption that the fluid is homogeneous (i.e. not a fluid comprising two or more substances of different density or viscosity).

We employ an augmented form of 
\eqref{eq:nast-incomp-1} which includes a vector field \emph{forcing} term \(\textbf{u}\),
\begin{align}
    \rho (\partial_t \textbf{v} + \textbf{v} \cdot \nabla \textbf{v}) &= - \nabla p + \eta \triangle \textbf{v} + \textbf{u}. 
    \label{eq:nast-incomp-2-p}
\end{align}
Non-periodic conditions are included to challenge models which perform well upon periodic domains, such as the FNO \citep{li2020fourier}.
The forcing term poses challenges based upon spatially heterogeneous dynamics.
Firstly, this allows us to see if the prediction methods can successfully learn to predict in the presence of heterogeneity.
Secondly, this permits us to use the spatially varying random field as a target for inverse inference.

Initial conditions \(\textbf{v}_0\) and inhomogeneous forcing parameters \(\textbf{u}\) are each drawn from isotropic Gaussian random fields with truncated power-law decay \(\tau\) of the power spectral density and scale \(\sigma\), where \(\tau_{\textbf{v}_0}=-3,\sigma_{\textbf{v}_0}=0.15,\tau_{\textbf{u}}=-1,\sigma_{\textbf{u}}=0.4\).
The variation in the resulting field is due to the alteration in the random seed.
We set the domain to the unit square $\Omega = \left[0, 1\right]^2$, the viscosity to \(\nu=0.01\).
Simulations are implemented using Phiflow \cite{HollPhiflow2020}.
Boundary conditions are Dirichlet, clamping field velocity to null at the perimeter.

\subsection{2D Shallow-Water Equations}
\label{sec:swe}
The shallow-water equations, derived from the general Navier-Stokes equations, present a suitable framework for modelling free-surface flow problems.
In 2D, these come in the form of the following system of hyperbolic PDEs,
\begin{subequations}
 \begin{align}
     \label{eq:swe1}
      \partial_t h + \partial_x hu + \partial_y hv &= 0\, , \\
      \label{eq:swe2}
      %\partial_t hu + \partial_x \left( u^2h + \frac{1}{2} g_r h^2 \right) &= - g_r h \partial_x b \, , \\
      \partial_t hu + \partial_x \left( u^2h + \frac{1}{2} g_r h^2 \right) + \partial_y uvh &= - g_r h \partial_x b \, , \\
      \label{eq:swe3}
      %\partial_t hv + \partial_y \left( v^2h + \frac{1}{2} g_r h^2 \right) &= - g_r h \partial_y b \, ,
      \partial_t hv + \partial_y \left( v^2h + \frac{1}{2} g_r h^2 \right)  + \partial_x uvh &= - g_r h \partial_y b \, ,
\end{align}
\end{subequations}

with $u,v$ being the velocities in horizontal and vertical direction, $h$ describing the water depth and $b$ describing a spatially varying bathymetry.
$hu, hv$ can be interpreted as the directional momentum components and $g_r$ describes the gravitational acceleration.

\begin{figure}[!]
    \centering
        \begin{subfigure}{0.24\textwidth}
         \centering
         \includegraphics[width=\textwidth]{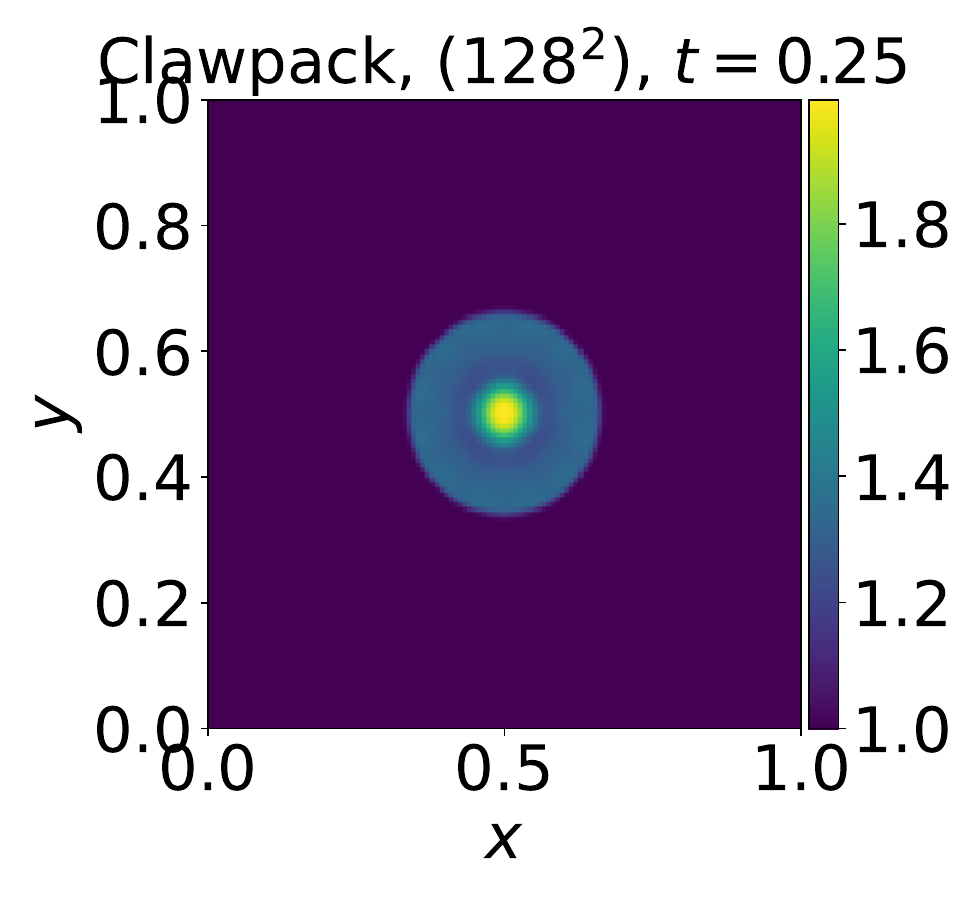}
         \caption{}%Plots of the MSE calculated at different unrolled time steps.}
    %\label{fig:vis_diff-react_data}
    \end{subfigure}
    \hfill
    \begin{subfigure}{0.24\textwidth}
         \centering
         \includegraphics[width=\textwidth]{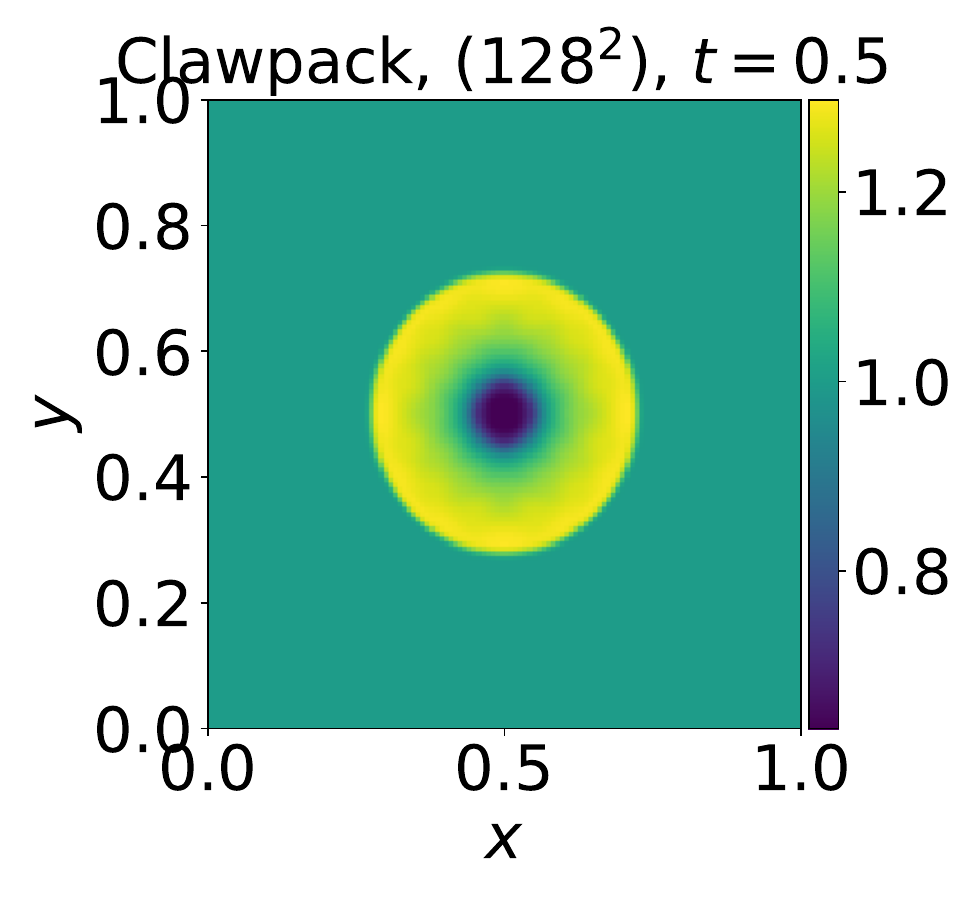}
         \caption{}
         %\label{fig:vis_diff-react_fno}
     \end{subfigure}
     \hfill
     \begin{subfigure}{0.24\textwidth}
         \centering
         \includegraphics[width=\textwidth]{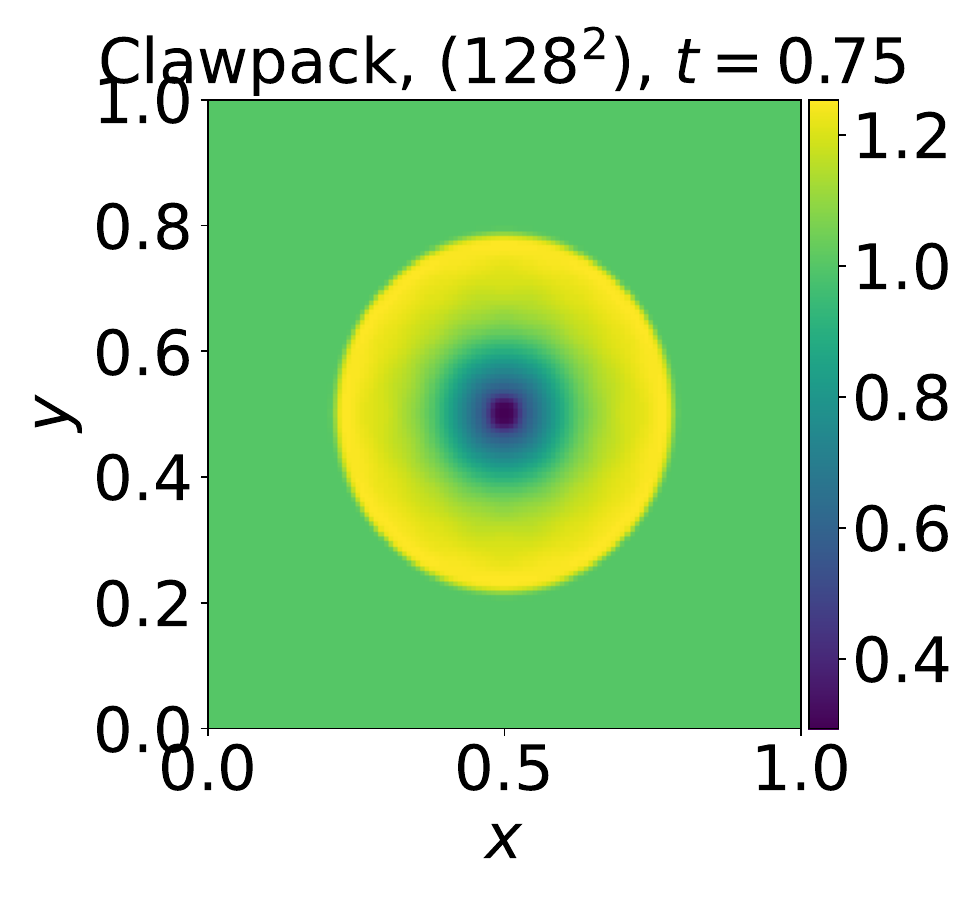}
         \caption{}
         %\label{fig:vis_diff-react_fno}
     \end{subfigure}
     \hfill
     \begin{subfigure}{0.24\textwidth}
         \centering
         \includegraphics[width=\textwidth]{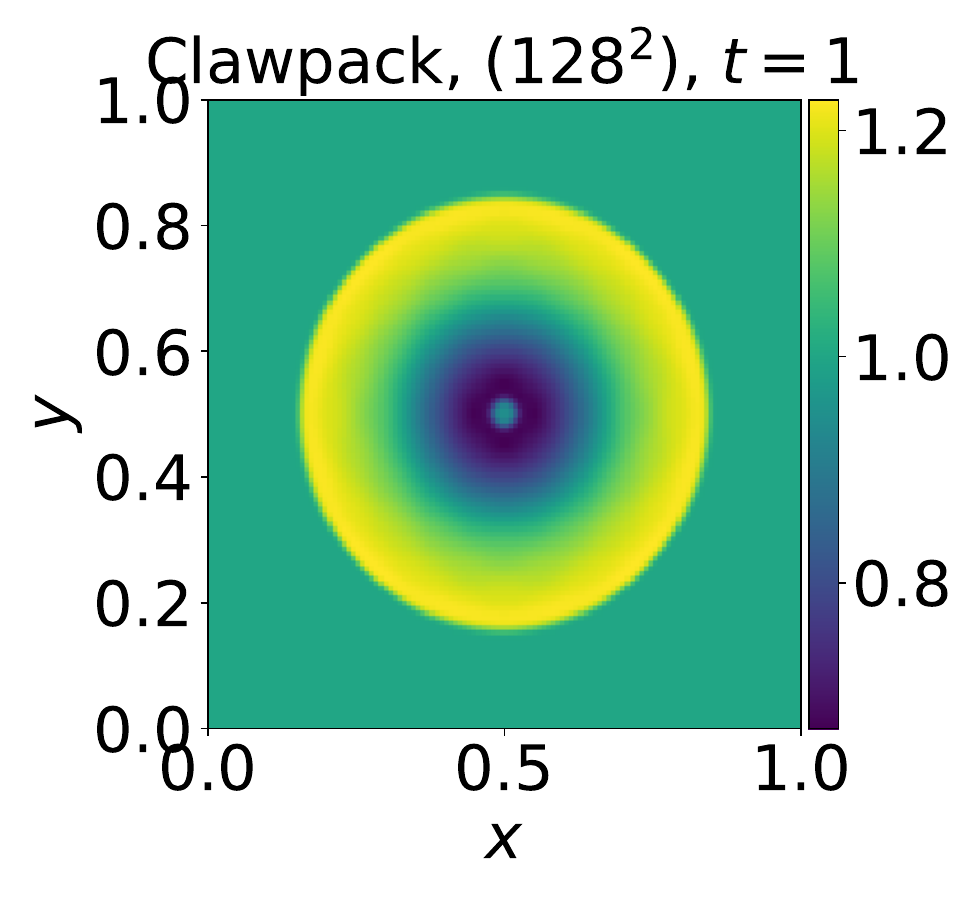}
         \caption{}
         %\label{fig:vis_diff-react_fno}
     \end{subfigure}
     \vspace{-2em} % REMOVING SUBFIGURE LABELS
     \caption{Visualization of the time evolution of the 2D shallow-water equations data.}
    \label{fig:vis_rdb_time}
\end{figure}

The specific simulation we include in our benchmark for the shallow-water equations problem as introduced in \ref{sec:swe} is a 2D radial dam break scenario.
On a square domain $\Omega = \left[-2.5, 2.5\right]^2$ we initialize the water height as a circular bump in the center of the domain
\begin{equation}
    h(t=0,x,y) = \begin{cases}
    2.0, & \text{for} \; r < \sqrt{x^2 + y^2} \\
    1.0, & \text{for} \; r \geq \sqrt{x^2 + y^2}
    \end{cases}
\end{equation}
with the radius $r$ randomly sampled from $\mathcal{U}(0.3,0.7)$.
For generating the datasets we simulate this problem using the PyClaw \cite{pyclaw} Python package which offers a comprehensive finite volume solver.
A time evolution visualization of the equation is shown in \autoref{fig:vis_rdb_time}.

\subsection{Diffusion-Sorption Equation}
\label{sec:diff-sorp}
The diffusion-sorption equation models a diffusion process which is retarded by a sorption process.
The equation is written as
\begin{align}
\label{eq:diff-sorp}
    \partial_t u(t,x) & = D/R(u) \partial_{xx} u(t,x),  ~~~ x  \in (0,1), t \in (0,500].
\end{align}
where $D$ is the effective diffusion coefficient, $R$ is the retardation factor representing the sorption that hinders the diffusion process.
Note that $R$ is dependent on the variable $u$. This equation is applicable to real world scenarios, one of the most prominent being groundwater contaminant transport.

This equation is retarded by the retardation factor $R$ which is dependent on $u$ based on the Freundlich sorption isotherm \cite{Limousin2007}:
\begin{align}
\label{eq:ret_diff-sorp}
    R(u) &= 1 + \frac{1-\phi}{\phi} \rho_s k n_f u^{n_f -1},
\end{align}
where $\phi = 0.29$ is the porosity of the porous medium, $\rho_s = 2\,880$ is the bulk density, $k=3.5 \times 10^{-4}$ is the Freundlich's parameter, $n_f = 0.874$ is the Freundlich's exponent, and the effective diffusion coefficient $D = 5 \times 10^{-4}$. The initial condition is generated with a uniform distribution $u(0,x) \sim \mathcal{U}(0,0.2)$ for $x \in (0,1)$. We provide datasets discretized into $N_x = 1024$ and $N_t = 501$, as well as the temporally downsampled version for the models training with $N_t = 101$. The spatial discretization is performed using the finite volume method \cite{Moukalled2016} and the time integration using the built-in fourth order Runge-Kutta method in the \emph{scipy} package \cite{2020SciPy-NMeth}.

\begin{figure}[!]
    \centering
        \includegraphics[width=0.25\textwidth]{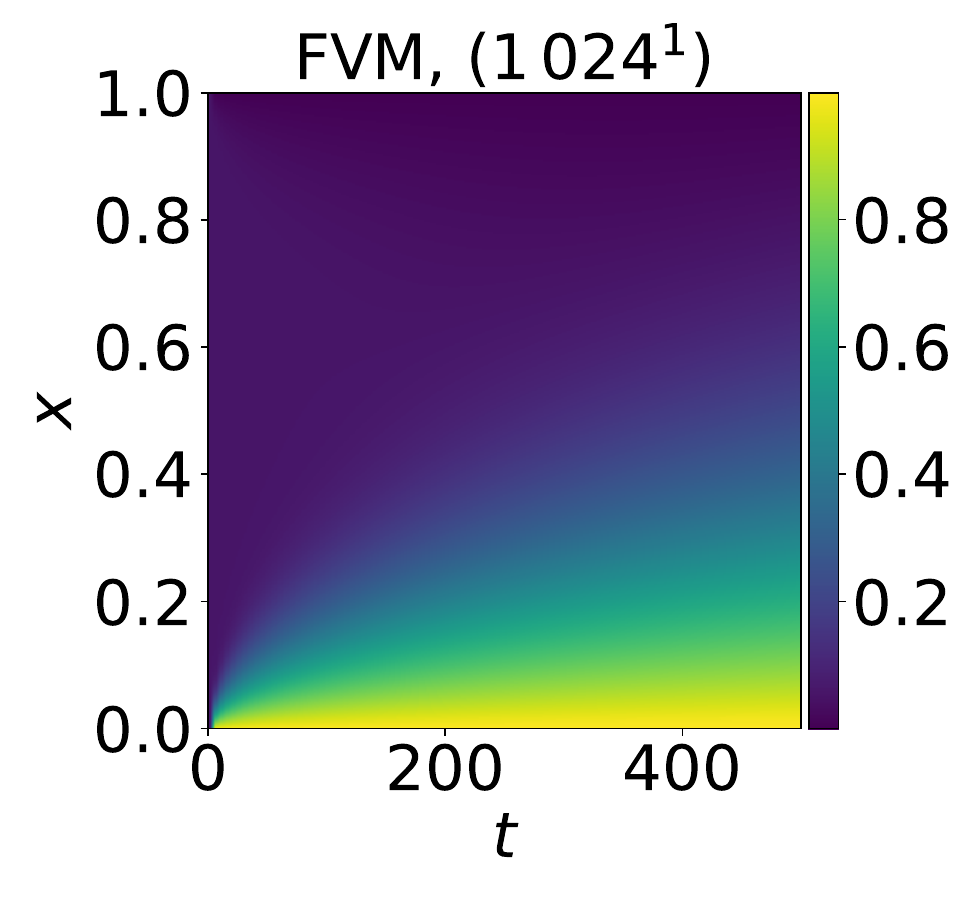}
     \caption{Visualization of the time evolution of the 1D diffusion-sorption equations data.}
    \label{fig:vis_diff-sorp_time}
\end{figure}

This particular example is interesting because of a few things. First, the diffusion coefficient becomes non-linear with dependency on $u$. And based on \autoref{eq:ret_diff-sorp}, it is clear that there is a singularity when $u=0$. Second, it is highly applicable to a real-world problem, namely the groundwater contaminant transport \cite{Nowak2016}. To date, application of machine learning to real-world physics problems is still rare. Third, we employ boundary conditions that are not the usual zero or periodic conditions that can be easily padded in models with a convolutional structure. Here, we use $u(t,0) = 1.0$ and $u(t,1) = D \partial_{x} u(t,1)$. The second boundary condition is particularly challenging since it uses a derivative instead of a constant value.
For generating the datasets we simulate this problem using a standard finite volume solver. A time evolution visualization of the equation is shown in \autoref{fig:vis_diff-sorp_time}.

\subsection{2D Diffusion-Reaction Equation}
\label{sec:2d_diff-react}
In addition to the 1D diffusion-reaction equation, which involves only a single variable, we also consider extending the application to a 2D domain, with two non-linearly coupled variables, namely the activator $u=u(t,x,y)$ and the inhibitor $v=v(t,x,y)$. The equation is written as
\begin{align} \label{eq:2d_diff-react}
    \partial_t u  = D_u \partial_{xx} u + D_u \partial_{yy} u + R_u, 
    ~~~ \partial_t v  = D_v \partial_{xx} v + D_v \partial_{yy} v + R_v \, ,
\end{align}
where $D_u$ and $D_v$ are the diffusion coefficient for the activator and inhibitor, respectively, $R_u=R_u(u,v)$ and $R_v=R_v(u,v)$ are the activator and inhibitor reaction function, respectively.
The domain of the simulation includes $x \in (-1,1), y \in (-1,1), t \in (0,5]$. This equation is applicable most prominently for modeling biological pattern formation.

The reaction functions for the activator and inhibitor are defined by the Fitzhugh-Nagumo equation \cite{Klaasen1984fhn}, written as:
\begin{align}
\label{eq:react_u_diff-sorp}
    R_u(u,v) &= u - u^3 - k - v, \\
\label{eq:react_v_diff-sorp}
    R_v(u,v) &= u - v,
\end{align}
where $k = 5 \times 10^{-3}$, and the diffusion coefficients for the activator and inhibitor are $D_u = 1 \times 10^{-3}$ and $D_v = 5 \times 10^{-3}$, respectively. The initial condition is generated as standard normal random noise $u(0,x,y) \sim \mathcal{N}(0,1.0)$ for $x \in (-1,1)$ and $y \in (-1,1)$. We provide datasets discretized into $N_x = 512$, $N_y = 512$ and $N_t = 501$, as well as the downsampled version for the models training with $N_x = 128$, $N_y = 128$, and $N_t = 101$. As in the 1D diffusion-sorption equation, the spatial discretization is performed using the finite volume method \cite{Moukalled2016}, and the time integration is performed using the built-in fourth order Runge-Kutta method in the \emph{scipy} package \cite{2020SciPy-NMeth}.

\begin{figure}[!]
    \centering
        \begin{subfigure}{0.24\textwidth}
         \centering
         \includegraphics[width=\textwidth]{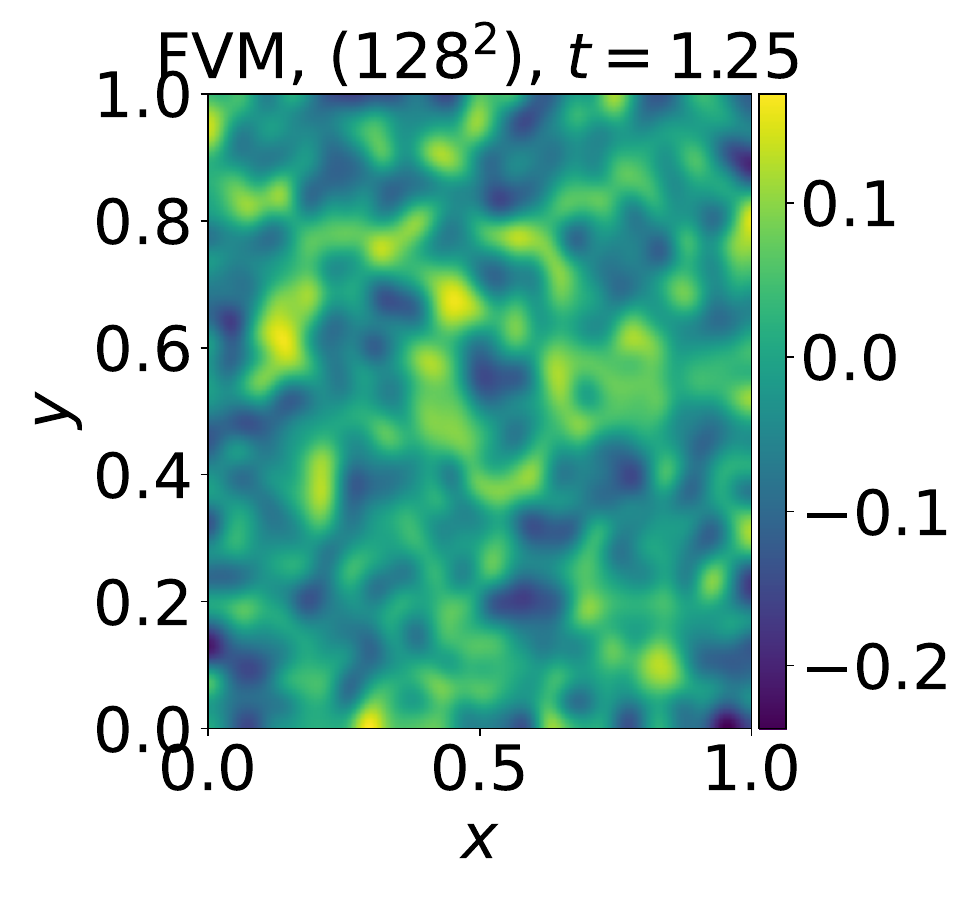}
         \caption{}%Plots of the MSE calculated at different unrolled time steps.}
    %\label{fig:vis_diff-react_data}
    \end{subfigure}
    \hfill
    \begin{subfigure}{0.24\textwidth}
         \centering
         \includegraphics[width=\textwidth]{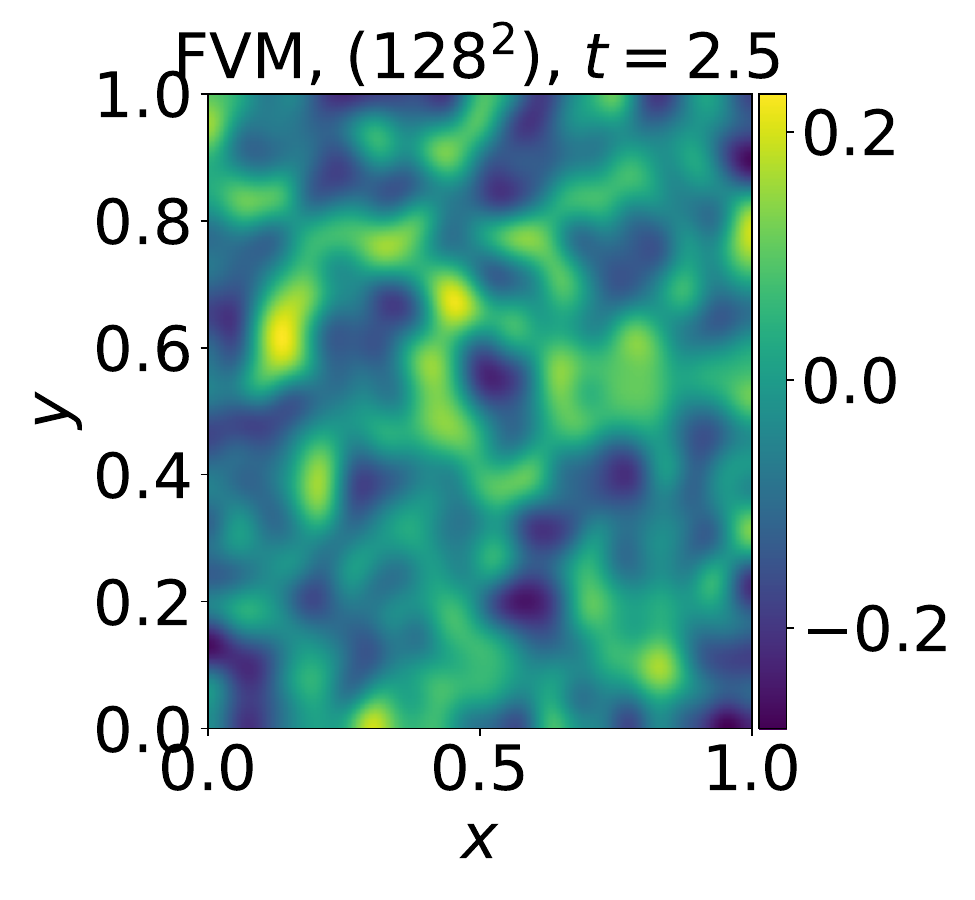}
         \caption{}
         %\label{fig:vis_diff-react_fno}
     \end{subfigure}
     \hfill
     \begin{subfigure}{0.24\textwidth}
         \centering
         \includegraphics[width=\textwidth]{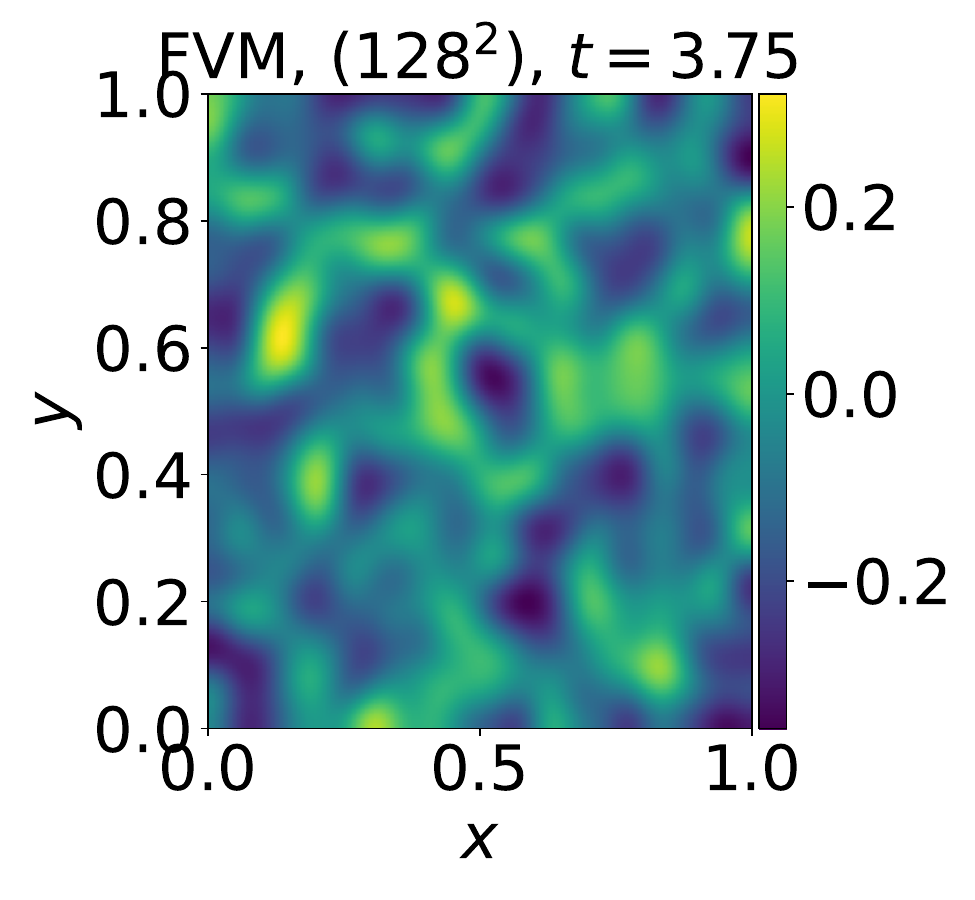}
         \caption{}
         %\label{fig:vis_diff-react_fno}
     \end{subfigure}
     \hfill
     \begin{subfigure}{0.24\textwidth}
         \centering
         \includegraphics[width=\textwidth]{figures/timothy/diff-react/2D_diff-react_NA_NA_data_4.pdf}
         \caption{}
         %\label{fig:vis_diff-react_fno}
     \end{subfigure}
     \vspace{-2em} % REMOVING SUBFIGURE LABELS
     \caption{Visualization of the time evolution of the 2D diffusion-reaction equations data.}
    \label{fig:vis_diff-react_time}
\end{figure}

We included the 2D diffusion-reaction equation as an example because it serves as a challenging benchmark problem. First, there are two variables of interest, namely the activator and inhibitor, which are non-linearly coupled. Second, it also has applicability in real-world problems, namely biological pattern formation \cite{Turing1952}. Third, we also employ a no-flow Neumann boundary condition, meaning that $D_u \partial_x u = 0$, $D_v \partial_x v = 0$, $D_u \partial_y u = 0$, and $D_v \partial_y v = 0$ for $x,y \in (-1,1)^2$.
For generating the datasets we simulate this problem using a standard finite volume solver. A time evolution visualization of the equation is shown in \autoref{fig:vis_diff-react_time}.

\begin{figure}[ht!]
    \centering
    \includegraphics[width=0.8\textwidth]{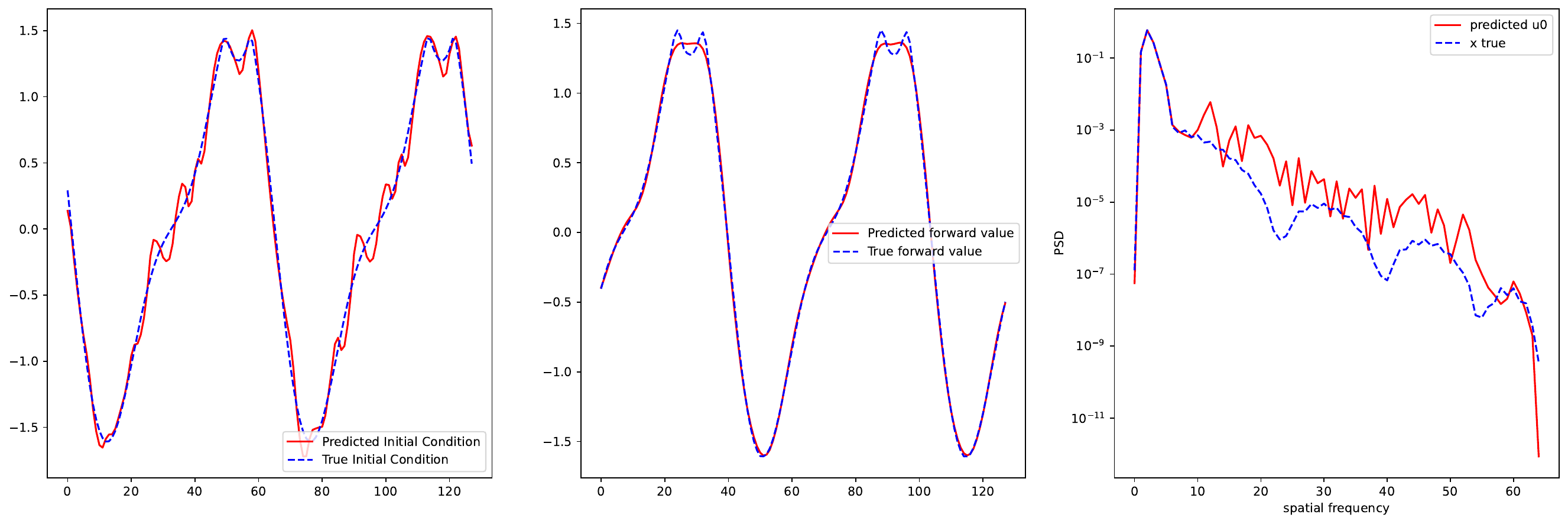}    
    \caption{Inverse problem for the 1d advection equation with $\beta=0.1$. The spectra density where most of the error is concentrated in the higher frequencies is depicted on the right.}
    \label{fig:ic11d}
\end{figure}

\subsection{Gradient-Based Inverse Method}
The inverse problem aims at solving an inverse inference by minimising the prediction loss\citep{cao2018inverse,nocedal1999numerical}, 
\begin{align*}
     & \mathcal{L}(u(t=T,x|u_0),u(t=T,x|\hat{u}_0)) \\
     & \text{where} ~~~ \hat{u}_0 \sim p_\theta(u_0|u(t=T,x))\,.
\end{align*}
The generation process $p_\theta(u_0|u(t=T,x))$ is a deterministic function, whose parameters $\theta$ use a bilinear interpolation to recover the initial condition \citep{MacKinlayModel2021}. 

\autoref{fig:ic11d} shows the solution of the inverse problem for the 1d advection equation. On the left, we see the true and estimated initial condition, and on the right the power density in the frequency domain. As we can see, the error is concentrated in the mid-high frequencies. In the middle we have the true and predicted value at time $t=T$. The error is smaller then in the plot on the left.

\autoref{tab:inverse_err1}, \autoref{tab:inverse_err2} and \autoref{tab:inverse_err3} show the error in the spatial and frequency domain of $4$ datasets and using FNO and U-Net as surrogate models. In Fig.\ref{fig:ic11d}, the left figure visualizes the true and the estimated initial condition, while the middle figure is the predicted and the true value. As shown in the figure on the right, the largest error is in the higher frequencies. This effect is also visible from the frequency metrics of Tab.\ref{tab:inverse_err2} and Tab.\ref{tab:inverse_err3}. In the experiment we use the same initial and boundary conditions of the forward problem. 

\begin{table}[]
    \centering
\begin{tabular}{llrr}
\toprule
&& \multicolumn{2}{c}{Forward model} \\
\cmidrule{3-4}
PDE & Metric &  \multicolumn{1}{c}{FNO} &    \multicolumn{1}{c}{U-Net} \\
\midrule
 \multirow{6}{*}{  Advection\_beta4 } & MSE & $2.4\times10^{-3} \pm 3.4 \times 10^{-3}$ & $1.0\times10^{+0} \pm 5.6 \times 10^{-2}$ \\
                  & nL2 & $3.9\times10^{-2} \pm 2.9 \times 10^{-2}$ & $1.0\times10^{+0} \pm 2.8 \times 10^{-2}$ \\
                  & nL3 & $4.4\times10^{-2} \pm 3.3 \times 10^{-2}$ & $1.0\times10^{+0} \pm 2.9 \times 10^{-2}$ \\
                  & MSE' & $2.9\times10^{-4} \pm 5.8 \times 10^{-4}$ & $9.9\times10^{-1} \pm 2.5 \times 10^{-2}$ \\
                  & nL2' & $1.4\times10^{-2} \pm 1.1 \times 10^{-2}$ & $1.0\times10^{+0} \pm 8.0 \times 10^{-3}$ \\
                  & nL3' & $1.6\times10^{-2} \pm 1.3 \times 10^{-2}$ & $1.0\times10^{+0} \pm 8.4 \times 10^{-3}$ \\ \midrule
 \multirow{6}{*}{  Burgers\_Nu1 } & MSE & $1.0\times10^{+0} \pm 2.2 \times 10^{-1}$ & $1.3\times10^{+0} \pm 2.3 \times 10^{-1}$ \\
                  & nL2 & $1.0\times10^{+0} \pm 1.0 \times 10^{-1}$ & $1.1\times10^{+0} \pm 1.0 \times 10^{-1}$ \\
                  & nL3 & $1.0\times10^{+0} \pm 1.0 \times 10^{-1}$ & $1.1\times10^{+0} \pm 1.1 \times 10^{-1}$ \\
                  & MSE' & $1.3\times10^{-4} \pm 2.8 \times 10^{-4}$ & $2.5\times10^{-3} \pm 1.9\times 10^{-3}$ \\
                  & nL2' & $7.0\times10^{-1} \pm 4.6 \times 10^{-1}$ & $1.6\times10^{+1} \pm 2.0 \times 10^{+1}$ \\
                  & nL3' & $7.0\times10^{-1} \pm 4.4 \times 10^{-1}$ & $1.7\times10^{+1} \pm 2.1 \times 10^{+1}$ \\ \midrule
 \multirow{6}{*}{  CFD\_Shock\_Trans } & MSE & $3.4\times10^{+0} \pm 5.3 \times 10^{-1}$ & $1.1\times10^{+2} \pm 2.0 \times 10^{+1}$ \\
                  & nL2 & $1.8\times10^{+0} \pm 1.4 \times 10^{-1}$ & $1.0\times10^{+1} \pm 1.1 \times 10^{+0}$ \\
                  & nL3 & $1.9\times10^{+0} \pm 2.7 \times 10^{-1}$ & $1.1\times10^{+1} \pm 1.6 \times 10^{+0}$ \\
                  & MSE' & $1.0\times10^{-1} \pm 5.9 \times 10^{-2}$ & $4.2\times10^{-1} \pm 9.2 \times 10^{-1}$ \\
                  & nL2' & $3.3\times10^{-1} \pm 8.5 \times 10^{-2}$ & $5.8\times10^{-1} \pm 3.9 \times 10^{-1}$ \\
                  & nL3' & $3.6\times10^{-1} \pm 9.6 \times 10^{-2}$ & $6.0\times10^{-1} \pm 4.0 \times 10^{-1}$ \\ \midrule
 \multirow{6}{*}{  ReacDiff\_Nu1\_Rho2 } & MSE & $1.7\times10^{+0} \pm 2.1 \times 10^{-1}$ & $2.0\times10^{+0} \pm 3.8 \times 10^{-1}$ \\
                  & nL2 & $1.3\times10^{+0} \pm 8.4 \times 10^{-2}$ & $1.4\times10^{+0} \pm 1.3 \times 10^{-1}$ \\
                  & nL3 & $1.3\times10^{+0} \pm 8.1 \times 10^{-2}$ & $1.5\times10^{+0} \pm 1.3 \times 10^{-1}$ \\
                  & MSE' & $5.4\times10^{-2} \pm 1.2 \times 10^{-1}$ & $6.4\times10^{-1} \pm 3.5 \times 10^{-1}$ \\
                  & nL2' & $1.2\times10^{-1} \pm 1.2 \times 10^{-1}$ & $7.3\times10^{-1} \pm 5.1 \times 10^{-2}$ \\
                  & nL3' & $1.2\times10^{-1} \pm 1.2 \times 10^{-1}$ & $7.3\times10^{-1} \pm 5.0 \times 10^{-2}$ \\
\bottomrule
\end{tabular}
    \caption{Error of the inverse problem. The prime indicates the error of the predition, for example MSE' is the MSE at time $t=T$. The MSE for example in the first row is one order of magnitude lower. nL2 and nL3 are the normalized L2 and L3 norm error, nLp = $||\hat{\bm{y}}-\bm{y}||_p / ||\bm{y}||_p, p=2,3$.}
    \label{tab:inverse_err1}
\end{table}

\begin{table}[]
    \centering
\begin{tabular}{llrr}
\toprule
&& \multicolumn{2}{c}{Forward model} \\
\cmidrule{3-4}
PDE & Metric &  \multicolumn{1}{c}{FNO} &    \multicolumn{1}{c}{U-Net} \\
\midrule
 \multirow{12}{*}{  Advection\_beta4 } & fMSE & 3.04$\times10^{-1}$ & 1.29$\times10^{+2}$ \\
                  & fMSE low & 5.56$\times10^{-1}$ & 1.29$\times10^{+2}$ \\
                  & fMSE mid & 5.26$\times10^{-2}$ & 1.29$\times10^{+2}$ \\
                  & fMSE high & 3.03$\times10^{-1}$ & 1.29$\times10^{+2}$ \\
                  & fMSE' & 3.67$\times10^{-2}$ & 9.92$\times10^{-1}$ \\
                  & fMSE' low & 1.60$\times10^{-2}$ & 9.92$\times10^{-1}$ \\
                  & fMSE' mid & 5.74$\times10^{-2}$ & 9.92$\times10^{-1}$ \\
                  & fMSE' high & 3.68$\times10^{-2}$ & 9.92$\times10^{-1}$ \\
                  & fL2 & 3.91$\times10^{-2}$ & 1.00$\times10^{+0}$ \\
                  & fL2 low & 3.75$\times10^{-2}$ & 1.01$\times10^{+0}$ \\
                  & fL2 mid & 1.41$\times10^{+1}$ & 0.00$\times10^{+0}$ \\
                  & fL2 high & 3.90$\times10^{-2}$ & 0.00$\times10^{+0}$ \\ \midrule
 \multirow{12}{*}{  Burgers\_Nu1 } & fMSE & 1.29$\times10^{+2}$ & 1.59$\times10^{+2}$ \\
                  & fMSE low & 2.58$\times10^{+2}$ & 1.59$\times10^{+2}$ \\
                  & fMSE mid & 1.19$\times10^{-1}$ & 1.59$\times10^{+2}$ \\
                  & fMSE high & 1.29$\times10^{+2}$ & 1.59$\times10^{+2}$ \\
                  & fMSE' & 1.67$\times10^{-2}$ & 2.46$\times10^{-3}$ \\
                  & fMSE' low & 3.36$\times10^{-2}$ & 2.46$\times10^{-3}$ \\
                  & fMSE' mid & 9.26$\times10^{-7}$ & 2.46$\times10^{-3}$ \\
                  & fMSE' high & 1.66$\times10^{-2}$ & 2.46$\times10^{-3}$ \\
                  & fL2 & 9.98$\times10^{-1}$ & 1.11$\times10^{+0}$ \\
                  & fL2 low & 9.98$\times10^{-1}$ & 1.11$\times10^{+0}$ \\
                  & fL2 mid & 3.50$\times10^{+0}$ & 0.00$\times10^{+0}$ \\
                  & fL2 high & 9.98$\times10^{-1}$ & 0.00$\times10^{+0}$ \\ \midrule
 \multirow{12}{*}{  CFD\_Shock\_Trans } & fMSE & 4.37$\times10^{+2}$ & 1.40$\times10^{+4}$ \\
                  & fMSE low & 4.37$\times10^{+2}$ & 1.40$\times10^{+4}$ \\
                  & fMSE mid & 4.37$\times10^{+2}$ & 1.40$\times10^{+4}$ \\
                  & fMSE high & 4.37$\times10^{+2}$ & 1.40$\times10^{+4}$ \\
                  & fMSE' & 1.28$\times10^{+1}$ & 2.19$\times10^{+2}$ \\
                  & fMSE' low & 3.21$\times10^{+1}$ & 2.19$\times10^{+2}$ \\
                  & fMSE' mid & 1.13$\times10^{+0}$ & 2.19$\times10^{+2}$ \\
                  & fMSE' high & 8.98$\times10^{+0}$ & 2.19$\times10^{+2}$ \\
                  & fL2 & 1.84$\times10^{+0}$ & 1.04$\times10^{+1}$ \\
                  & fL2 low & 1.51$\times10^{+0}$ & 9.95$\times10^{+0}$ \\
                  & fL2 mid & 0.00$\times10^{+0}$ & 0.00$\times10^{+0}$ \\
                  & fL2 high & 0.00$\times10^{+0}$ & 0.00$\times10^{+0}$ \\ \midrule
 \multirow{12}{*}{  ReacDiff\_Nu1\_Rho2 } & fMSE & 2.17$\times10^{+2}$ & 2.55$\times10^{+2}$ \\
                  & fMSE low & 6.10$\times10^{+2}$ & 2.55$\times10^{+2}$ \\
                  & fMSE mid & 1.48$\times10^{-2}$ & 2.55$\times10^{+2}$ \\
                  & fMSE high & 1.28$\times10^{+2}$ & 2.55$\times10^{+2}$ \\
                  & fMSE' & 6.94$\times10^{+0}$ & 6.35$\times10^{-1}$ \\
                  & fMSE' low & 2.77$\times10^{+1}$ & 6.35$\times10^{-1}$ \\
                  & fMSE' mid & 1.14$\times10^{-5}$ & 6.35$\times10^{-1}$ \\
                  & fMSE' high & 1.29$\times10^{-4}$ & 6.35$\times10^{-1}$ \\
                  & fL2 & 1.30$\times10^{+0}$ & 1.41$\times10^{+0}$ \\
                  & fL2 low & 1.54$\times10^{+0}$ & 1.60$\times10^{+0}$ \\
                  & fL2 mid & 7.45$\times10^{+0}$ & 0.00$\times10^{+0}$ \\
                  & fL2 high & 1.00$\times10^{+0}$ & 0.00$\times10^{+0}$ \\
\bottomrule
\end{tabular}
    \caption{Frequency error of the inverse problem. fMSE, fL2 and fL3 are the frequency version of the MSE, normalized L2 and L3 norm metrics. Low, mid and high is the range of frequencies.  Prime is used for the error in the prediction, without the error of the initial condition estimation. Normalised metric are not well defined, when the original signal is zero. }
    \label{tab:inverse_err2}
\end{table}
\begin{table}[]
    \centering
\begin{tabular}{llrr}
\toprule
&& \multicolumn{2}{c}{Forward model} \\
\cmidrule{3-4}
PDE & Metric &  \multicolumn{1}{c}{FNO} &    \multicolumn{1}{c}{U-Net} \\
\midrule
 \multirow{12}{*}{ Advection\_beta4 } & fL2' & 1.36$\times10^{-2}$ & 1.13$\times10^{+1}$ \\
                  & fL2' low & 7.50$\times10^{-3}$ & 5.66$\times10^{+0}$ \\
                  & fL2' mid & 1.61$\times10^{+0}$ & 5.27$\times10^{+1}$ \\
                  & fL2' high & 1.36$\times10^{-2}$ & 8.01$\times10^{+0}$ \\
                  & fL3 & 3.14$\times10^{-2}$ & 1.00$\times10^{+0}$ \\
                  & fL3 low & 3.12$\times10^{-2}$ & 1.00$\times10^{+0}$ \\
                  & fL3 mid & 1.75$\times10^{+1}$ & 0.00$\times10^{+0}$ \\
                  & fL3 high & 3.14$\times10^{-2}$ & 0.00$\times10^{+0}$ \\
                  & fL3' & 9.51$\times10^{-3}$ & 5.04$\times10^{+0}$ \\
                  & fL3' low & 5.62$\times10^{-3}$ & 3.18$\times10^{+0}$ \\
                  & fL3' mid & 1.47$\times10^{+0}$ & 2.77$\times10^{+1}$ \\
                  & fL3' high & 9.51$\times10^{-3}$ & 4.00$\times10^{+0}$ \\ \midrule
 \multirow{12}{*}{  Burgers\_Nu1 } & fL2' & 7.00$\times10^{-1}$ & 1.82$\times10^{+2}$ \\
                  & fL2' low & 7.99$\times10^{-1}$ & 6.28$\times10^{+1}$ \\
                  & fL2' mid & 1.03$\times10^{+2}$ & 5.84$\times10^{+5}$ \\
                  & fL2' high & 5.39$\times10^{-1}$ & 1.31$\times10^{+2}$ \\
                  & fL3 & 9.97$\times10^{-1}$ & 1.04$\times10^{+0}$ \\
                  & fL3 low & 9.97$\times10^{-1}$ & 1.04$\times10^{+0}$ \\
                  & fL3 mid & 3.58$\times10^{+0}$ & 0.00$\times10^{+0}$ \\
                  & fL3 high & 9.97$\times10^{-1}$ & 0.00$\times10^{+0}$ \\
                  & fL3' & 7.21$\times10^{-1}$ & 4.88$\times10^{+1}$ \\
                  & fL3' low & 7.99$\times10^{-1}$ & 2.40$\times10^{+1}$ \\
                  & fL3' mid & 9.35$\times10^{+1}$ & 2.98$\times10^{+5}$ \\
                  & fL3' high & 5.39$\times10^{-1}$ & 3.92$\times10^{+1}$ \\ \midrule
 \multirow{12}{*}{   CFD\_Shock\_Trans } & fL2' & 3.34$\times10^{-1}$ & 2.12$\times10^{+0}$ \\
                  & fL2' low & 2.68$\times10^{-1}$ & 2.14$\times10^{+0}$ \\
                  & fL2' mid & 0.00$\times10^{+0}$ & 0.00$\times10^{+0}$ \\
                  & fL2' high & 0.00$\times10^{+0}$ & 0.00$\times10^{+0}$ \\
                  & fL3 & 1.26$\times10^{+0}$ & 9.41$\times10^{+0}$ \\
                  & fL3 low & 1.11$\times10^{+0}$ & 9.36$\times10^{+0}$ \\
                  & fL3 mid & 0.00$\times10^{+0}$ & 0.00$\times10^{+0}$ \\
                  & fL3 high & 0.00$\times10^{+0}$ & 0.00$\times10^{+0}$ \\
                  & fL3' & 2.16$\times10^{-1}$ & 2.19$\times10^{+0}$ \\
                  & fL3' low & 1.96$\times10^{-1}$ & 2.20$\times10^{+0}$ \\
                  & fL3' mid & 0.00$\times10^{+0}$ & 0.00$\times10^{+0}$ \\
                  & fL3' high & 0.00$\times10^{+0}$ & 0.00$\times10^{+0}$ \\ \midrule
 \multirow{12}{*}{ReacDiff\_Nu1\_Rho2 }  & fL2' & 1.23$\times10^{-1}$ & 1.18$\times10^{+1}$ \\
                  & fL2' low & 1.23$\times10^{-1}$ & 5.83$\times10^{+0}$ \\
                  & fL2' mid & 1.89$\times10^{+18}$ & 1.90$\times10^{+21}$ \\
                  & fL2' high & 9.03$\times10^{+18}$ & 3.93$\times10^{+21}$ \\
                  & fL3 & 1.27$\times10^{+0}$ & 1.29$\times10^{+0}$ \\
                  & fL3 low & 1.45$\times10^{+0}$ & 1.47$\times10^{+0}$ \\
                  & fL3 mid & 7.07$\times10^{+0}$ & 0.00$\times10^{+0}$ \\
                  & fL3 high & 1.00$\times10^{+0}$ & 0.00$\times10^{+0}$ \\
                  & fL3' & 1.23$\times10^{-1}$ & 5.08$\times10^{+0}$ \\
                  & fL3' low & 1.23$\times10^{-1}$ & 3.18$\times10^{+0}$ \\
                  & fL3' mid & 1.07$\times10^{+18}$ & 7.25$\times10^{+20}$ \\
                  & fL3' high & 6.54$\times10^{+18}$ & 1.14$\times10^{+21}$ \\
\bottomrule
\end{tabular}
    \caption{Frequency error of the prediction of the inverse problem. fMSE, fL2 and fL3 are the frequency version sof the MSE, normalized L2 and L3 norm metrics. Low, mid and high is the range of the frequencies. Prime is used for the error in the prediction, without the error of the initial condition estimation. Normalised metric are not well defined, when the original signal is zero.}
    \label{tab:inverse_err3}
\end{table}

% \clearpage
\FloatBarrier
\section{Detailed Baseline Score}
\label{sec:app:baseline}

\begin{table*}[h!]
    \caption{Summary of the baseline models' performance for different evaluation metrics: RMSE, normalised RMSE (nRMSE), RMSE from conserved value (cRMSE), maximum error, RMSE at the boundaries (bRMSE), RMSE in Fourier space at low (fRMSE low), medium (fRMSE mid), and high frequency (fRMSE high) ranges applied to the diffusion-sorption, 2D diffusion-reaction, and shallow-water equations.}
    \label{tab:baseline}
    \centering
    \begin{tabularx}{\textwidth}{lllrrr}
        \toprule
        & & & \multicolumn{3}{c}{Baseline model} \\
        \cmidrule{4-6}
        PDE & Parameter & Metric & \multicolumn{1}{c}{U-Net} & \multicolumn{1}{c}{FNO} & \multicolumn{1}{c}{PINN} \\
        \midrule
       \multirow{8}{*}{Diffusion-sorption} & \multirow{8}{*}{N/A} & RMSE & $5.8\times10^{-2}$ & $5.9\times10^{-4}$ & $9.9\times10^{-2}$ \\
        & & nRMSE & $1.5\times10^{-1}$ & $1.7\times10^{-3}$ & $2.2\times10^{-1}$ \\
        & & max error & $2.9\times10^{-1}$ & $7.8\times10^{-3}$ & $2.2\times10^{-1}$ \\
        & & cRMSE & $4.8\times10^{-2}$ & $1.9\times10^{-4}$ & $7.5\times10^{-2}$ \\
        & & bRMSE & $6.1\times10^{-3}$ & $2.0\times10^{-3}$ & $1.4\times10^{-1}$ \\
        & & fRMSE low & $1.9\times10^{-2}$ & $1.5\times10^{-4}$ & $3.5\times10^{-2}$ \\
        & & fRMSE mid & $4.7\times10^{-3}$ & $5.0\times10^{-5}$ & $5.2\times10^{-3}$ \\
        & & fRMSE high & $1.9\times10^{-4}$ & $7.1\times10^{-6}$ & $2.7\times10^{-4}$ \\
        \midrule
        \multirow{8}{*}{2D diffusion-reaction} & \multirow{8}{*}{N/A} & RMSE & $6.1\times10^{-2}$ & $8.1\times10^{-3}$ & $1.9\times10^{-1}$ \\
        & & nRMSE & $8.4\times10^{-1}$ & $1.2\times10^{-1}$ & $1.6\times10^{+0}$ \\
        & & max error & $1.9\times10^{-1}$ & $9.1\times10^{-2}$ & $5.0\times10^{-1}$ \\
        & & cRMSE & $3.9\times10^{-2}$ & $1.7\times10^{-3}$ & $1.3\times10^{-1}$ \\
        & & bRMSE & $7.8\times10^{-2}$ & $2.7\times10^{-2}$ & $2.2\times10^{-1}$ \\
        & & fRMSE low & $1.7\times10^{-2}$ & $8.2\times10^{-4}$ & $5.7\times10^{-2}$ \\
        & & fRMSE mid & $5.4\times10^{-3}$ & $7.7\times10^{-4}$ & $1.3\times10^{-2}$ \\
        & & fRMSE high & $6.8\times10^{-4}$ & $4.1\times10^{-4}$ & $1.5\times10^{-3}$ \\
        \midrule
        \multirow{8}{*}{Shallow-water equation} & \multirow{8}{*}{N/A} & RMSE & $8.6\times10^{-2}$ & $4.5\times10^{-3}$ & $1.7\times10^{-2}$ \\
        & & nRMSE & $8.3\times10^{-2}$ & $4.4\times10^{-3}$ & $1.7\times10^{-2}$ \\
        & & max error & $4.4\times10^{-1}$ & $4.5\times10^{-2}$ & $1.3\times10^{-3}$ \\
        & & cRMSE & $1.3\times10^{-2}$ & $2.0\times10^{-4}$ & $1.7\times10^{-2}$ \\
        & & bRMSE & $4.2\times10^{-3}$ & $1.4\times10^{-3}$ & $1.5\times10^{-1}$ \\
        & & fRMSE low & $2.0\times10^{-2}$ & $2.6\times10^{-4}$ & $5.9\times10^{-3}$ \\
        & & fRMSE mid & $7.0\times10^{-3}$ & $3.1\times10^{-4}$ & $1.9\times10^{-3}$ \\
        & & fRMSE high & $8.6\times10^{-4}$ & $2.5\times10^{-4}$ & $6.0\times10^{-4}$ \\
        \bottomrule
    \end{tabularx}
\end{table*}

\begin{table*}[h!]
    \caption{Summary of the baseline models' performance for different evaluation metrics: RMSE, normalised RMSE (nRMSE), RMSE from conserved value (cRMSE), maximum error, RMSE at the boundaries (bRMSE), RMSE in Fourier space at low (fRMSE low), medium (fRMSE mid), and high frequency (fRMSE high) ranges applied to the advection equation with different parameter values.}
    \label{tab:baseline_advection}
    \centering
    \begin{tabularx}{\textwidth}{lp{3.3cm}lrrr}
        \toprule
        & & & \multicolumn{3}{c}{Baseline model} \\
        \cmidrule{4-6}
        PDE & Parameter & Metric & \multicolumn{1}{c}{U-Net} & \multicolumn{1}{c}{FNO} & \multicolumn{1}{c}{PINN} \\
        \midrule
        \multirow{32}{*}{Advection} & \multirow{8}{3.3cm}{$\beta=0.1$} 
          & RMSE & $3.1\times10^{-2}$ & $4.1\times10^{-3}$ & $6.7\times10^{-3}$ \\
        & & nRMSE & $5.0\times10^{-2}$ & $7.7\times10^{-3}$ & $7.8\times10^{-3}$ \\
        & & max error & $5.1\times10^{-1}$ & $1.1\times10^{-1}$ & $2.0\times10^{-2}$ \\
        & & cRMSE & $1.5\times10^{-2}$ & $3.8\times10^{-4}$ & $1.5\times10^{-3}$ \\
        & & bRMSE & $6.6\times10^{-2}$ & $4.0\times10^{-3}$ & $1.7\times10^{-2}$ \\
        & & fRMSE low & $8.7\times10^{-2}$ & $3.5\times10^{-4}$ & $2.2\times10^{-3}$ \\
        & & fRMSE mid & $4.5\times10^{-3}$ & $4.4\times10^{-4}$ & $3.9\times10^{-4}$ \\
        & & fRMSE high & $9.5\times10^{-4}$ & $2.4\times10^{-4}$ & $4.8\times10^{-6}$ \\
        \cmidrule{2-6}
        & \multirow{8}{3.3cm}{$\beta=0.4$} 
          & RMSE & $1.5\times10^{-1}$ & $5.3\times10^{-3}$ & $2.6\times10^{-2}$ \\
        & & nRMSE & $2.3\times10^{-1}$ & $1.0\times10^{-2}$ & $3.0\times10^{-2}$ \\
        & & max error & $8.8\times10^{-1}$ & $1.6\times10^{-1}$ & $7.0\times10^{-2}$ \\
        & & cRMSE & $6.1\times10^{-2}$ & $4.2\times10^{-4}$ & $7.7\times10^{-3}$ \\
        & & bRMSE & $1.4\times10^{-1}$ & $4.6\times10^{-3}$ & $3.9\times10^{-2}$ \\
        & & fRMSE low & $5.5\times10^{-2}$ & $4.3\times10^{-4}$ & $6.6\times10^{-3}$ \\
        & & fRMSE mid & $1.3\times10^{-2}$ & $4.5\times10^{-4}$ & $3.3\times10^{-3}$ \\
        & & fRMSE high & $1.0\times10^{-3}$ & $3.1\times10^{-4}$ & $2.6\times10^{-5}$ \\
        \cmidrule{2-6}
        & \multirow{8}{3.3cm}{$\beta=1.0$} 
          & RMSE & $1.4\times10^{-1}$ & $5.2\times10^{-3}$ & $1.1\times10^{-2}$ \\
        & & nRMSE & $2.3\times10^{-1}$ & $9.7\times10^{-3}$ & $1.3\times10^{-2}$ \\
        & & max error & $9.5\times10^{-1}$ & $2.0\times10^{-1}$ & $2.0\times10^{-2}$ \\
        & & cRMSE & $5.2\times10^{-2}$ & $4.5\times10^{-4}$ & $2.7\times10^{-3}$ \\
        & & bRMSE & $1.4\times10^{-1}$ & $4.6\times10^{-3}$ & $8.3\times10^{-3}$ \\
        & & fRMSE low & $5.7\times10^{-2}$ & $3.5\times10^{-4}$ & $3.0\times10^{-3}$ \\
        & & fRMSE mid & $1.3\times10^{-2}$ & $3.7\times10^{-4}$ & $9.4\times10^{-4}$ \\
        & & fRMSE high & $9.8\times10^{-4}$ & $3.0\times10^{-4}$ & $4.8\times10^{-6}$ \\
        \cmidrule{2-6}
         & \multirow{8}{3.3cm}{$\beta=4.0$} 
           & RMSE & $1.3\times10^{-2}$ & $3.9\times10^{-3}$ & $6.6\times10^{-1}$ \\
        & & nRMSE & $2.4\times10^{-2}$ & $6.7\times10^{-3}$ & $7.7\times10^{-1}$ \\
        & & max error & $1.3\times10^{-1}$ & $9.0\times10^{-2}$ & $1.0\times10^{+0}$ \\
        & & cRMSE & $4.1\times10^{-3}$ & $2.4\times10^{-4}$ & $1.8\times10^{-2}$ \\
        & & bRMSE & $2.4\times10^{-2}$ & $3.1\times10^{-3}$ & $4.9\times10^{-1}$ \\
        & & fRMSE low & $3.6\times10^{-3}$ & $3.0\times10^{-4}$ & $1.5\times10^{-1}$ \\
        & & fRMSE mid & $1.7\times10^{-3}$ & $3.0\times10^{-4}$ & $2.1\times10^{-4}$ \\
        & & fRMSE high & $3.9\times10^{-4}$ & $2.2\times10^{-4}$ & $7.2\times10^{-6}$ \\
        \bottomrule
    \end{tabularx}
\end{table*}

\begin{table*}[h!]
    \caption{Summary of the baseline models' performance for different evaluation metrics: RMSE, normalised RMSE (nRMSE), RMSE from conserved value (cRMSE), maximum error, RMSE at the boundaries (bRMSE), RMSE in Fourier space at low (fRMSE low), medium (fRMSE mid), and high frequency (fRMSE high) ranges applied to the Burgers' equation with different parameter values.}
    \label{tab:baseline_burgers}
    \centering
    \begin{tabularx}{\textwidth}{lp{3.3cm}lrrr}
        \toprule
        & & & \multicolumn{3}{c}{Baseline model} \\
        \cmidrule{4-6}
        PDE & Parameter & Metric & \multicolumn{1}{c}{U-Net} & \multicolumn{1}{c}{FNO} & \multicolumn{1}{c}{PINN} \\
        \midrule
        \multirow{32}{*}{Burgers'} & \multirow{8}{3.3cm}{$\nu=0.001$} 
          & RMSE & $1.3\times10^{-1}$ & $9.6\times10^{-3}$ & $2.2\times10^{-1}$ \\
        & & nRMSE & $3.7\times10^{-1}$ & $2.9\times10^{-2}$ & $3.9\times10^{-1}$ \\
        & & max error & $6.0\times10^{-1}$ & $2.3\times10^{-1}$ & $4.3\times10^{-1}$ \\
        & & cRMSE & $8.5\times10^{-2}$ & $8.6\times10^{-4}$ & $1.3\times10^{-1}$ \\
        & & bRMSE & $1.2\times10^{-1}$ & $9.1\times10^{-3}$ & $2.1\times10^{-1}$ \\
        & & fRMSE low & $4.6\times10^{-2}$ & $8.5\times10^{-4}$ & $7.8\times10^{-2}$ \\
        & & fRMSE mid & $1.1\times10^{-2}$ & $1.1\times10^{-3}$ & $1.1\times10^{-2}$ \\
        & & fRMSE high & $1.5\times10^{-3}$ & $5.1\times10^{-4}$ & $2.7\times10^{-4}$ \\
        \cmidrule{2-6}
        & \multirow{8}{3.3cm}{$\nu=0.01$} 
          & RMSE & $7.0\times10^{-2}$ & $2.7\times10^{-3}$ & $4.7\times10^{-1}$ \\
        & & nRMSE & $2.2\times10^{-1}$ & $7.8\times10^{-3}$ & $8.5\times10^{-1}$ \\
        & & max error & $4.6\times10^{-1}$ & $6.4\times10^{-2}$ & $6.8\times10^{-1}$ \\
        & & cRMSE & $2.7\times10^{-2}$ & $5.0\times10^{-4}$ & $4.7\times10^{-1}$ \\
        & & bRMSE & $7.1\times10^{-2}$ & $4.0\times10^{-3}$ & $6.8\times10^{-1}$ \\
        & & fRMSE low & $2.6\times10^{-2}$ & $5.3\times10^{-4}$ & $1.3\times10^{-1}$ \\
        & & fRMSE mid & $7.1\times10^{-3}$ & $4.7\times10^{-4}$ & $7.9\times10^{-3}$ \\
        & & fRMSE high & $4.9\times10^{-4}$ & $8.7\times10^{-5}$ & $6.7\times10^{-5}$ \\
        \cmidrule{2-6}
        & \multirow{8}{3.3cm}{$\nu=0.1$} 
          & RMSE & $4.6\times10^{-2}$ & $7.6\times10^{-4}$ & $2.5\times10^{-1}$ \\
        & & nRMSE & $2.3\times10^{-1}$ & $2.9\times10^{-3}$ & $4.6\times10^{-1}$ \\
        & & max error & $2.9\times10^{-1}$ & $9.6\times10^{-3}$ & $3.3\times10^{-1}$ \\
        & & cRMSE & $2.5\times10^{-2}$ & $2.2\times10^{-4}$ & $2.4\times10^{-1}$ \\
        & & bRMSE & $6.2\times10^{-2}$ & $1.1\times10^{-3}$ & $2.4\times10^{-1}$ \\
        & & fRMSE low & $1.8\times10^{-2}$ & $3.1\times10^{-4}$ & $7.1\times10^{-2}$ \\
        & & fRMSE mid & $2.8\times10^{-3}$ & $6.7\times10^{-5}$ & $1.2\times10^{-4}$ \\
        & & fRMSE high & $8.3\times10^{-4}$ & $8.5\times10^{-6}$ & $4.4\times10^{-6}$ \\
        \cmidrule{2-6}
         & \multirow{8}{3.3cm}{$\nu=1.0$} 
          & RMSE & $2.7\times10^{-2}$ & $1.2\times10^{-3}$ & $1.1\times10^{-2}$ \\
        & & nRMSE & $2.4\times10^{-1}$ & $4.0\times10^{-3}$ & $1.9\times10^{-2}$ \\
        & & max error & $1.7\times10^{-1}$ & $8.0\times10^{-3}$ & $1.6\times10^{-2}$ \\
        & & cRMSE & $1.7\times10^{-2}$ & $1.1\times10^{-4}$ & $9.6\times10^{-3}$ \\
        & & bRMSE & $2.6\times10^{-2}$ & $1.2\times10^{-3}$ & $5.6\times10^{-3}$ \\
        & & fRMSE low & $1.0\times10^{-2}$ & $4.2\times10^{-4}$ & $3.2\times10^{-3}$ \\
        & & fRMSE mid & $1.8\times10^{-3}$ & $1.6\times10^{-5}$ & $6.0\times10^{-5}$ \\
        & & fRMSE high & $2.3\times10^{-4}$ & $1.5\times10^{-6}$ & $3.2\times10^{-6}$ \\
        \bottomrule
    \end{tabularx}
\end{table*}

\begin{table*}[h!]
    \caption{Summary of the baseline models' performance for different evaluation metrics: RMSE, normalised RMSE (nRMSE), RMSE from conserved value (cRMSE), maximum error, RMSE at the boundaries (bRMSE), RMSE in Fourier space at low (fRMSE low), medium (fRMSE mid), and high frequency (fRMSE high) ranges applied to the Darcy flow equation with different parameter values.}
    \label{tab:baseline_darcy}
    \centering
    \begin{tabularx}{\textwidth}{lXlrr}
        \toprule
        & & & \multicolumn{2}{c}{Baseline model} \\
        \cmidrule{4-5}
        PDE & Parameter & Metric & \multicolumn{1}{c}{U-Net} & \multicolumn{1}{c}{FNO} \\
        \midrule
        \multirow{40}{*}{DarcyFlow} & \multirow{8}{3.3cm}{$\beta=0.01$} & RMSE & $4.0\times10^{-3}$ & $8.0\times10^{-3}$\\
        & & nRMSE & $1.1\times10^{+0}$ & $2.5\times10^{+0}$\\
        & & max error & $6.8\times10^{-2}$ & $1.5\times10^{-1}$\\
        & & cRMSE & $5.8\times10^{-3}$ & $1.3\times10^{-2}$\\
        & & bRMSE & $6.3\times10^{-4}$ & $4.7\times10^{-3}$\\
        & & fRMSE low & $2.5\times10^{-3}$ & $5.2\times10^{-3}$\\
        & & fRMSE mid & $1.3\times10^{-4}$ & $1.5\times10^{-4}$\\
        & & fRMSE high & $2.1\times10^{-5}$ & $1.6\times10^{-5}$\\
        \cmidrule{2-5}
        & \multirow{8}{3.3cm}{$\beta=0.1$} & RMSE & $4.8\times10^{-3}$ & $6.2\times10^{-3}$\\
        & & nRMSE & $1.8\times10^{-1}$ & $2.2\times10^{-1}$\\
        & & max error & $7.0\times10^{-2}$ & $8.9\times10^{-2}$\\
        & & cRMSE & $6.0\times10^{-3}$ & $7.7\times10^{-3}$\\
        & & bRMSE & $8.6\times10^{-4}$ & $5.0\times10^{-3}$\\
        & & fRMSE low & $2.6\times10^{-3}$ & $3.6\times10^{-3}$\\
        & & fRMSE mid & $1.9\times10^{-4}$ & $2.6\times10^{-4}$\\
        & & fRMSE high & $4.4\times10^{-5}$ & $4.5\times10^{-5}$\\
        \cmidrule{2-5}
        & \multirow{8}{3.3cm}{$\beta=1.0$} & RMSE & $6.4\times10^{-3}$ & $1.2\times10^{-2}$\\
        & & nRMSE & $3.3\times10^{-2}$ & $6.4\times10^{-2}$\\
        & & max error & $9.0\times10^{-2}$ & $1.1\times10^{-1}$\\
        & & cRMSE & $6.0\times10^{-3}$ & $1.1\times10^{-2}$\\
        & & bRMSE & $3.5\times10^{-3}$ & $5.5\times10^{-3}$\\
        & & fRMSE low & $3.0\times10^{-3}$ & $5.2\times10^{-3}$\\
        & & fRMSE mid & $3.4\times10^{-4}$ & $5.1\times10^{-4}$\\
        & & fRMSE high & $1.3\times10^{-4}$ & $1.5\times10^{-4}$\\
        \cmidrule{2-5}
         & \multirow{8}{3.3cm}{$\beta=10.0$} & RMSE & $1.4\times10^{-2}$ & $2.1\times10^{-2}$\\
        & & nRMSE & $8.2\times10^{-3}$ & $1.2\times10^{-2}$\\
        & & max error & $2.4\times10^{-1}$ & $3.2\times10^{-1}$\\
        & & cRMSE & $9.9\times10^{-3}$ & $1.5\times10^{-2}$\\
        & & bRMSE & $9.4\times10^{-3}$ & $1.6\times10^{-2}$\\
        & & fRMSE low & $5.8\times10^{-3}$ & $8.3\times10^{-3}$\\
        & & fRMSE mid & $9.8\times10^{-4}$ & $1.3\times10^{-3}$\\
        & & fRMSE high & $3.6\times10^{-4}$ & $5.7\times10^{-4}$\\
        \cmidrule{2-5}
         & \multirow{8}{3.3cm}{$\beta=100.0$} & RMSE & $7.3\times10^{-2}$ & $1.1\times10^{-1}$\\
        & & nRMSE & $4.4\times10^{-3}$ & $6.4\times10^{-3}$\\
        & & max error & $1.7\times10^{+0}$ & $2.1\times10^{+0}$\\
        & & cRMSE & $5.1\times10^{-2}$ & $8.9\times10^{-2}$\\
        & & bRMSE & $4.6\times10^{-2}$ & $7.9\times10^{-2}$\\
        & & fRMSE low & $2.9\times10^{-2}$ & $4.6\times10^{-2}$\\
        & & fRMSE mid & $5.3\times10^{-3}$ & $7.6\times10^{-3}$\\
        & & fRMSE high & $2.5\times10^{-3}$ & $3.6\times10^{-3}$\\
        \bottomrule
    \end{tabularx}
\end{table*}

\begin{table*}[h!]
    \caption{Summary of the baseline models' performance for different evaluation metrics: RMSE, normalised RMSE (nRMSE), RMSE from conserved value (cRMSE), maximum error, RMSE at the boundaries (bRMSE), RMSE in Fourier space at low (fRMSE low), medium (fRMSE mid), and high frequency (fRMSE high) ranges applied to the 1d diffusion-reaction equation with different parameter values.}
    \label{tab:baseline_reacdiff}
    \centering
    \begin{tabularx}{\textwidth}{lp{3.3cm}lrrr}
        \toprule
        & & & \multicolumn{3}{c}{Baseline model} \\
        \cmidrule{4-6}
        PDE & Parameter & Metric & \multicolumn{1}{c}{U-Net} & \multicolumn{1}{c}{FNO} & \multicolumn{1}{c}{PINN} \\
        \midrule
        \multirow{32}{*}{ReacDiff} & \multirow{8}{3.3cm}{$\nu=0.5, \rho=1.0$} & RMSE & $3.1\times10^{-3}$ & $6.3\times10^{-4}$ & $4.5\times10^{-2}$ \\
        & & nRMSE & $6.0\times10^{-3}$ & $1.4\times10^{-3}$ & $8.0\times10^{-2}$ \\
        & & max error & $1.8\times10^{-2}$ & $8.7\times10^{-3}$ & $7.6\times10^{-2}$ \\
        & & cRMSE & $2.5\times10^{-3}$ & $1.3\times10^{-3}$ & $4.3\times10^{-2}$ \\
        & & bRMSE & $3.7\times10^{-3}$ & $6.7\times10^{-4}$ & $7.5\times10^{-2}$ \\
        & & fRMSE low & $1.1\times10^{-3}$ & $4.1\times10^{-4}$ & $1.4\times10^{-2}$ \\
        & & fRMSE mid & $1.8\times10^{-4}$ & $9.1\times10^{-6}$ & $2.4\times10^{-4}$ \\
        & & fRMSE high & $1.8\times10^{-5}$ & $1.7\times10^{-6}$ & $3.7\times10^{-6}$ \\
        \cmidrule{2-6}
        & \multirow{8}{3.3cm}{$\nu=0.5, \rho=10.0$} & RMSE & $6.2\times10^{-8}$ & $0.0\times10^{+0}$ & $1.4\times10^{-2}$ \\
        & & nRMSE & $6.5\times10^{-8}$ & $0.0\times10^{+0}$ & $1.4\times10^{-2}$ \\
        & & max error & $6.2\times10^{-8}$ & $0.0\times10^{+0}$ & $2.6\times10^{-2}$ \\
        & & cRMSE & $6.2\times10^{-8}$ & $0.0\times10^{+0}$ & $6.2\times10^{-3}$ \\
        & & bRMSE & $6.2\times10^{-8}$ & $0.0\times10^{+0}$ & $2.3\times10^{-2}$ \\
        & & fRMSE low & $1.6\times10^{-8}$ & $0.0\times10^{+0}$ & $4.3\times10^{-3}$ \\
        & & fRMSE mid & $0.0\times10^{+0}$ & $0.0\times10^{+0}$ & $2.5\times10^{-4}$ \\
        & & fRMSE high & $0.0\times10^{+0}$ & $0.0\times10^{+0}$ & $2.9\times10^{-6}$ \\
        \cmidrule{2-6}
        & \multirow{8}{3.3cm}{$\nu=2.0, \rho=1.0$} & RMSE & $2.3\times10^{-3}$ & $2.9\times10^{-4}$ & $3.9\times10^{-1}$ \\
        & & nRMSE & $4.5\times10^{-3}$ & $7.0\times10^{-4}$ & $7.3\times10^{-1}$ \\
        & & max error & $2.0\times10^{-2}$ & $4.2\times10^{-3}$ & $3.9\times10^{-1}$ \\
        & & cRMSE & $1.9\times10^{-3}$ & $6.4\times10^{-4}$ & $3.9\times10^{-1}$ \\
        & & bRMSE & $1.8\times10^{-3}$ & $4.1\times10^{-4}$ & $3.9\times10^{-1}$ \\
        & & fRMSE low & $7.7\times10^{-4}$ & $1.9\times10^{-4}$ & $9.7\times10^{-2}$ \\
        & & fRMSE mid & $1.7\times10^{-4}$ & $9.2\times10^{-6}$ & $6.2\times10^{-5}$ \\
        & & fRMSE high & $2.6\times10^{-5}$ & $1.8\times10^{-6}$ & $3.4\times10^{-6}$ \\
        \cmidrule{2-6}
         & \multirow{8}{3.3cm}{$\nu=2.0, \rho=10.0$} & RMSE & $3.1\times10^{-8}$ & $6.2\times10^{-8}$ & $3.2\times10^{-2}$ \\
        & & nRMSE & $3.2\times10^{-8}$ & $6.5\times10^{-8}$ & $3.3\times10^{-2}$ \\
        & & max error & $3.1\times10^{-8}$ & $6.2\times10^{-8}$ & $3.2\times10^{-2}$ \\
        & & cRMSE & $3.1\times10^{-8}$ & $6.2\times10^{-8}$ & $3.2\times10^{-2}$ \\
        & & bRMSE & $3.1\times10^{-8}$ & $6.2\times10^{-8}$ & $3.1\times10^{-2}$ \\
        & & fRMSE low & $7.8\times10^{-9}$ & $1.6\times10^{-8}$ & $8.0\times10^{-3}$ \\
        & & fRMSE mid & $0.0\times10^{+0}$ & $0.0\times10^{+0}$ & $6.4\times10^{-6}$ \\
        & & fRMSE high & $0.0\times10^{+0}$ & $0.0\times10^{+0}$ & $2.7\times10^{-7}$ \\
        \bottomrule
    \end{tabularx}
\end{table*}

\begin{table*}[h!]
    \caption{Summary of the baseline models' performance for different evaluation metrics: RMSE, normalised RMSE (nRMSE), RMSE from conserved value (cRMSE), maximum error, RMSE at the boundaries (bRMSE), RMSE in Fourier space at low (fRMSE low), medium (fRMSE mid), and high frequency (fRMSE high) ranges applied to the 1d compressible Navier-Stokes equation with different parameter values.}
    \label{tab:baseline_1dcfd}
    \centering
    \begin{tabularx}{\textwidth}{lXlrr}
        \toprule
        & & & \multicolumn{2}{c}{Baseline model} \\
        \cmidrule{4-5}
        PDE & Parameter & Metric & \multicolumn{1}{c}{U-Net} & \multicolumn{1}{c}{FNO} \\
        \midrule
        \multirow{40}{*}{1DCFD} & \multirow{8}{3.3cm}{$\eta = \zeta = 0.01$\ Rand\ periodic} & RMSE & $9.9\times10^{-1}$ & $2.7\times10^{-1}$\\
        & & nRMSE & $3.6\times10^{-1}$ & $9.5\times10^{-2}$\\
        & & max error & $7.8\times10^{+0}$ & $4.1\times10^{+0}$\\
        & & cRMSE & $3.6\times10^{-1}$ & $5.0\times10^{-2}$\\
        & & bRMSE & $1.0\times10^{+0}$ & $2.2\times10^{-1}$\\
        & & fRMSE low & $3.6\times10^{-1}$ & $7.3\times10^{-2}$\\
        & & fRMSE mid & $1.2\times10^{-1}$ & $5.5\times10^{-2}$\\
        & & fRMSE high & $9.2\times10^{-3}$ & $3.7\times10^{-3}$\\
        \cmidrule{2-5}
        & \multirow{8}{3.3cm}{$\eta=\zeta=0.1$\ Rand\ periodic} & RMSE & $6.6\times10^{-1}$ & $9.3\times10^{-2}$\\
        & & nRMSE & $7.2\times10^{-1}$ & $6.8\times10^{-2}$\\
        & & max error & $5.3\times10^{+0}$ & $1.5\times10^{+0}$\\
        & & cRMSE & $3.5\times10^{-1}$ & $2.7\times10^{-2}$\\
        & & bRMSE & $6.8\times10^{-1}$ & $7.6\times10^{-2}$\\
        & & fRMSE low & $2.5\times10^{-1}$ & $2.8\times10^{-2}$\\
        & & fRMSE mid & $5.7\times10^{-2}$ & $1.3\times10^{-2}$\\
        & & fRMSE high & $7.7\times10^{-3}$ & $2.0\times10^{-3}$\\
        \cmidrule{2-5}
        & \multirow{8}{3.3cm}{inviscid\ Rand\ periodic} & RMSE & $1.7\times10^{+1}$ & $4.7\times10^{-1}$\\
        & & nRMSE & $1.1\times10^{+0}$ & $1.2\times10^{-1}$\\
        & & max error & $2.0\times10^{+1}$ & $7.1\times10^{+0}$\\
        & & cRMSE & $1.7\times10^{+1}$ & $6.7\times10^{-2}$\\
        & & bRMSE & $1.6\times10^{+1}$ & $3.5\times10^{-1}$\\
        & & fRMSE low & $5.3\times10^{-1}$ & $4.5\times10^{+0}$\\
        & & fRMSE mid & $1.9\times10^{-1}$ & $1.6\times10^{-1}$\\
        & & fRMSE high & $2.1\times10^{-2}$ & $2.6\times10^{-3}$\\
        \cmidrule{2-5}
         & \multirow{8}{3.3cm}{inviscid\ Rand\ Outgoing} & RMSE & $1.6\times10^{+0}$ & $2.6\times10^{-1}$\\
        & & nRMSE & $1.1\times10^{+1}$ & $6.7\times10^{+0}$\\
        & & max error & $1.2\times10^{+1}$ & $4.3\times10^{+0}$\\
        & & cRMSE & $1.5\times10^{+0}$ & $1.5\times10^{-1}$\\
        & & bRMSE & $1.8\times10^{+0}$ & $3.6\times10^{-1}$\\
        & & fRMSE low & $6.8\times10^{-1}$ & $9.0\times10^{-2}$\\
        & & fRMSE mid & $1.2\times10^{-1}$ & $4.5\times10^{-2}$\\
        & & fRMSE high & $1.6\times10^{-2}$ & $6.7\times10^{-3}$\\
        \cmidrule{2-5}
         & \multirow{8}{3.3cm}{inviscid\ Shock\ Outgoing} & RMSE & $4.1\times10^{-1}$ & $1.6\times10^{-1}$\\
        & & nRMSE & $1.7\times10^{-1}$ & $4.7\times10^{-2}$\\
        & & max error & $6.6\times10^{+0}$ & $3.8\times10^{+0}$\\
        & & cRMSE & $2.1\times10^{-1}$ & $5.3\times10^{-2}$\\
        & & bRMSE & $5.6\times10^{-1}$ & $2.4\times10^{-1}$\\
        & & fRMSE low & $1.4\times10^{-1}$ & $3.7\times10^{-2}$\\
        & & fRMSE mid & $5.3\times10^{-2}$ & $2.6\times10^{-2}$\\
        & & fRMSE high & $1.1\times10^{-2}$ & $6.7\times10^{-3}$\\
        \bottomrule
    \end{tabularx}
\end{table*}

\begin{table*}[h!]
    \caption{Summary of the baseline models' performance for different evaluation metrics: RMSE, normalised RMSE (nRMSE), RMSE from conserved value (cRMSE), maximum error, RMSE at the boundaries (bRMSE), RMSE in Fourier space at low (fRMSE low), medium (fRMSE mid), and high frequency (fRMSE high) ranges applied to the 2d compressible Navier-Stokes equation with different parameter values (first part).}
    \label{tab:baseline_2dcfd_1}
    \centering
    \begin{tabularx}{\textwidth}{lXlrr}
        \toprule
        & & & \multicolumn{2}{c}{Baseline model} \\
        \cmidrule{4-5}
        PDE & Parameter & Metric & \multicolumn{1}{c}{U-Net} & \multicolumn{1}{c}{FNO} \\
        \midrule
        \multirow{32}{*}{2DCFD} & \multirow{8}{3.3cm}{$M=0.1$, inviscid\ Rand\ periodic} & RMSE & $4.0\times10^{-1}$ & $2.6\times10^{-1}$\\
        & & nRMSE & $6.6\times10^{-1}$ & $2.8\times10^{-1}$\\
        & & max error & $5.1\times10^{+0}$ & $4.2\times10^{+0}$\\
        & & cRMSE & $1.5\times10^{-1}$ & $1.6\times10^{-2}$\\
        & & bRMSE & $4.3\times10^{-1}$ & $2.6\times10^{-1}$\\
        & & fRMSE low & $1.1\times10^{-1}$ & $4.5\times10^{-2}$\\
        & & fRMSE mid & $4.8\times10^{-2}$ & $4.4\times10^{-2}$\\
        & & fRMSE high & $1.7\times10^{-2}$ & $1.6\times10^{-2}$\\
        \cmidrule{2-5}
        & \multirow{8}{3.3cm}{$M=0.1,\eta=\zeta=0.01$\ Rand\ periodic} & RMSE & $9.1\times10^{-2}$ & $2.3\times10^{-2}$\\
        & & nRMSE & $7.1\times10^{-1}$ & $1.7\times10^{-1}$\\
        & & max error & $1.1\times10^{+0}$ & $4.0\times10^{-1}$\\
        & & cRMSE & $3.6\times10^{-2}$ & $5.3\times10^{-3}$\\
        & & bRMSE & $1.1\times10^{-1}$ & $2.2\times10^{-2}$\\
        & & fRMSE low & $2.7\times10^{-2}$ & $5.7\times10^{-3}$\\
        & & fRMSE mid & $8.2\times10^{-3}$ & $2.7\times10^{-3}$\\
        & & fRMSE high & $2.6\times10^{-3}$ & $6.3\times10^{-4}$\\
        \cmidrule{2-5}
        & \multirow{8}{3.3cm}{$M=0.1,\eta=\zeta=0.1$\ Rand\ periodic} & RMSE & $4.7\times10^{-2}$ & $4.9\times10^{-3}$\\
        & & nRMSE & $5.1\times10^{+0}$ & $3.6\times10^{-1}$\\
        & & max error & $6.7\times10^{-1}$ & $8.7\times10^{-2}$\\
        & & cRMSE & $3.2\times10^{-2}$ & $3.2\times10^{-3}$\\
        & & bRMSE & $6.6\times10^{-2}$ & $4.3\times10^{-3}$\\
        & & fRMSE low & $1.3\times10^{-2}$ & $1.4\times10^{-3}$\\
        & & fRMSE mid & $4.2\times10^{-3}$ & $4.3\times10^{-4}$\\
        & & fRMSE high & $2.2\times10^{-3}$ & $1.4\times10^{-4}$\\
        \cmidrule{2-5}
         & \multirow{8}{3.3cm}{$M=1.0$, inviscid\ Rand\ periodic} & RMSE & $1.5\times10^{+0}$ & $1.4\times10^{+0}$\\
        & & nRMSE & $4.7\times10^{-1}$ & $3.5\times10^{-1}$\\
        & & max error & $1.6\times10^{+1}$ & $1.6\times10^{+1}$\\
        & & cRMSE & $4.8\times10^{-1}$ & $1.6\times10^{-1}$\\
        & & bRMSE & $1.5\times10^{+0}$ & $1.3\times10^{+0}$\\
        & & fRMSE low & $4.8\times10^{-1}$ & $4.0\times10^{-1}$\\
        & & fRMSE mid & $1.2\times10^{-1}$ & $1.2\times10^{-1}$\\
        & & fRMSE high & $3.9\times10^{-2}$ & $3.9\times10^{-2}$\\
        \bottomrule
    \end{tabularx}
\end{table*}

\begin{table*}[h!]
    \caption{Summary of the baseline models' performance for different evaluation metrics: RMSE, normalised RMSE (nRMSE), RMSE from conserved value (cRMSE), maximum error, RMSE at the boundaries (bRMSE), RMSE in Fourier space at low (fRMSE low), medium (fRMSE mid), and high frequency (fRMSE high) ranges applied to the 2d compressible Navier-Stokes equation with different parameter values (second part).}
    \label{tab:baseline_2dcfd_2}
    \centering
    \begin{tabularx}{\textwidth}{lXlrr}
        \toprule
        & & & \multicolumn{2}{c}{Baseline model} \\
        \cmidrule{4-5}
        PDE & Parameter & Metric & \multicolumn{1}{c}{U-Net} & \multicolumn{1}{c}{FNO} \\
        \midrule
        \multirow{32}{*}{2DCFD} & \multirow{8}{3.3cm}{$M=1.0,\eta=\zeta=0.01$\ Rand\ periodic} & RMSE & $3.4\times10^{-1}$ & $1.2\times10^{-1}$\\
        & & nRMSE & $3.6\times10^{-1}$ & $9.6\times10^{-2}$\\
        & & max error & $3.7\times10^{+0}$ & $1.7\times10^{+0}$\\
        & & cRMSE & $1.1\times10^{-1}$ & $1.8\times10^{-2}$\\
        & & bRMSE & $3.6\times10^{-1}$ & $1.3\times10^{-1}$\\
        & & fRMSE low & $1.1\times10^{-1}$ & $3.3\times10^{-2}$\\
        & & fRMSE mid & $2.7\times10^{-2}$ & $1.5\times10^{-2}$\\
        & & fRMSE high & $6.2\times10^{-3}$ & $3.6\times10^{-3}$\\
        \cmidrule{2-5}
        & \multirow{8}{3.3cm}{$M=1.0,\eta=\zeta=0.1$\ Rand\ periodic} & RMSE & $1.1\times10^{-1}$ & $1.5\times10^{-2}$\\
        & & nRMSE & $9.2\times10^{-1}$ & $9.8\times10^{-2}$\\
        & & max error & $1.3\times10^{+0}$ & $2.4\times10^{-1}$\\
        & & cRMSE & $4.8\times10^{-2}$ & $4.8\times10^{-3}$\\
        & & bRMSE & $1.5\times10^{-1}$ & $1.7\times10^{-2}$\\
        & & fRMSE low & $3.0\times10^{-2}$ & $3.2\times10^{-3}$\\
        & & fRMSE mid & $1.3\times10^{-2}$ & $1.5\times10^{-3}$\\
        & & fRMSE high & $4.3\times10^{-3}$ & $8.9\times10^{-4}$\\
        \cmidrule{2-5}
        & \multirow{8}{3.3cm}{$M=0.1$, inviscid\ Turb\ periodic} & RMSE & $3.3\times10^{-1}$ & $2.8\times10^{-1}$\\
        & & nRMSE & $1.9\times10^{-1}$ & $1.6\times10^{-1}$\\
        & & max error & $2.2\times10^{+0}$ & $1.8\times10^{+0}$\\
        & & cRMSE & $1.5\times10^{-2}$ & $1.2\times10^{-2}$\\
        & & bRMSE & $3.6\times10^{-1}$ & $2.8\times10^{-1}$\\
        & & fRMSE low & $6.5\times10^{-2}$ & $5.0\times10^{-2}$\\
        & & fRMSE mid & $3.2\times10^{-2}$ & $3.1\times10^{-2}$\\
        & & fRMSE high & $8.5\times10^{-3}$ & $6.5\times10^{-3}$\\
        \cmidrule{2-5}
         & \multirow{8}{3.3cm}{$M=1.0$, inviscid \ Turb\ periodic} & RMSE & $9.5\times10^{-2}$ & $9.2\times10^{-2}$\\
        & & nRMSE & $1.4\times10^{-1}$ & $1.3\times10^{-1}$\\
        & & max error & $8.2\times10^{-1}$ & $7.9\times10^{-1}$\\
        & & cRMSE & $6.5\times10^{-3}$ & $4.3\times10^{-3}$\\
        & & bRMSE & $1.1\times10^{-1}$ & $9.7\times10^{-1}$\\
        & & fRMSE low & $1.3\times10^{-2}$ & $1.1\times10^{-2}$\\
        & & fRMSE mid & $1.2\times10^{-2}$ & $1.2\times10^{-2}$\\
        & & fRMSE high & $5.2\times10^{-3}$ & $5.2\times10^{-3}$\\
        \bottomrule
    \end{tabularx}
\end{table*}

\begin{table*}[h!]
    \caption{Summary of the baseline models' performance for different evaluation metrics: RMSE, normalised RMSE (nRMSE), RMSE from conserved value (cRMSE), maximum error, RMSE at the boundaries (bRMSE), RMSE in Fourier space at low (fRMSE low), medium (fRMSE mid), and high frequency (fRMSE high) ranges applied to the 3d compressible Navier-Stokes equation with different parameter values.}
    \label{tab:baseline_3dcfd}
    \centering
    \begin{tabularx}{\textwidth}{lXlrr}
        \toprule
        & & & \multicolumn{2}{c}{Baseline model} \\
        \cmidrule{4-5}
        PDE & Parameter & Metric & \multicolumn{1}{c}{U-Net} & \multicolumn{1}{c}{FNO} \\
        \midrule
        \multirow{16}{*}{3DCFD} & \multirow{8}{3.3cm}{$M=1.0$ \ inviscid \ Rand\ periodic} & RMSE & $2.2\times10^{+0}$ & $6.0\times10^{-1}$\\
        & & nRMSE & $1.0\times10^{+0}$ & $3.7\times10^{-1}$\\
        & & max error & $9.0\times10^{+0}$ & $3.6\times10^{+0}$\\
        & & cRMSE & $2.3\times10^{+0}$ & $8.1\times10^{-2}$\\
        & & bRMSE & $2.1\times10^{+0}$ & $6.0\times10^{-1}$\\
        & & fRMSE low & $7.3\times10^{-1}$ & $1.1\times10^{-1}$\\
        & & fRMSE mid & $7.6\times10^{-2}$ & $4.4\times10^{-2}$\\
        & & fRMSE high & $2.3\times10^{-2}$ & $9.3\times10^{-3}$\\
        \cmidrule{2-5}
        & \multirow{8}{3.3cm}{$M=1.0$ \ inviscid \ Turb\ periodic} & RMSE & $8.1\times10^{-2}$ & $8.2\times10^{-2}$\\
        & & nRMSE & $2.3\times10^{-1}$ & $2.4\times10^{-1}$\\
        & & max error & $5.0\times10^{-1}$ & $4.5\times10^{-1}$\\
        & & cRMSE & $7.3\times10^{-3}$ & $2.8\times10^{-3}$\\
        & & bRMSE & $9.9\times10^{-2}$ & $8.6\times10^{-2}$\\
        & & fRMSE low & $1.1\times10^{-2}$ & $7.2\times10^{-3}$\\
        & & fRMSE mid & $8.0\times10^{-3}$ & $9.4\times10^{-3}$\\
        & & fRMSE high & $1.7\times10^{-3}$ & $4.5\times10^{-3}$\\
        \bottomrule
    \end{tabularx}
\end{table*}

\FloatBarrier

\section{Detailed Runtime Comparison}
\label{sec:ap:runtime}
In this section we present the detailed comparison of computation time between the PDE solver used to generate the data and the baseline models used in this work, summarized in \autoref{tab:runtime_comparison}.
The system listed in \autoref{tab:hw1} was used to run all timing measurements regarding the Diffusion-sorption, 2D diffusion-reaction and Shallow-water equation scenarios.
PyClaw~\citep{pyclaw}, a well-optimized finite-volume Fortran code, is used as PDE solver for the shallow-water equation data generation.
Note that the experiment is only running on a single core due to its small size.
Because the PINN model is not discretized, the inference time includes evaluating the trained model at the same discretization points of the reference simulation for the last $20$ time steps of the data. E.g. the 2D diffusion-reaction scenario is evaluated at $128^2 \times 20$ discrete points. Additionally, autoregression is not required and therefore, it leads to significantly faster computation time relative to FNO and U-Net.

\begin{table}[h!]
    \caption{System configuration 1}
    \label{tab:hw1}
    \centering
    \begin{tabularx}{0.6\textwidth}{c|X}
        \toprule
        CPU & 2 $\times$ AMD EPYC 7742 \\
        GPU & 1 $\times$ NVIDIA Volta V100 \\
        Software & PyTorch@1.11, CUDA@11.3 \\
        \bottomrule
    \end{tabularx}
\end{table}

\begin{table*}[h!]
    \caption{Comparison of computation time between the PDE solver used to generate a single data sample and single forward runs of FNO, U-Net, and PINN. Training time of the baseline models for one epoch are also presented in this table. The unit used for the time is seconds.}
    \label{tab:runtime_comparison}
    \centering
    \begin{tabularx}{\textwidth}{Xrlrrr}
        \toprule
        PDE & Resolution & Model & Training time ($\frac{\text{s}}{\text{epoch}}$) & Epochs & Inference time (s) \\
        \midrule
        \multirow{4}{*}{\parbox{2cm}{Diffusion-sorption}} & \multirow{4}{*}{$1\,024^1$} & PDE solver & -- & -- & $59.83$\\
        & & FNO & $97.52$ & 500 & $0.32$ \\
        & & U-Net & $96.75$ & 500 & $0.32$ \\
        & & PINN & $0.011$ & 15\,000 & $0.0027$ \\
        \midrule
        \multirow{4}{*}{\parbox{2cm}{2D diffusion-reaction}} & \multirow{4}{*}{$128^2$} & PDE solver & -- & -- & $2.21$\\
        & & FNO & $108.28$ & 500 & $0.40$ \\
        & & U-Net & $83.19$ & 500 & $0.61$ \\
        & & PINN & $0.022$ & 100 & $0.0077$ \\
        \midrule
        \multirow{4}{*}{\parbox{2.5cm}{Shallow-water equation}} & \multirow{4}{*}{$128^2$} & PDE solver & -- & -- & $0.62$ \\
        & & FNO & $105.16$ & 500 & $0.37$ \\
        & & U-Net & $83.32$ & 500 & $0.56$ \\
        & & PINN & $0.041$ & 15\,000 & $0.00673$ \\
        \bottomrule
    \end{tabularx}
\end{table*}

As the case for 3D data, 
we also performed a similar experiment whose results are summarized in \autoref{tab:runtime_comparison-3D}. 
The used system information is listed in \autoref{tab:hw2}. 
Because of the severe memory usage, 
the resolution was reduced to $64^3$, though we provided a data with resolution $128^3$ in our official dataset. 
Note that the training and inference time are shorter than the 2D cases in \autoref{tab:runtime_comparison}. This is because the number of time-step and sample numbers are less than the 2D cases to reduce dataset size. 

\begin{table}[h!]
    \caption{System configuration 2}
    \label{tab:hw2}
    \centering
    \begin{tabularx}{0.7\textwidth}{c|X}
        \toprule
        GPU & 1 $\times$ NVIDIA GeForce RTX 3090 \\
        Software (ML methods) & PyTorch@1.11, CUDA@11.3 \\
        Software (simulations) & JAX@0.2.26, CUDA@11.3 \\        
        \bottomrule
    \end{tabularx}
\end{table}

\begin{table*}[h!]
    \caption{Comparison of computation time between the PDE solver used to generate a single data sample and single forward runs of FNO, U-Net. Training time of the baseline models for one epoch are also presented in this table. The unit used for the time is seconds.}
    \label{tab:runtime_comparison-3D}
    \centering
    \begin{tabularx}{\textwidth}{Xrlrrr}
        \toprule
        PDE & Resolution & Model & Training time ($\frac{\text{s}}{\text{epoch}}$) & Epochs & Inference time (s) \\
        \midrule
        \multirow{4}{*}{\parbox{2cm}{3D CFD}} & \multirow{4}{*}{$64^3$} & PDE solver & -- & -- & $60.07$\\
        & & FNO & $24.77$ & 500 & $0.14$ \\
        & & U-Net & $62.22$ & 500 & $0.27$ \\
        %\midrule
        % & \multirow{4}{*}{$128^3$} & PDE solver & -- & -- & $380.53$\\
        %& & FNO & -- & 500 & $1.01$ \\
        %& & U-Net & -- & 500 & -- \\
       \bottomrule
    \end{tabularx}
\end{table*}

\section{Resolution Sensitivity of Inference Time}
\label{sec:time_sensitivity}

\begin{figure}[h!]
   \centering
    \includegraphics[width=\textwidth]{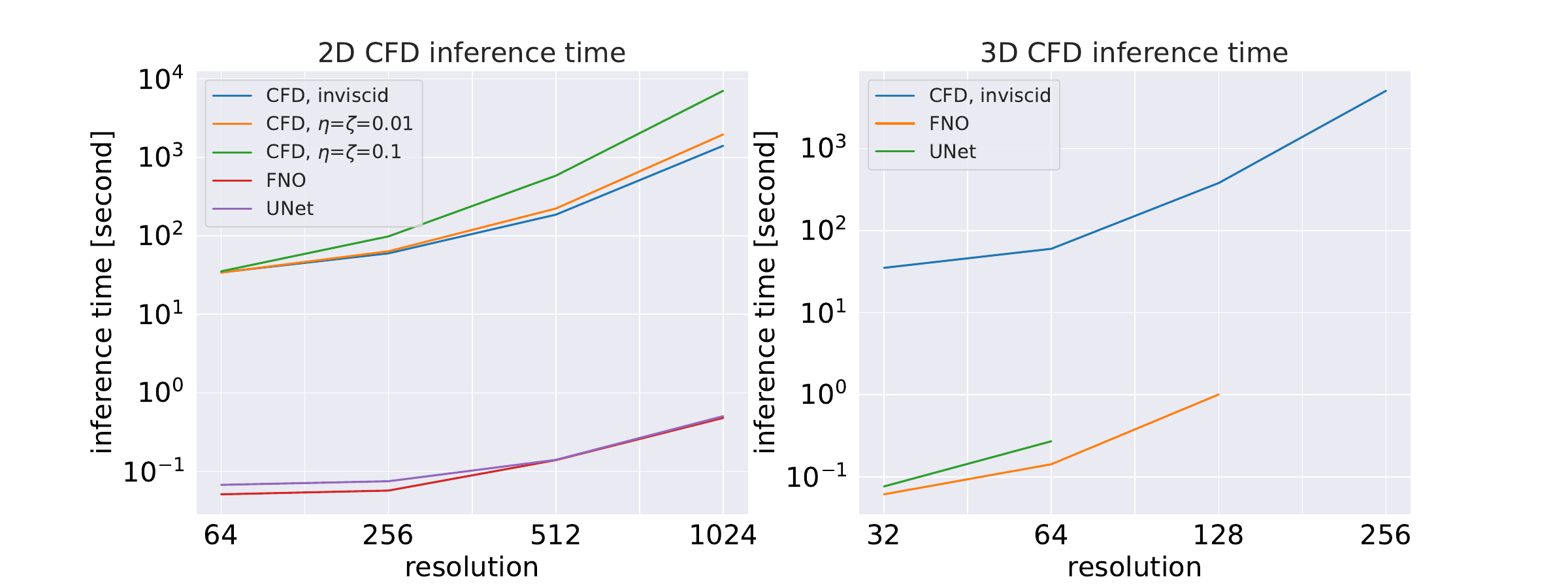}
    \caption{Plots inference time for 2D/3D CFD cases.}
    \label{fig:2D_CFD_inference_time}
\end{figure}

\autoref{fig:2D_CFD_inference_time} plots the resolution dependence of the inference time of classical simulation and ML methods for 2D/3D compressible Navier-Stokes equations cases. 
To calculate the inference times, we used the same hardware resources to be a "fair" comparison as listed in \autoref{tab:hw2}.  

The figure clearly shows that the ML inference time is nearly 3-order of magnitude smaller than that of the classical simulations. 
Concerning the resolution dependence, both of the ML models show a similar dependence to the inviscid classical  simulation method. 
Importantly, the inference time of ML models is in general independent of the diffusion coefficient, such as viscosity. On the other hand, 
the classical simulation methods increase their computation time with diffusion coefficient because of the stability condition, known as Courant-Friedrich-Lewy (CFL) condition, $\Delta t \propto \Delta x^2/\eta$ in the case of the explicit method. Here $\Delta x, \Delta t$ are time-step size and mesh size, respectively, and $\eta$ is the diffusion coefficient. 
This is much severer restriction than the inviscid case whose CFL condition is $\Delta t \propto \Delta x$. 
Hence, we can conclude that ML methods could even be suitable for solving for the problem with including strong-diffusive regime.  

\section{Error Comparison with PDE Solver}

To further assess the benefit of the trained baseline models, we generated the 2D diffusion-reaction data using a PDE solver with higher resolution ($512 \times 512$), and downsampled them to lower resolution ($128 \times 128$). These downsampled data were assumed as the ground truth (low discretization error) and then were used to train the baseline models. The trained baseline model predictions were compared against data that were generated using the same PDE solver but with coarser resolution (higher discretization error). The error comparison is summarized in \autoref{tab:resolution_error}. We observed that generating the data with lower resolution already accumulates high discretization error, relative to the baseline model prediction error. However, further sensitivity analysis with regards to different resolutions is required in future works to determine if the resolution is fine enough to be assumed as the ground truth.

\begin{table*}[h!]
    \caption{Error comparison between of U-Net, FNO, and PINN prediction, as well as low-resolution PDE solver data, against the high-resolution PDE solver data (assumed as the ground truth) for the 2D diffusion-reaction scenario.}
    \label{tab:resolution_error}
    \centering
    \begin{tabularx}{0.85\textwidth}{Xrrrr}
        \toprule
        Error metric & U-Net & FNO & PINN & low-res PDE solver \\
        \midrule
        RMSE & $6.1 \times 10^{-2}$ & $8.1 \times 10^{-3}$ & $1.9 \times 10^{-1}$ & $1.8 \times 10^{-1}$ \\
        nRMSE & $8.4 \times 10^{-1}$ & $1.2 \times 10^{-1}$ & $1.6 \times 10^{+0}$ & $2.8 \times 10^{+0}$ \\
        max error & $1.9 \times 10^{-1}$ & $9.1 \times 10^{-2}$ & $5.0 \times 10^{-1}$ & $8.9 \times 10^{-1}$ \\
        cRMSE & $3.9 \times 10^{-2}$ & $1.7 \times 10^{-3}$ & $1.3 \times 10^{-1}$ & $4.9 \times 10^{-2}$ \\
        bRMSE & $7.8 \times 10^{-2}$ & $2.7 \times 10^{-2}$ & $2.2 \times 10^{-1}$ & $2.1 \times 10^{-1}$ \\
        fRMSE low & $1.7 \times 10^{-2}$ & $8.2 \times 10^{-4}$ & $5.7 \times 10^{-2}$ & $4.9 \times 10^{-2}$ \\
        fRMSE mid & $5.4 \times 10^{-3}$ & $7.7 \times 10^{-4}$ & $1.3 \times 10^{-2}$ & $2.2 \times 10^{-2}$ \\
        fRMSE high & $6.8 \times 10^{-4}$ & $4.1 \times 10^{-4}$ & $1.5 \times 10^{-3}$ & $3.4 \times 10^{-3}$ \\
        \bottomrule
    \end{tabularx}
\end{table*}

\section{Visualization of Model Predictions}
\begin{figure}[!]
    \centering
        \begin{subfigure}{0.32\textwidth}
         \centering
         \includegraphics[width=\textwidth]{figures/timothy/diff-sorp/1D_diff-sorp_NA_NA_data.pdf}
         \caption{}%Plots of the RMSE calculated at different unrolled time steps.}
    %\label{fig:vis_diff-react_data}
    \end{subfigure}
    \hfill
    \begin{subfigure}{0.32\textwidth}
         \centering
         \includegraphics[width=\textwidth]{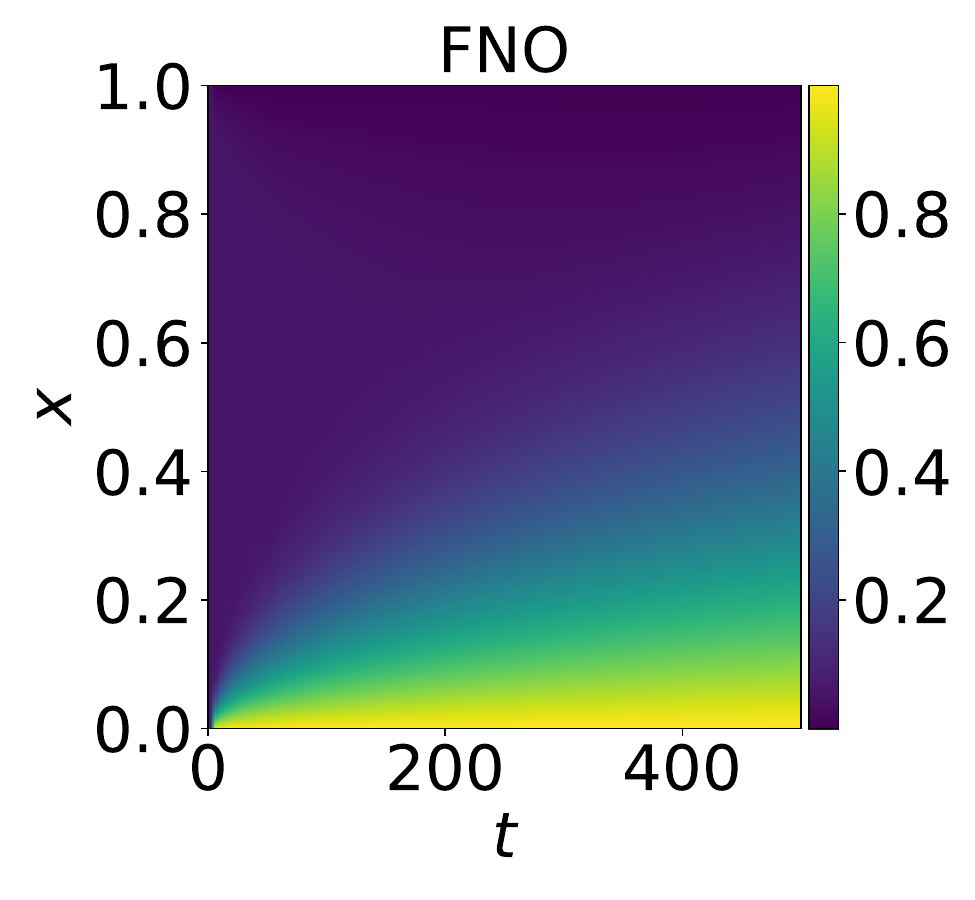}
         \caption{}
         %\label{fig:vis_diff-react_fno}
     \end{subfigure}
     \hfill
     \begin{subfigure}{0.32\textwidth}
         \centering
         \includegraphics[width=\textwidth]{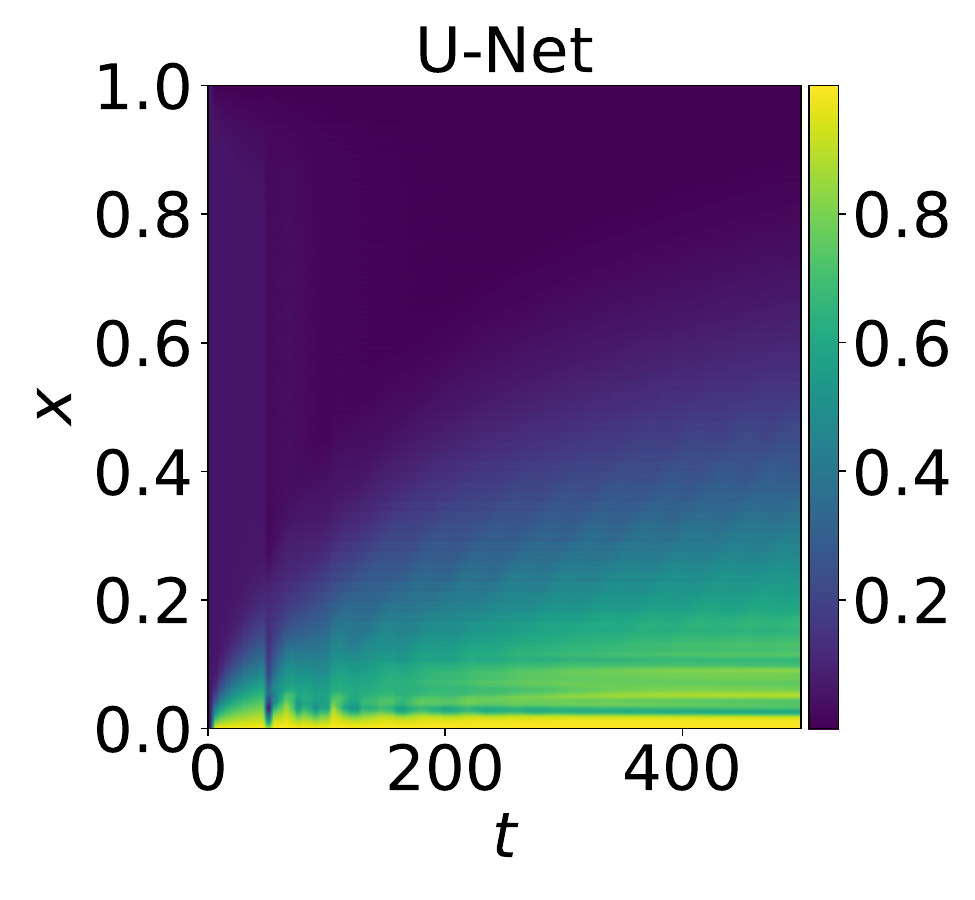}
         \caption{}
         %\label{fig:vis_diff-react_fno}
     \end{subfigure}
     \vspace{-2em} % REMOVING SUBFIGURE LABELS
     \caption{Visualization of the diffusion-sorption equation (a) data, (b) FNO prediction, and (c) U-Net prediction.}
    \label{fig:vis_diff-sorp_data-fno-unet}
\end{figure}

\begin{figure}[!]
    \centering
        \begin{subfigure}{0.24\textwidth}
         \centering
         \includegraphics[width=\textwidth]{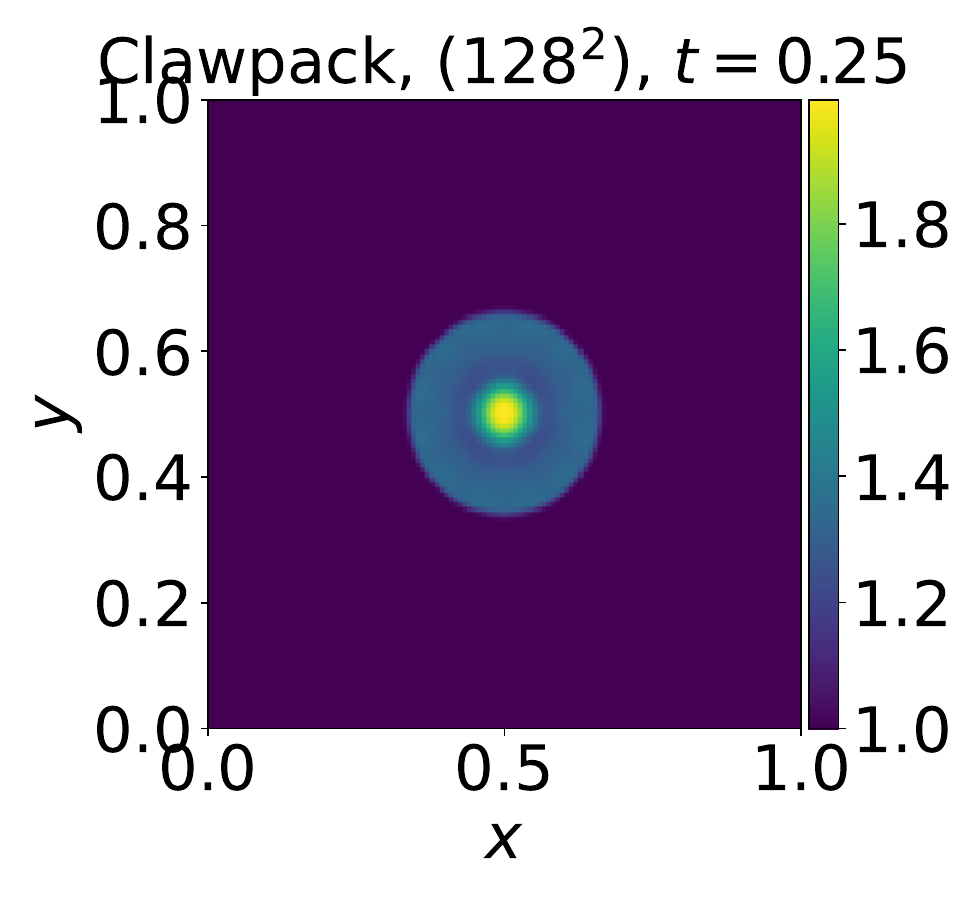}
         \caption{}%Plots of the RMSE calculated at different unrolled time steps.}
    %\label{fig:vis_diff-react_data}
    \end{subfigure}
    \hfill
    \begin{subfigure}{0.24\textwidth}
         \centering
         \includegraphics[width=\textwidth]{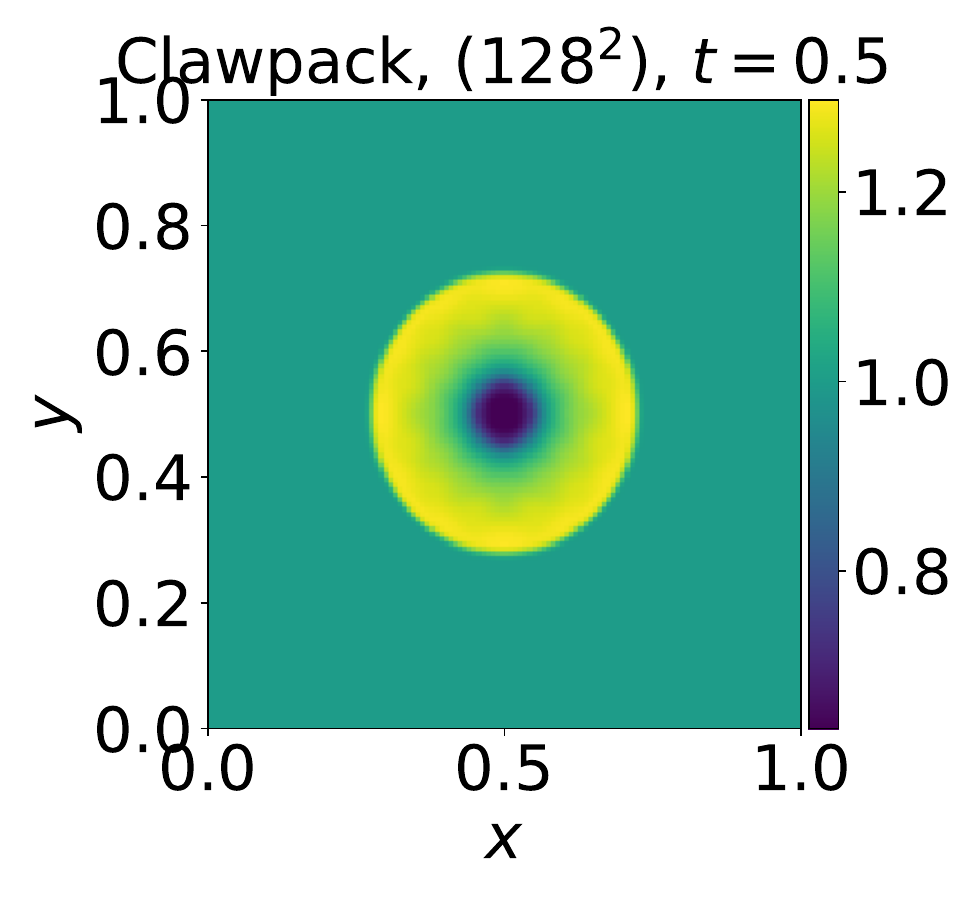}
         \caption{}
         %\label{fig:vis_diff-react_fno}
     \end{subfigure}
     \hfill
     \begin{subfigure}{0.24\textwidth}
         \centering
         \includegraphics[width=\textwidth]{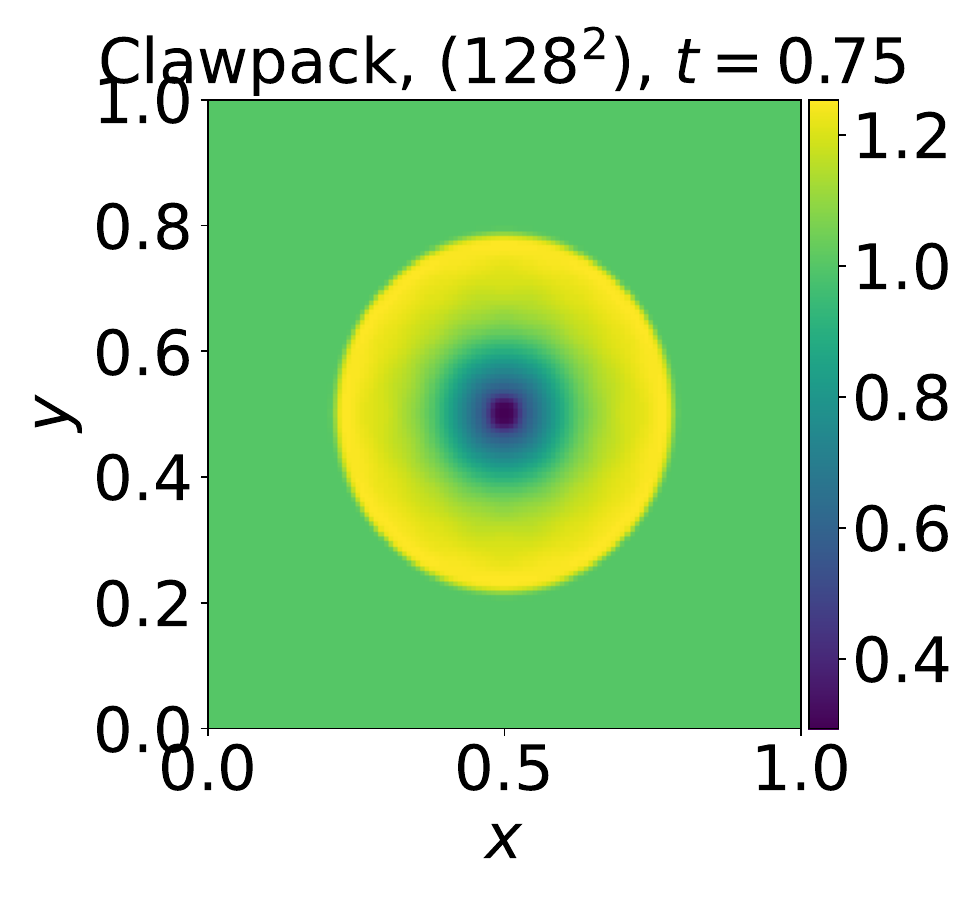}
         \caption{}
         %\label{fig:vis_diff-react_fno}
     \end{subfigure}
     \hfill
     \begin{subfigure}{0.24\textwidth}
         \centering
         \includegraphics[width=\textwidth]{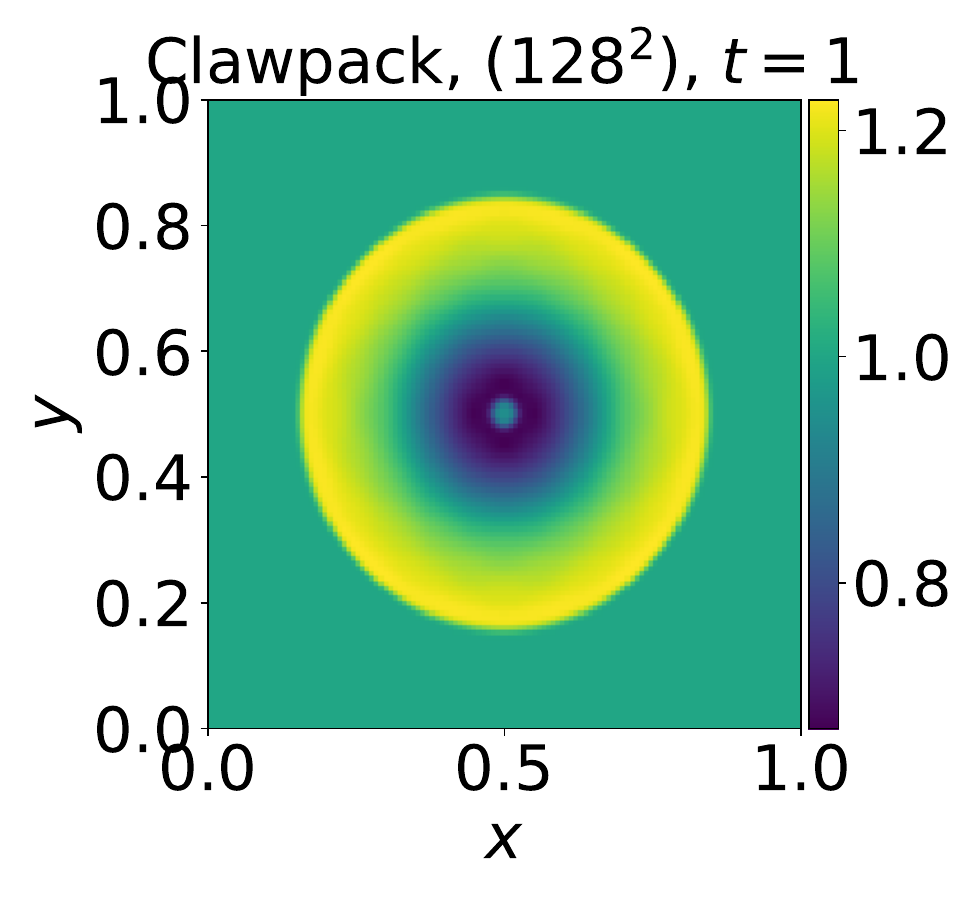}
         \caption{}
         %\label{fig:vis_diff-react_fno}
     \end{subfigure}
     \vspace{-2em} % REMOVING SUBFIGURE LABELS
     \caption{Visualization of the time evolution of the shallow water equation data.}
    \label{fig:vis_rdb_data}
\end{figure}

\begin{figure}[!]
    \centering
        \begin{subfigure}{0.24\textwidth}
         \centering
         \includegraphics[width=\textwidth]{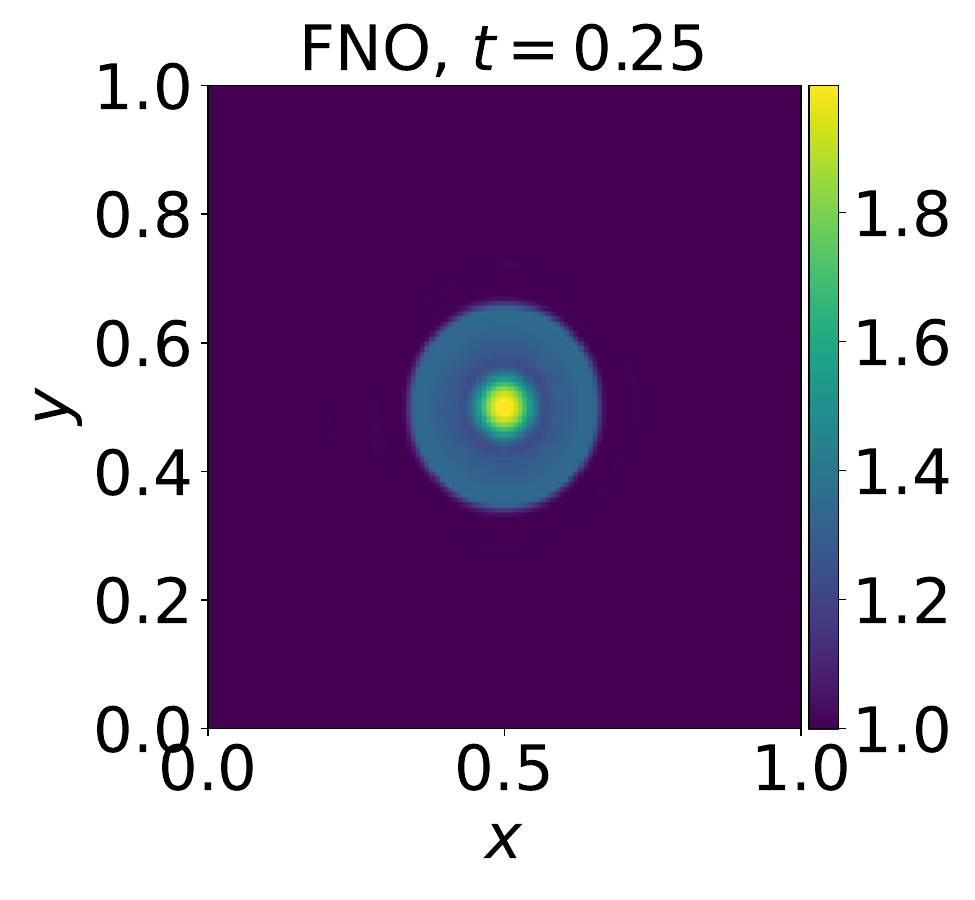}
         \caption{}%Plots of the RMSE calculated at different unrolled time steps.}
    %\label{fig:vis_diff-react_data}
    \end{subfigure}
    \hfill
    \begin{subfigure}{0.24\textwidth}
         \centering
         \includegraphics[width=\textwidth]{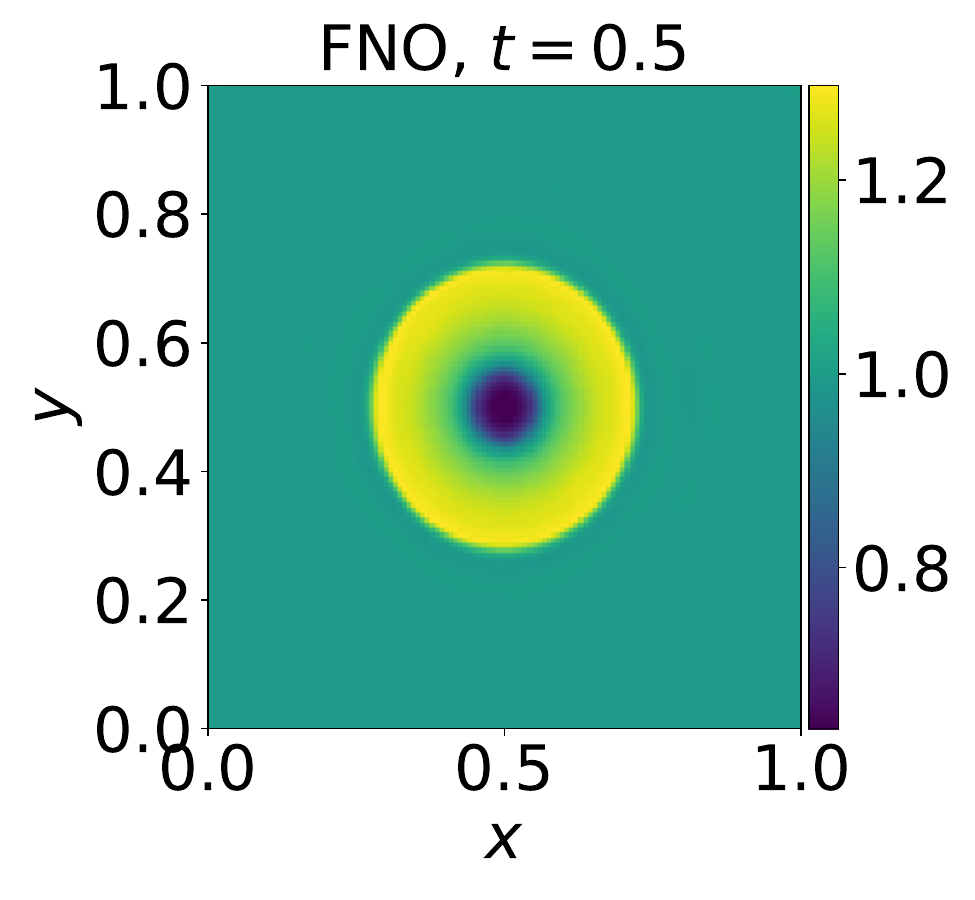}
         \caption{}
         %\label{fig:vis_diff-react_fno}
     \end{subfigure}
     \hfill
     \begin{subfigure}{0.24\textwidth}
         \centering
         \includegraphics[width=\textwidth]{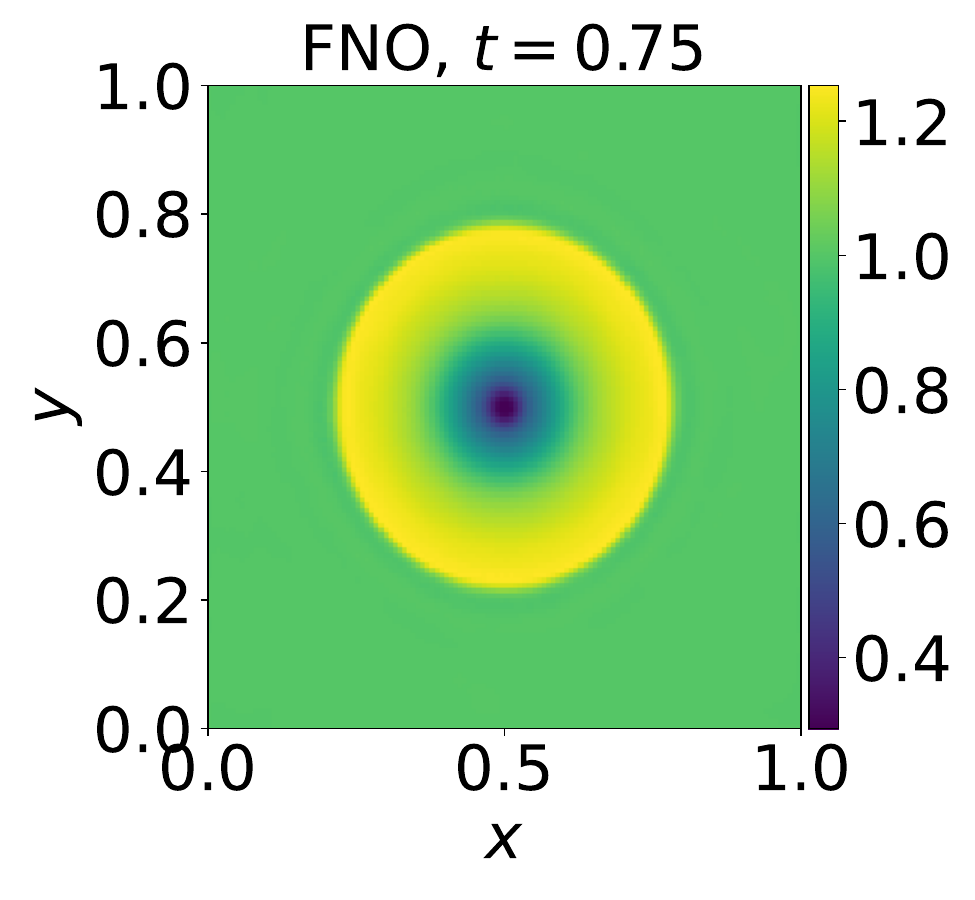}
         \caption{}
         %\label{fig:vis_diff-react_fno}
     \end{subfigure}
     \hfill
     \begin{subfigure}{0.24\textwidth}
         \centering
         \includegraphics[width=\textwidth]{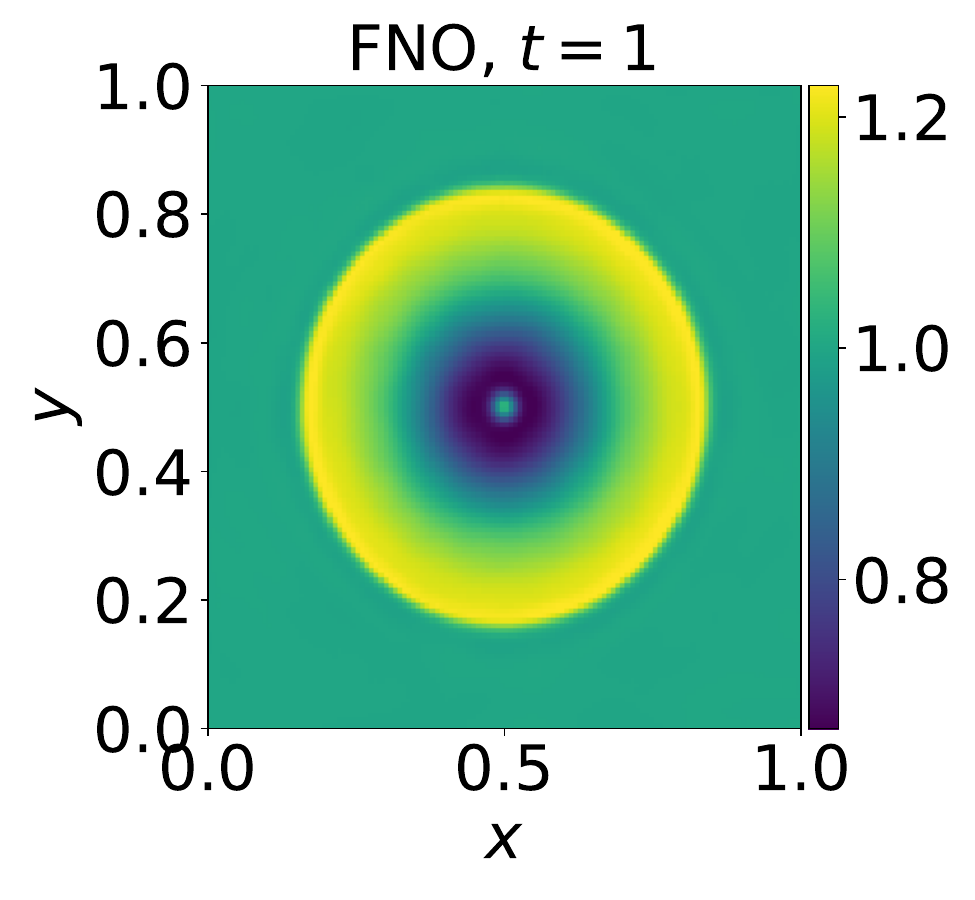}
         \caption{}
         %\label{fig:vis_diff-react_fno}
     \end{subfigure}
     \vspace{-2em} % REMOVING SUBFIGURE LABELS
     \caption{Visualization of the time evolution of the shallow water equation predicted using FNO.}
    \label{fig:vis_rdb_fno}
\end{figure}

\begin{figure}[!]
    \centering
        \begin{subfigure}{0.24\textwidth}
         \centering
         \includegraphics[width=\textwidth]{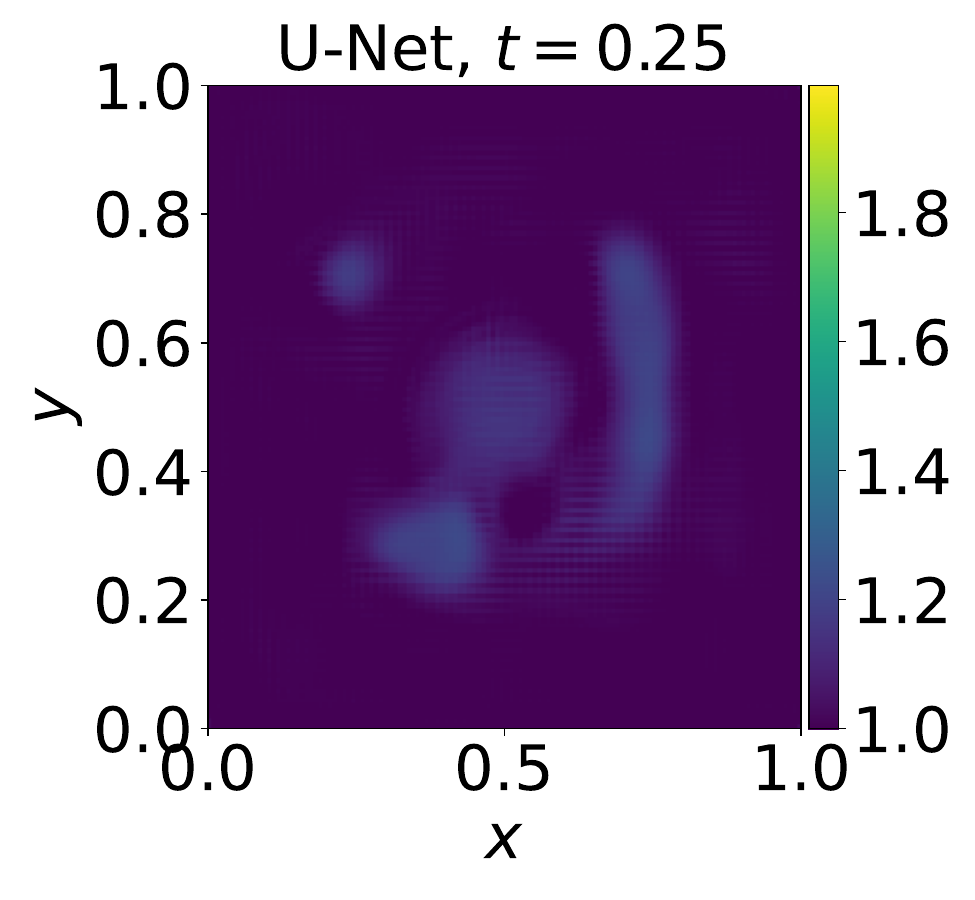}
         \caption{}%Plots of the RMSE calculated at different unrolled time steps.}
    %\label{fig:vis_diff-react_data}
    \end{subfigure}
    \hfill
    \begin{subfigure}{0.24\textwidth}
         \centering
         \includegraphics[width=\textwidth]{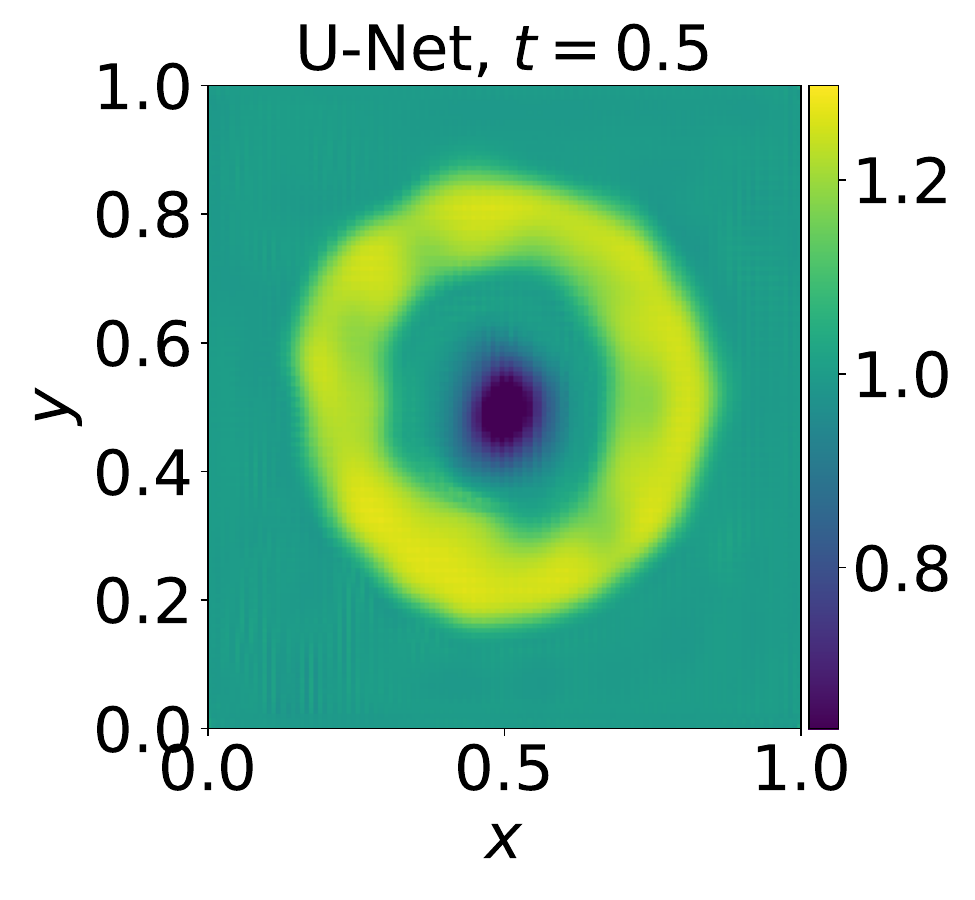}
         \caption{}
         %\label{fig:vis_diff-react_fno}
     \end{subfigure}
     \hfill
     \begin{subfigure}{0.24\textwidth}
         \centering
         \includegraphics[width=\textwidth]{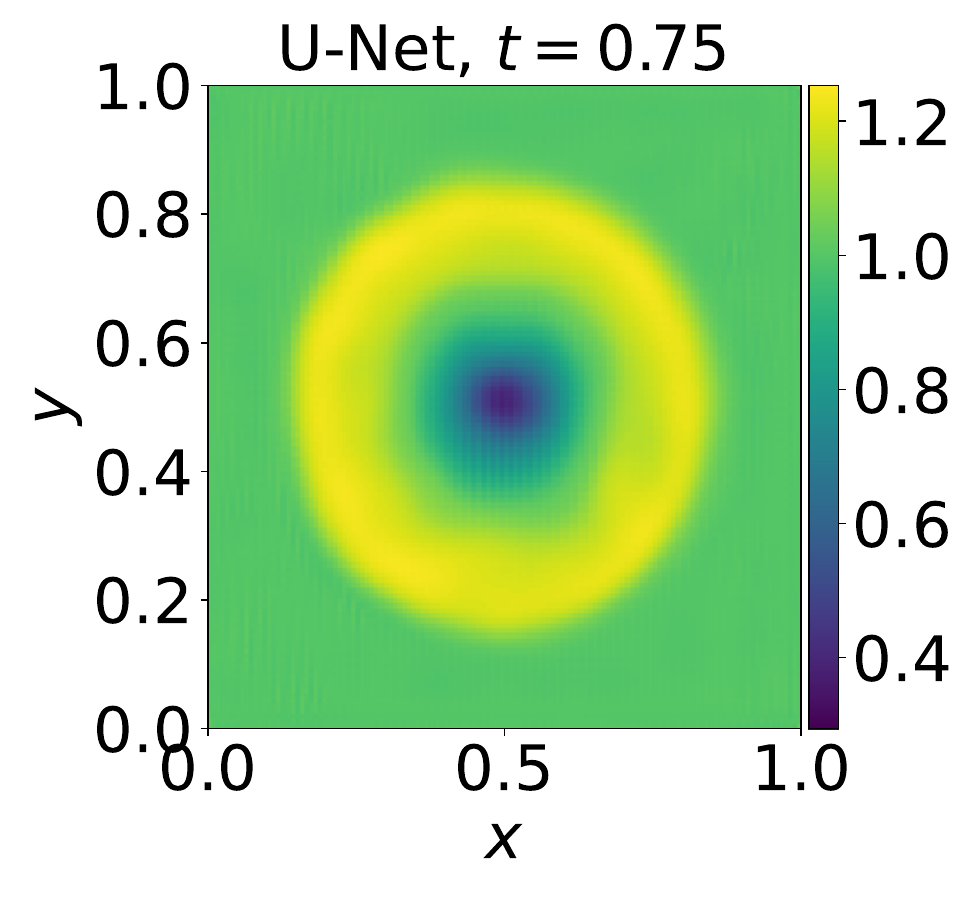}
         \caption{}
         %\label{fig:vis_diff-react_fno}
     \end{subfigure}
     \hfill
     \begin{subfigure}{0.24\textwidth}
         \centering
         \includegraphics[width=\textwidth]{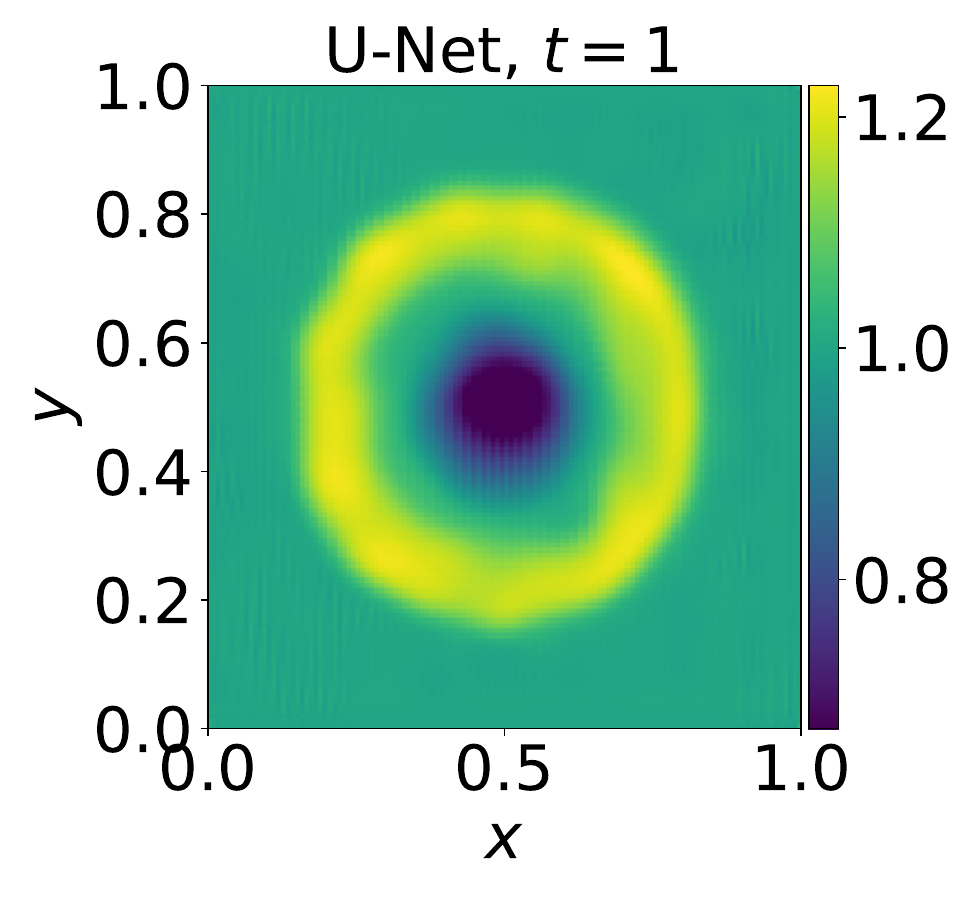}
         \caption{}
         %\label{fig:vis_diff-react_fno}
     \end{subfigure}
     \vspace{-2em} % REMOVING SUBFIGURE LABELS
     \caption{Visualization of the time evolution of the shallow water equation predicted using U-Net.}
    \label{fig:vis_rdb_unet}
\end{figure}

\begin{figure}[!]
    \centering
    \includegraphics[width=\textwidth]{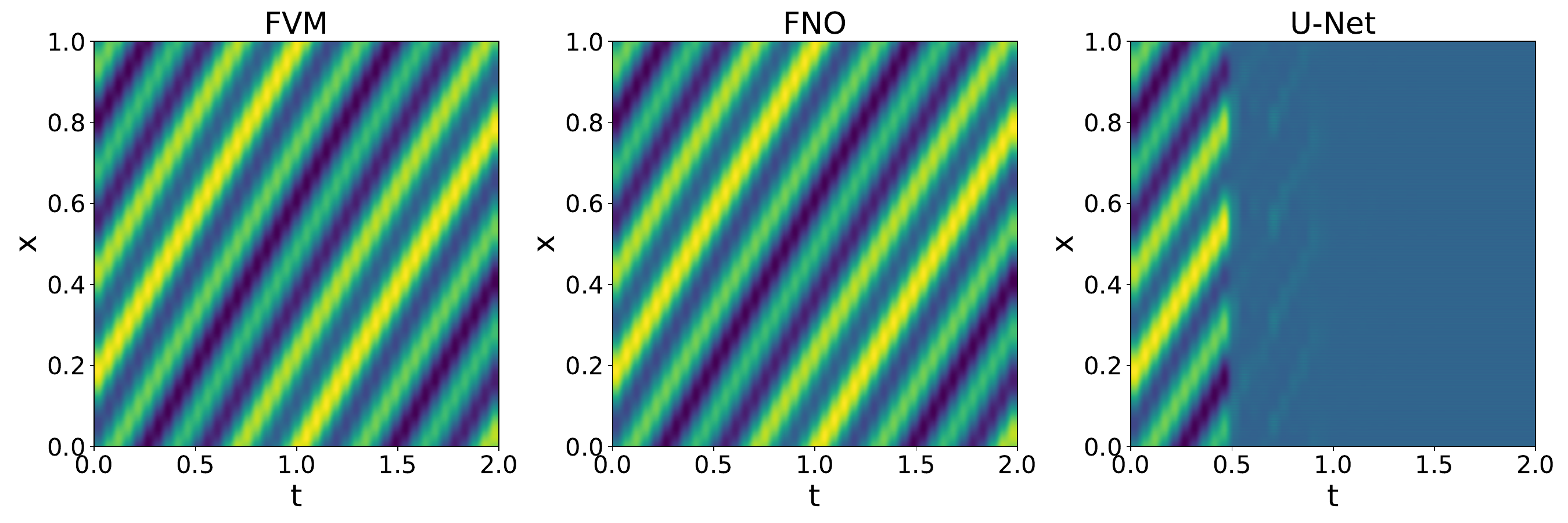}
    \caption{Plots of the predictions for 1D Advection equation.}
    \label{fig:vis_1d_adv_pred}
\end{figure}

\begin{figure}[!]
    \centering
    \includegraphics[width=\textwidth]{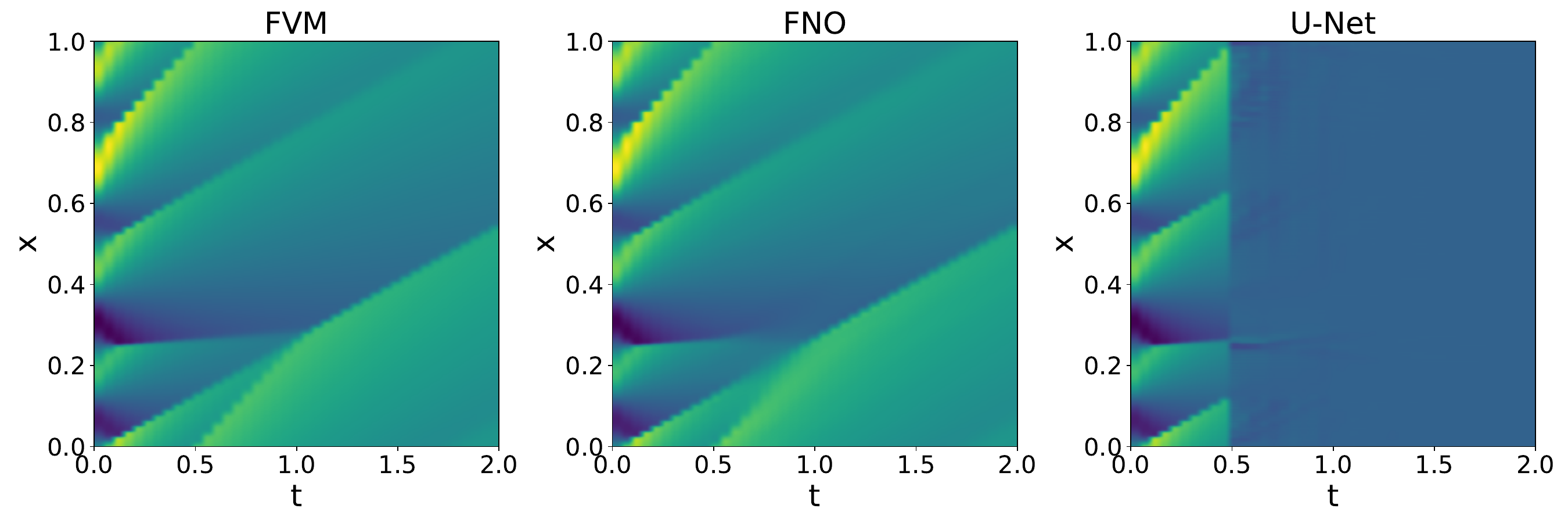}
    \caption{Plots of the predictions for 1D Burgers equation.}
    \label{fig:vis_1d_bgs_pred}
\end{figure}

\begin{figure}[!]
    \centering
    \includegraphics[width=\textwidth]{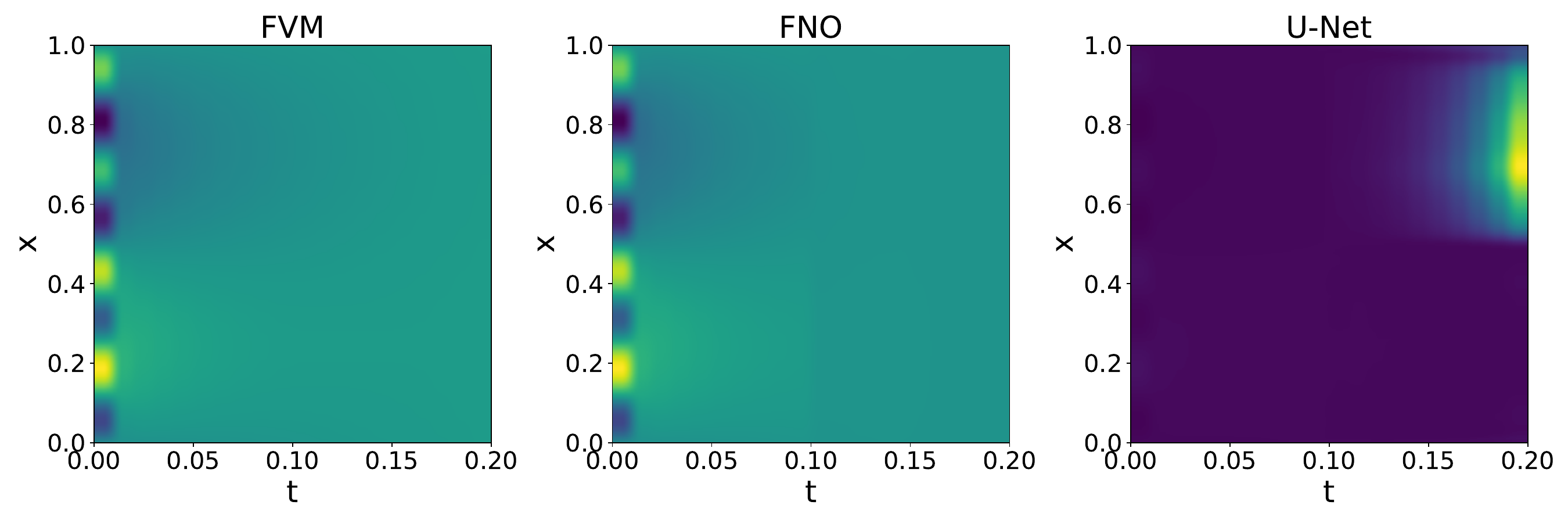}
    \caption{Plots of the predictions for 1D Reaction-Diffusion equation.}
    \label{fig:vis_1d_reac-diff_pred}
\end{figure}

\begin{figure}[!]
    \centering
    \includegraphics[width=\textwidth]{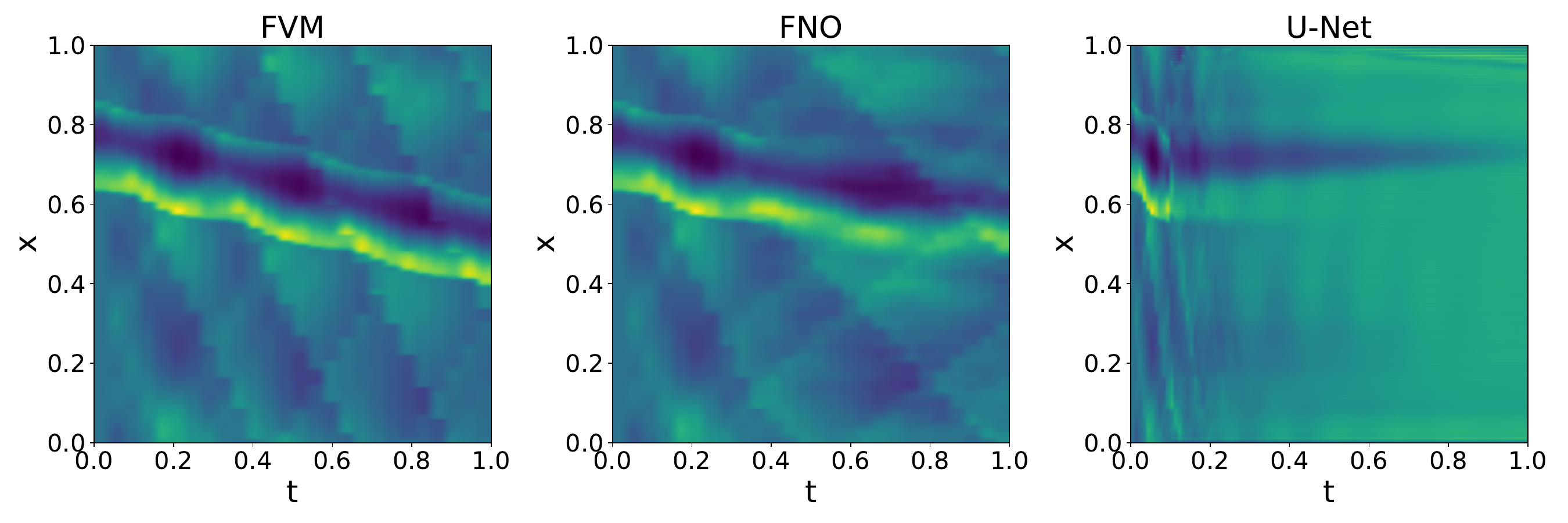}
    \caption{Plots of the predictions of density for 1D compressible NS equations.}
    \label{fig:vis_1d_cfd_pred}
\end{figure}

\begin{figure}[!]
    \centering
    \includegraphics[width=\textwidth]{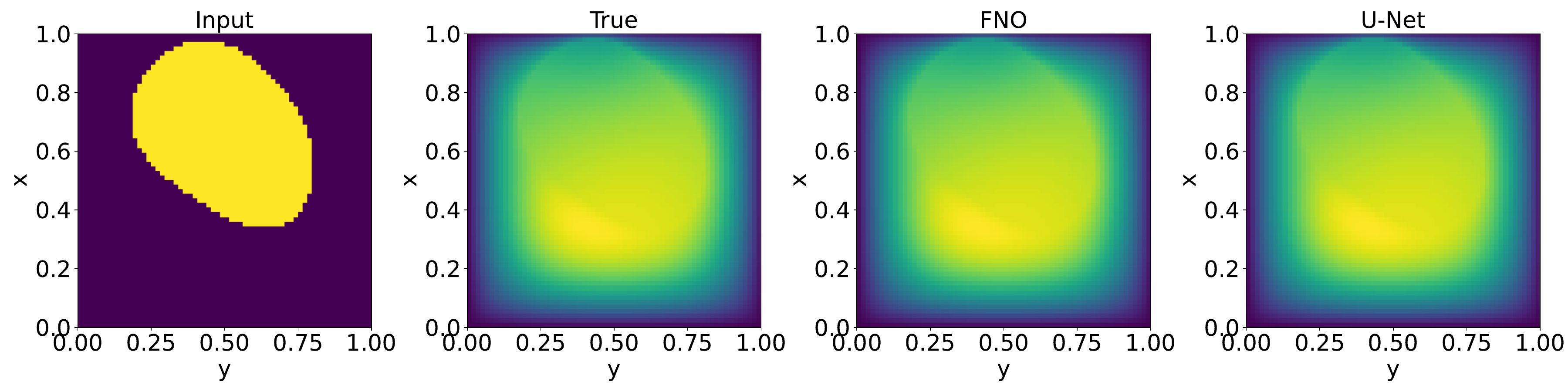}
    \caption{Plots of the predictions for 2D Darcy Flow.}
    \label{fig:vis_2d_darcy_pred}
\end{figure}

\begin{figure}[!]
    \centering
    \includegraphics[width=\textwidth]{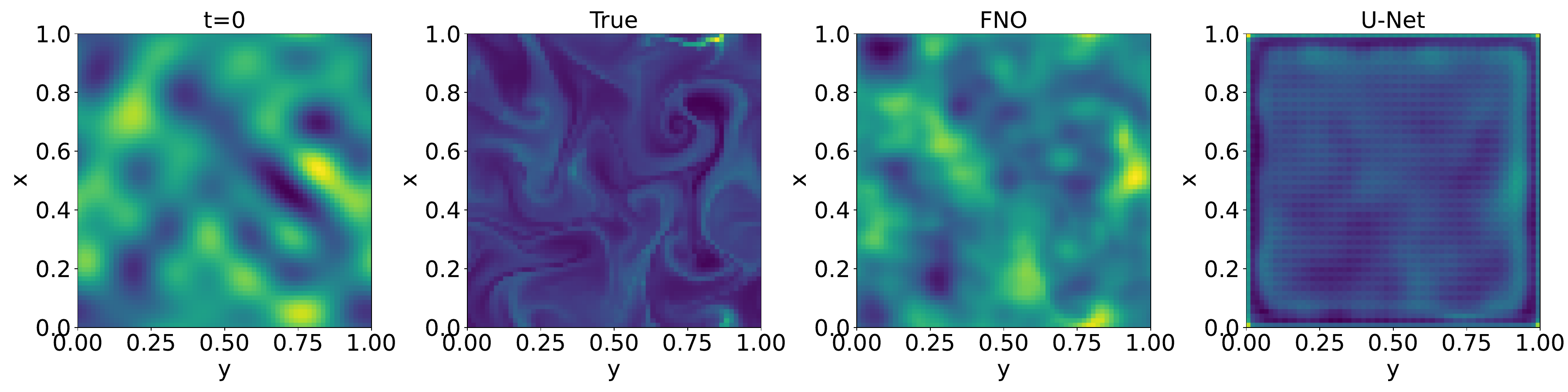}
    \caption{Plots of the predictions of the density for 2D compressible NS equations at the final time-step.}
    \label{fig:vis_2d_CFD_pred}
\end{figure}

In this section, we present visualizations of the baseline model predictions, compared against the generated datasets for the diffusion-sorption equation (\autoref{fig:vis_diff-sorp_data-fno-unet}), 2D diffusion-reaction equation (\autoref{fig:vis_diff-react_data}, \autoref{fig:vis_diff-react_fno}, and \autoref{fig:vis_diff-react_unet}), the shallow-water equation (\autoref{fig:vis_rdb_data}, \autoref{fig:vis_rdb_fno}, and \autoref{fig:vis_rdb_unet}), 1D Advection equation \autoref{fig:vis_1d_adv_pred}, 1D Burgers equation \autoref{fig:vis_1d_bgs_pred}, 1D Reaction-Diffusion equation \autoref{fig:vis_1d_reac-diff_pred}, 1D compressible NS equations \autoref{fig:vis_1d_cfd_pred}, 2D Darcy flow \autoref{fig:vis_2d_darcy_pred}, and 2D compressible NS equations \autoref{fig:vis_2d_CFD_pred}.

\begin{figure}[!]
    \centering
        \begin{subfigure}{0.24\textwidth}
         \centering
         \includegraphics[width=\textwidth]{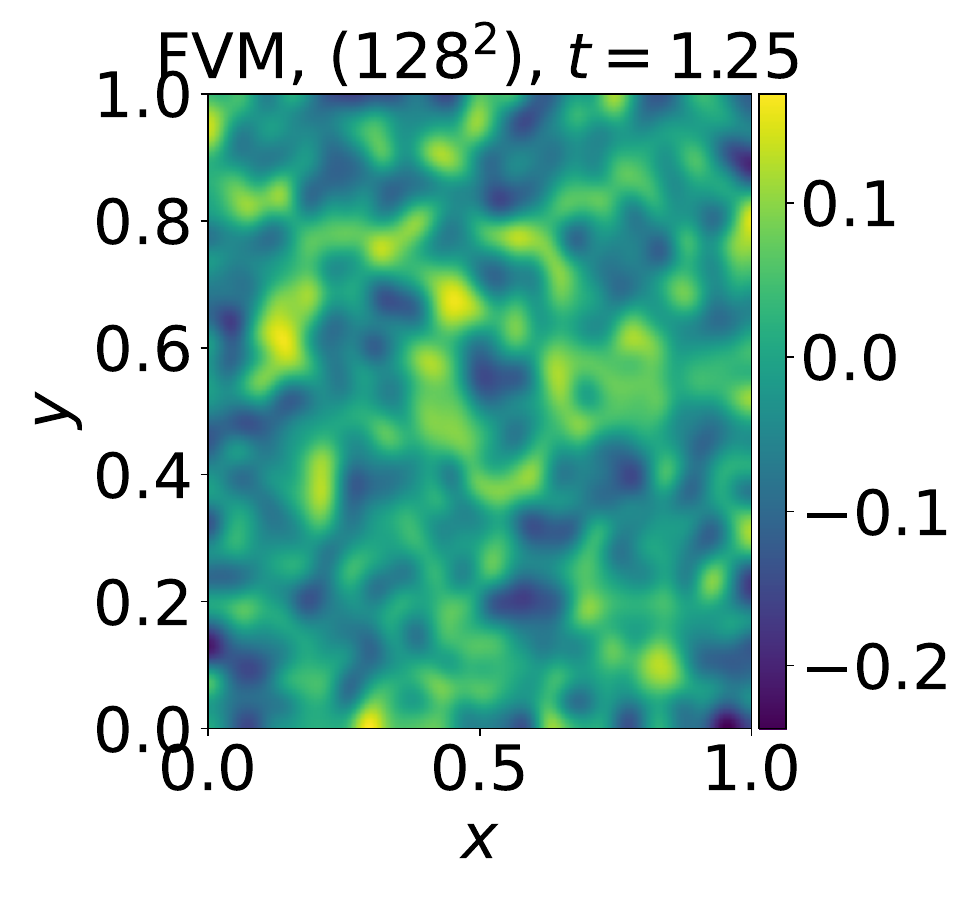}
         \caption{}%Plots of the RMSE calculated at different unrolled time steps.}
    %\label{fig:vis_diff-react_data}
    \end{subfigure}
    \hfill
    \begin{subfigure}{0.24\textwidth}
         \centering
         \includegraphics[width=\textwidth]{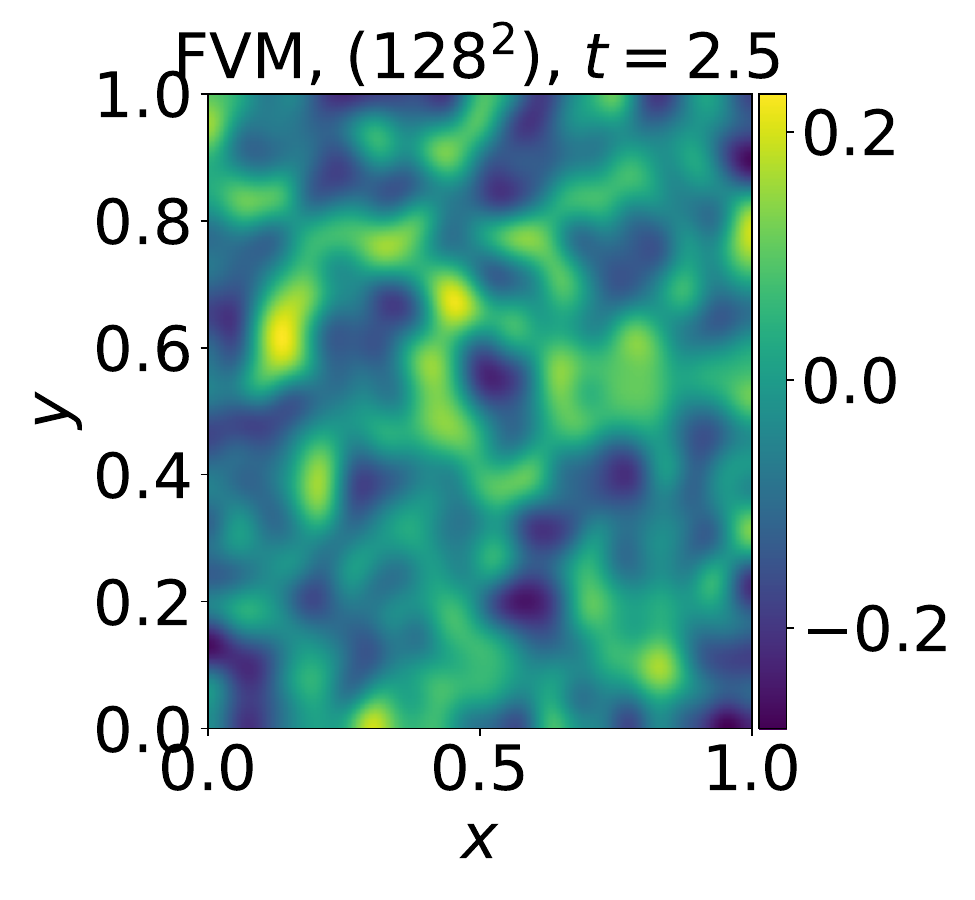}
         \caption{}
         %\label{fig:vis_diff-react_fno}
     \end{subfigure}
     \hfill
     \begin{subfigure}{0.24\textwidth}
         \centering
         \includegraphics[width=\textwidth]{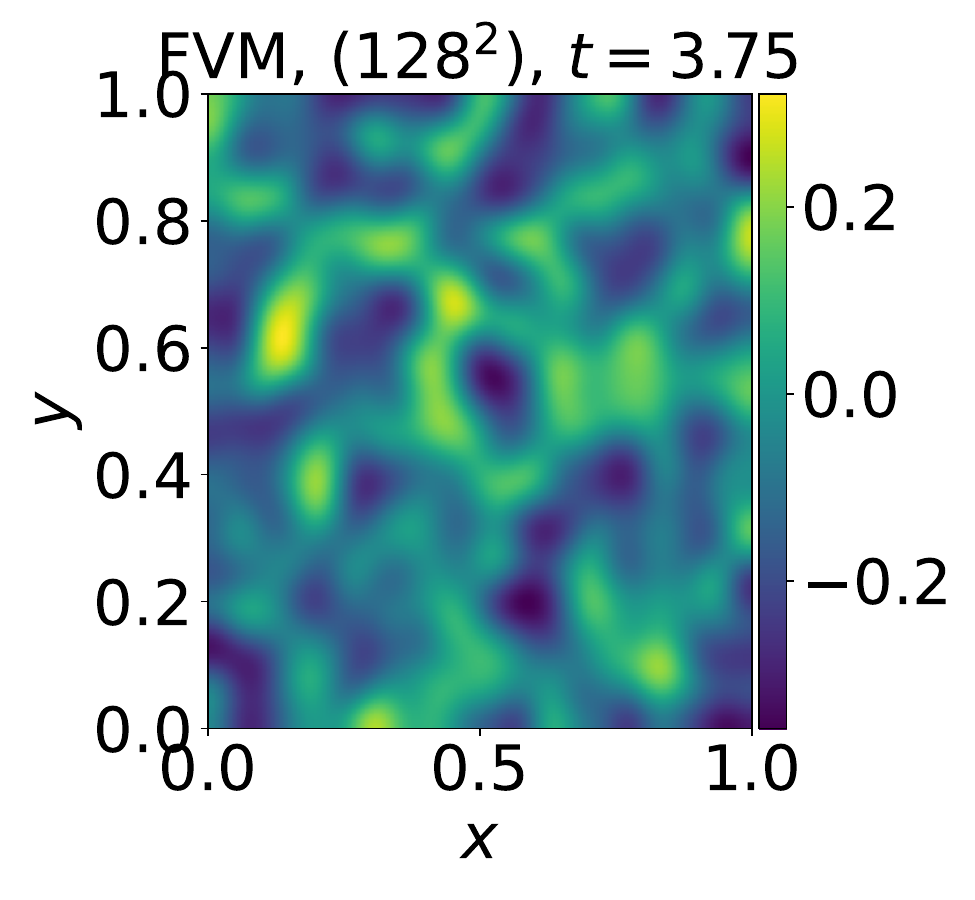}
         \caption{}
         %\label{fig:vis_diff-react_fno}
     \end{subfigure}
     \hfill
     \begin{subfigure}{0.24\textwidth}
         \centering
         \includegraphics[width=\textwidth]{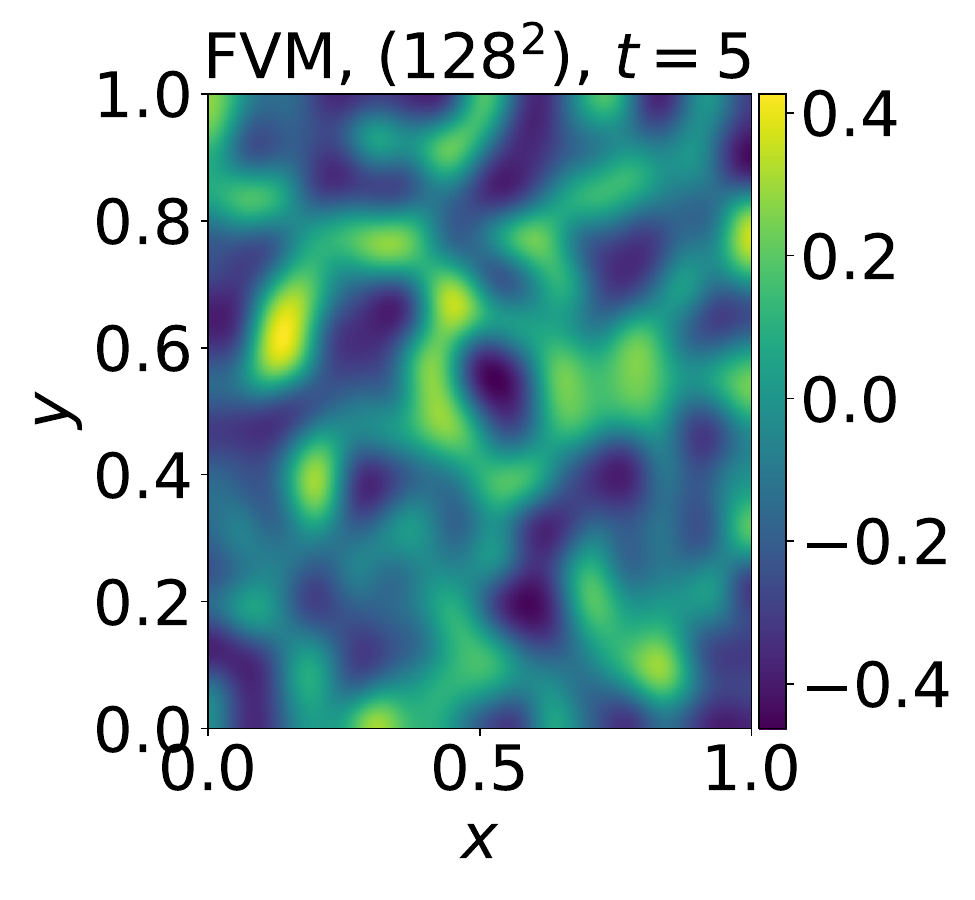}
         \caption{}
         %\label{fig:vis_diff-react_fno}
     \end{subfigure}
     \vspace{-2em} % REMOVING SUBFIGURE LABELS
     \caption{Visualization of the time evolution of the 2D diffusion-reaction equation data.}
    \label{fig:vis_diff-react_data}
\end{figure}

\begin{figure}[!]
    \centering
        \begin{subfigure}{0.24\textwidth}
         \centering
         \includegraphics[width=\textwidth]{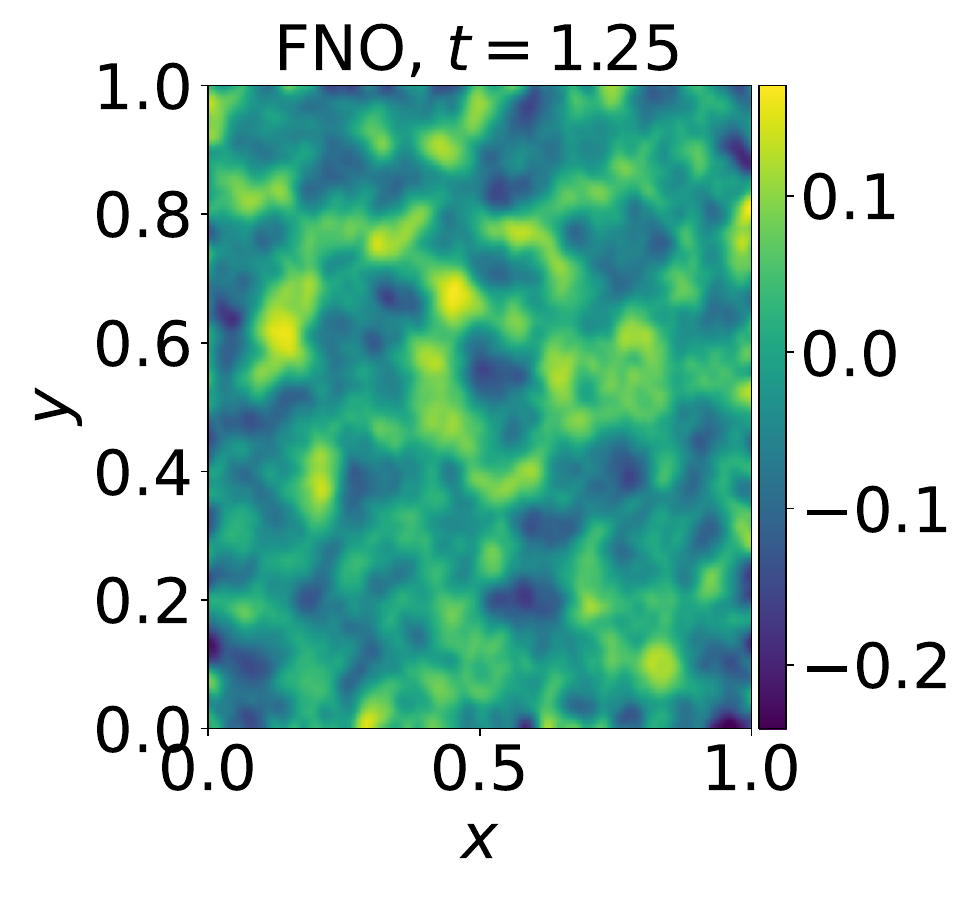}
         \caption{}%Plots of the RMSE calculated at different unrolled time steps.}
    %\label{fig:vis_diff-react_data}
    \end{subfigure}
    \hfill
    \begin{subfigure}{0.24\textwidth}
         \centering
         \includegraphics[width=\textwidth]{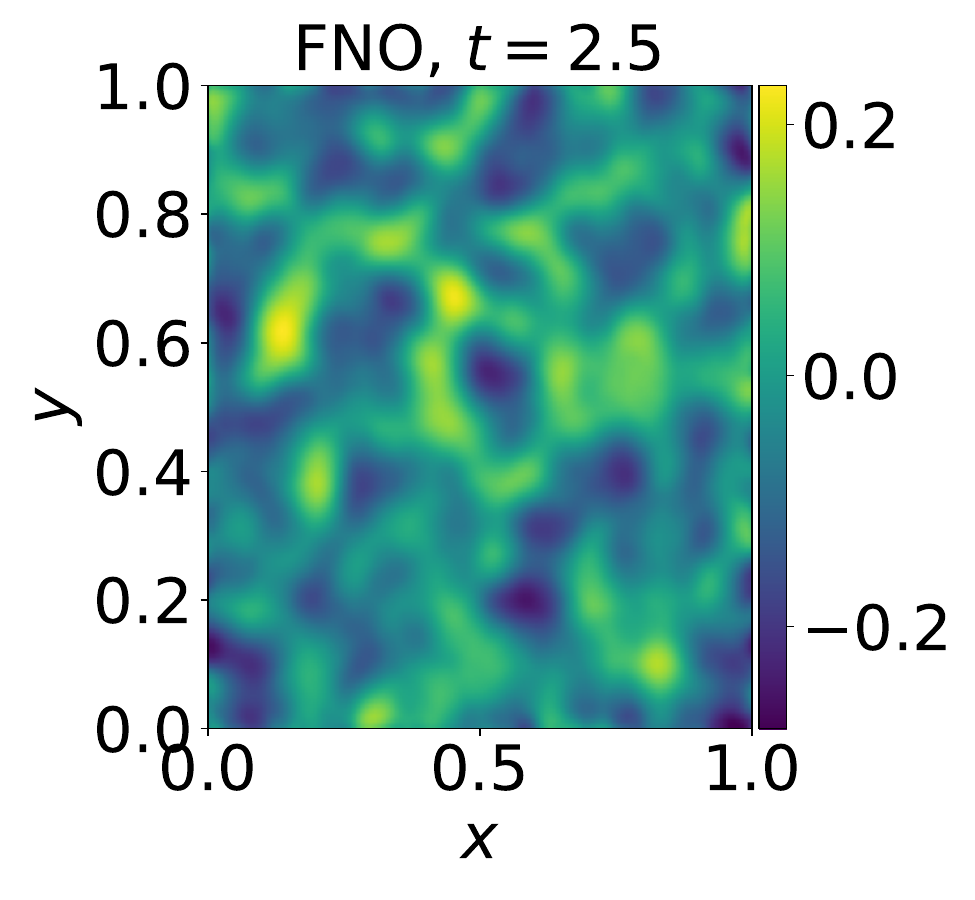}
         \caption{}
         %\label{fig:vis_diff-react_fno}
     \end{subfigure}
     \hfill
     \begin{subfigure}{0.24\textwidth}
         \centering
         \includegraphics[width=\textwidth]{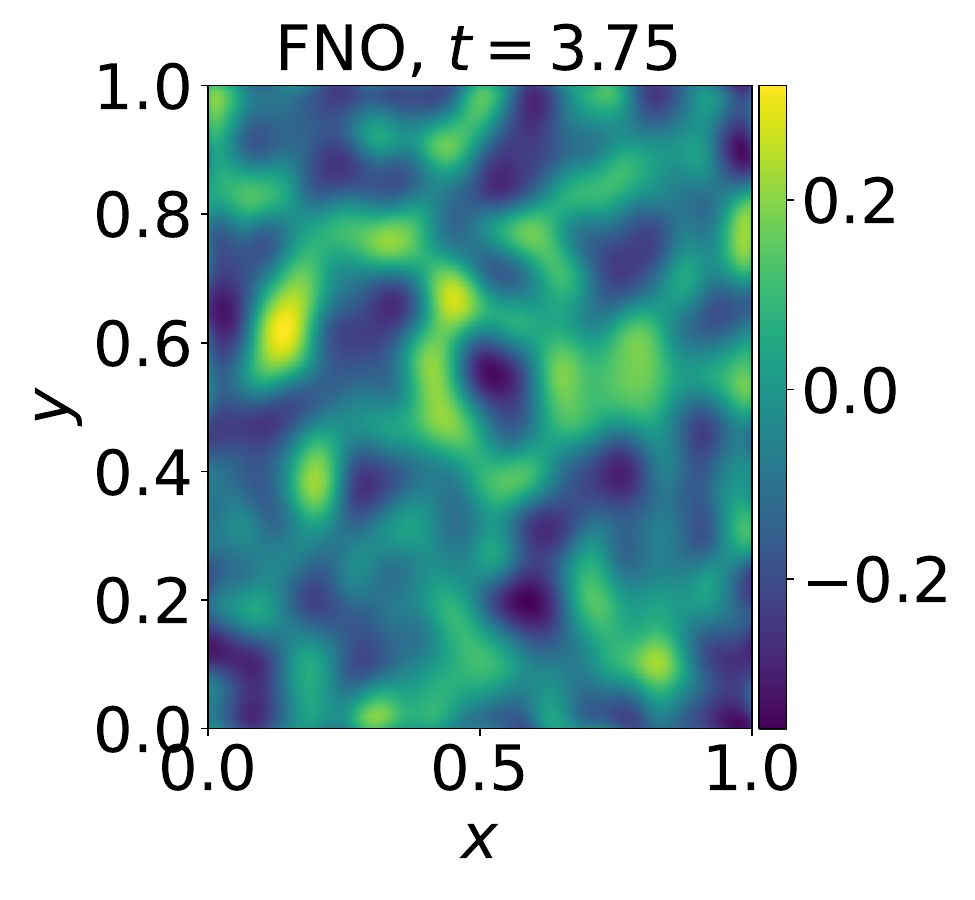}
         \caption{}
         %\label{fig:vis_diff-react_fno}
     \end{subfigure}
     \hfill
     \begin{subfigure}{0.24\textwidth}
         \centering
         \includegraphics[width=\textwidth]{figures/timothy/diff-react/2D_diff-react_NA_NA_FNO_4.pdf}
         \caption{}
         %\label{fig:vis_diff-react_fno}
     \end{subfigure}
     \vspace{-2em} % REMOVING SUBFIGURE LABELS
     \caption{Visualization of the time evolution of the 2D diffusion-reaction equation predicted using FNO.}
    \label{fig:vis_diff-react_fno}
\end{figure}

\begin{figure}[!]
    \centering
        \begin{subfigure}{0.24\textwidth}
         \centering
         \includegraphics[width=\textwidth]{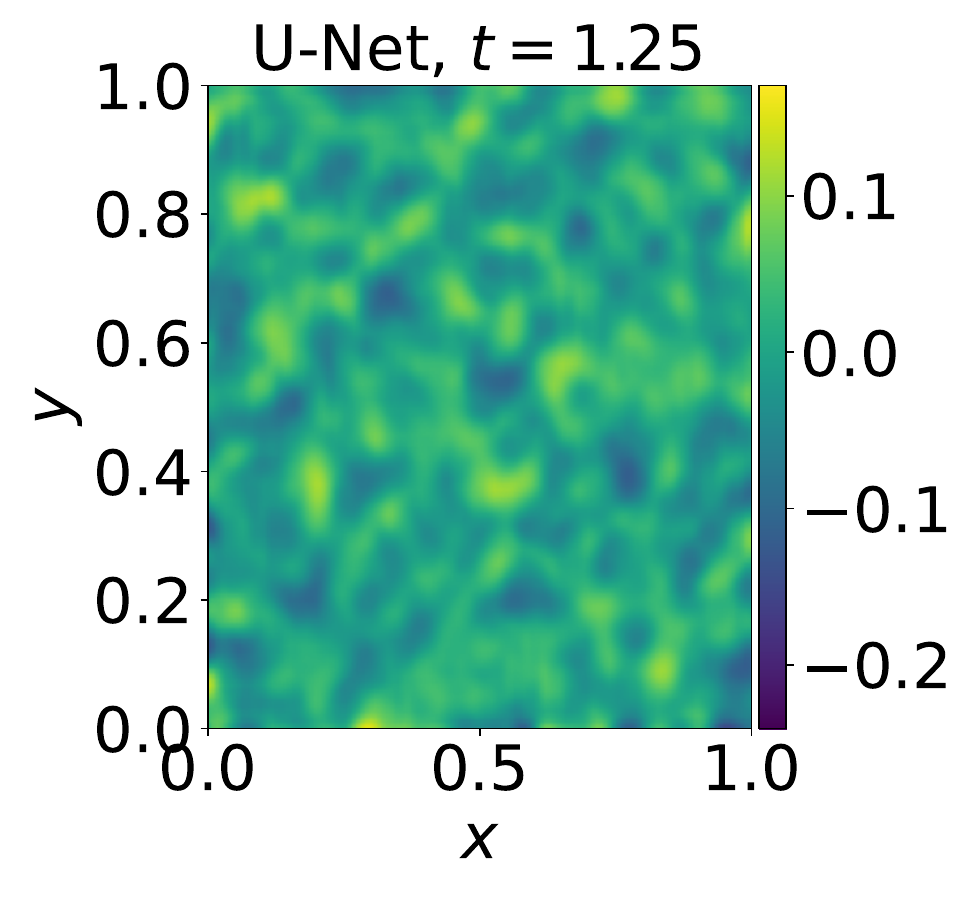}
         \caption{}%Plots of the RMSE calculated at different unrolled time steps.}
    %\label{fig:vis_diff-react_data}
    \end{subfigure}
    \hfill
    \begin{subfigure}{0.24\textwidth}
         \centering
         \includegraphics[width=\textwidth]{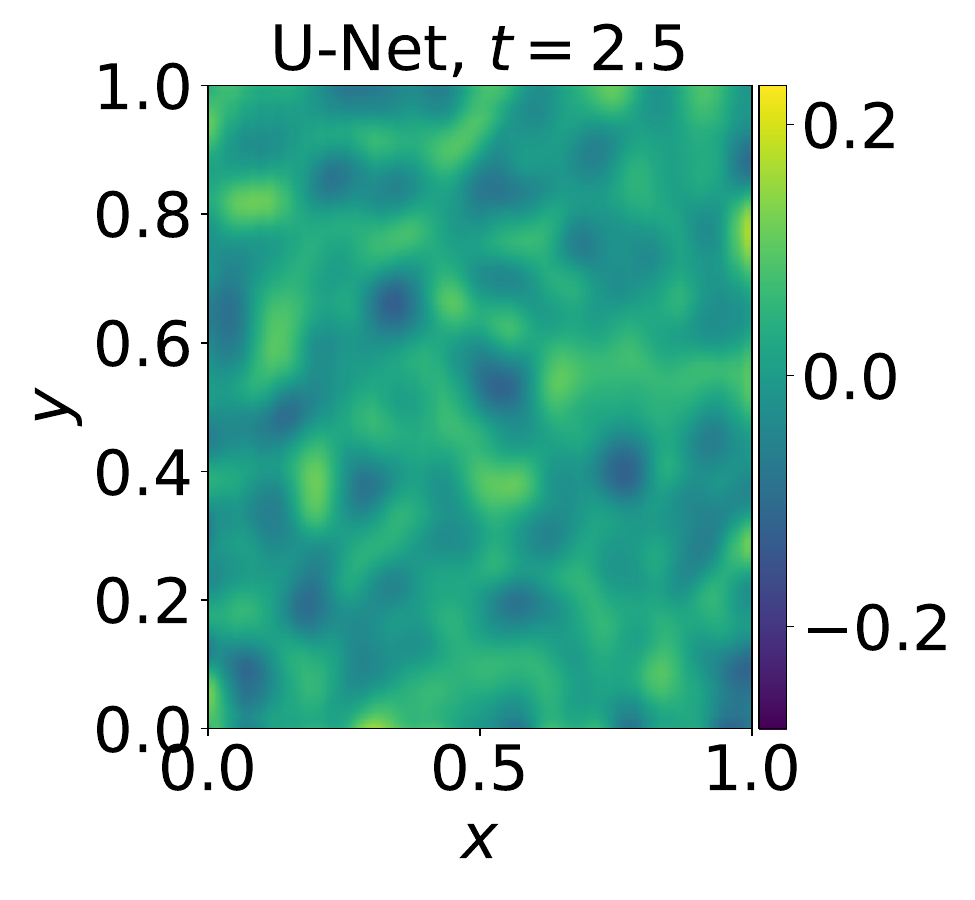}
         \caption{}
         %\label{fig:vis_diff-react_fno}
     \end{subfigure}
     \hfill
     \begin{subfigure}{0.24\textwidth}
         \centering
         \includegraphics[width=\textwidth]{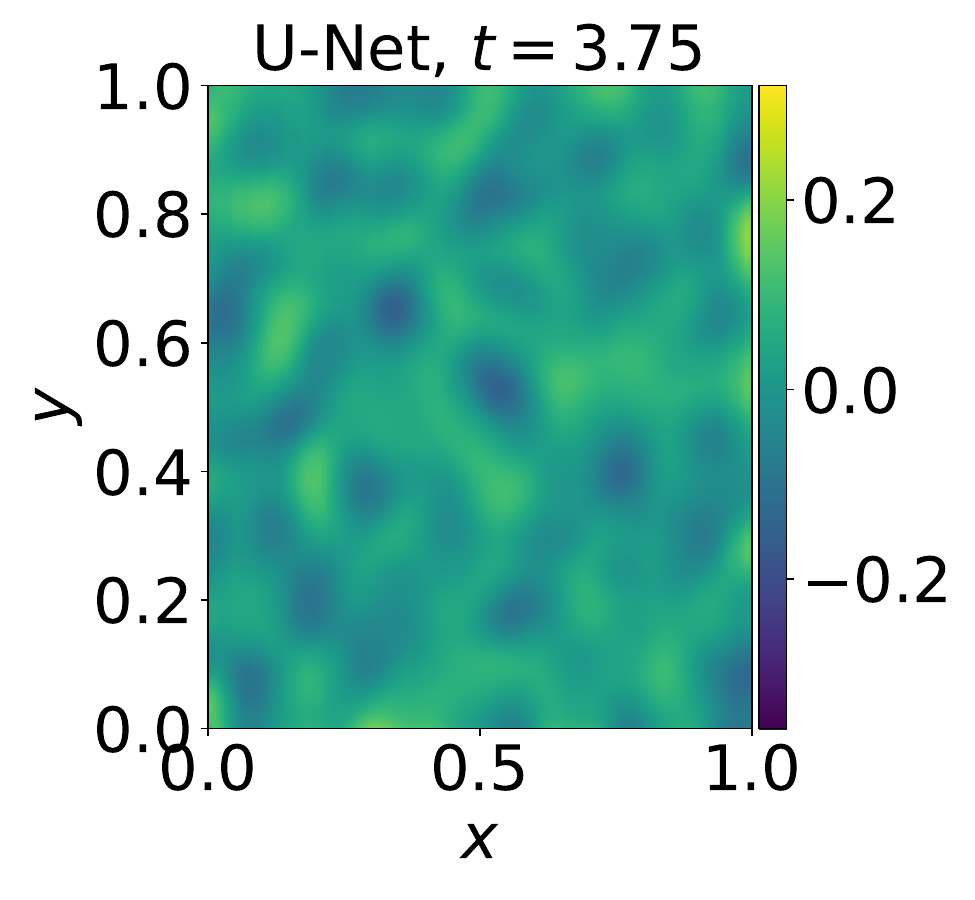}
         \caption{}
         %\label{fig:vis_diff-react_fno}
     \end{subfigure}
     \hfill
     \begin{subfigure}{0.24\textwidth}
         \centering
         \includegraphics[width=\textwidth]{figures/timothy/diff-react/2D_diff-react_NA_NA_Unet_PF_20_4.pdf}
         \caption{}
         %\label{fig:vis_diff-react_fno}
     \end{subfigure}
     \vspace{-2em} % REMOVING SUBFIGURE LABELS
     \caption{Visualization of the time evolution of the 2D diffusion-reaction equation predicted using U-Net.}
    \label{fig:vis_diff-react_unet}
\end{figure}

\section{Visualization of Initial Conditions}
\begin{figure}[!]
    \centering
        \begin{subfigure}{0.19\textwidth}
         \centering
         \includegraphics[width=\textwidth]{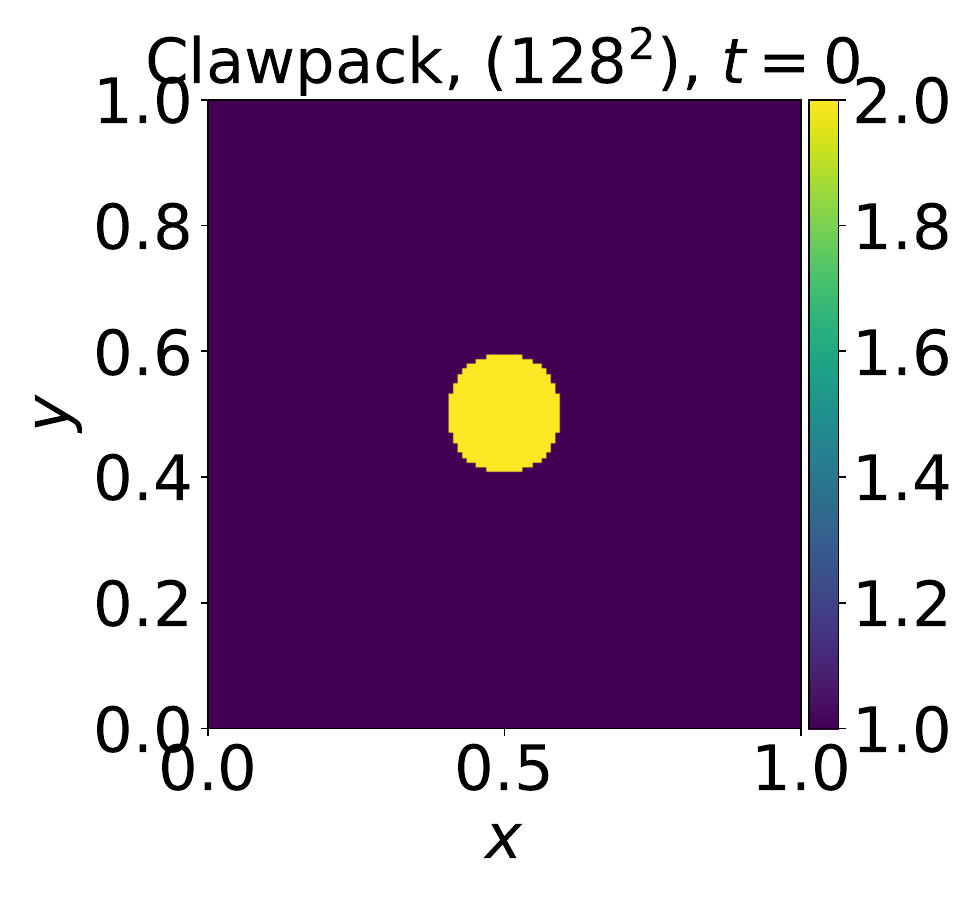}
         \caption{}
     \end{subfigure}    
    \hfill
    \begin{subfigure}{0.19\textwidth}
         \centering
         \includegraphics[width=\textwidth]{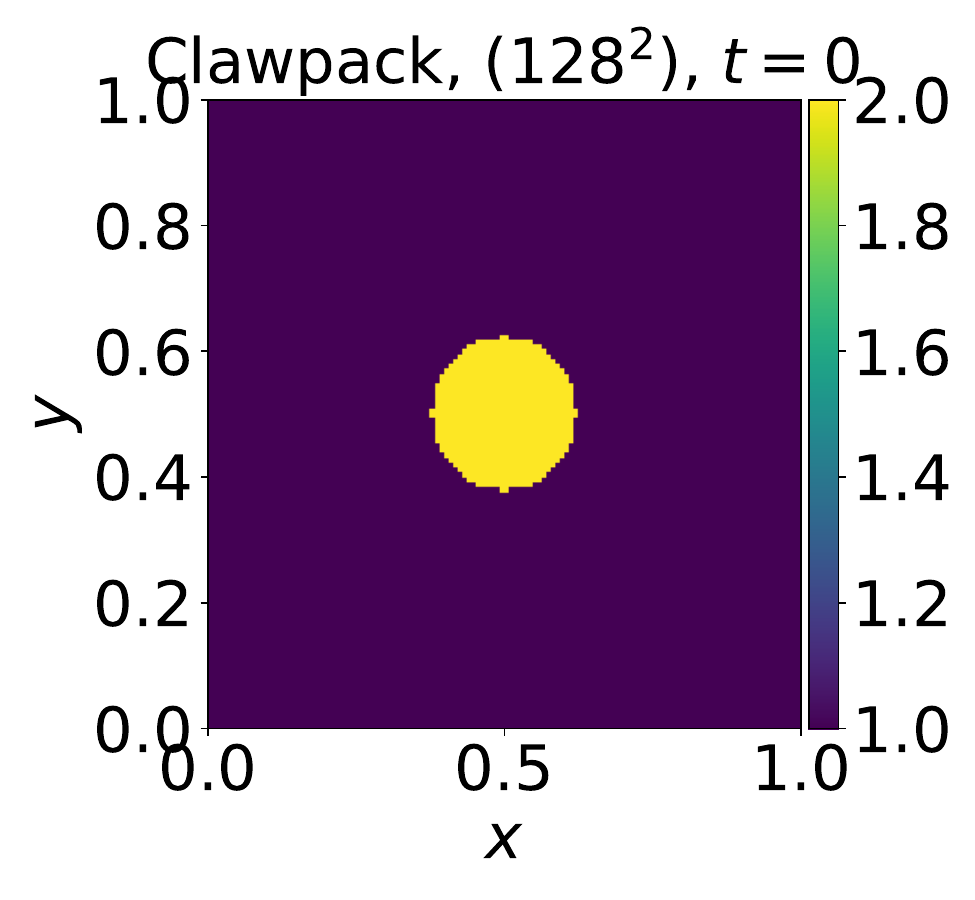}
         \caption{}
     \end{subfigure}
     \hfill
     \begin{subfigure}{0.19\textwidth}
         \centering
         \includegraphics[width=\textwidth]{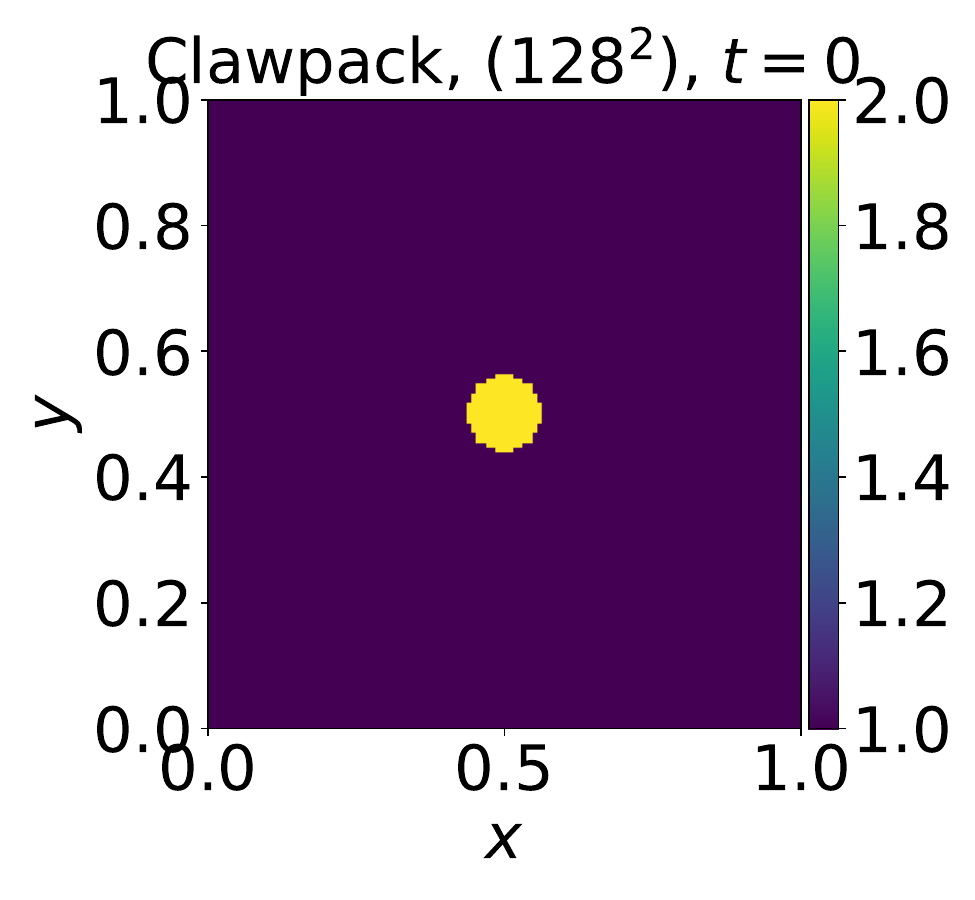}
         \caption{}
     \end{subfigure}
     \hfill
     \begin{subfigure}{0.19\textwidth}
         \centering
         \includegraphics[width=\textwidth]{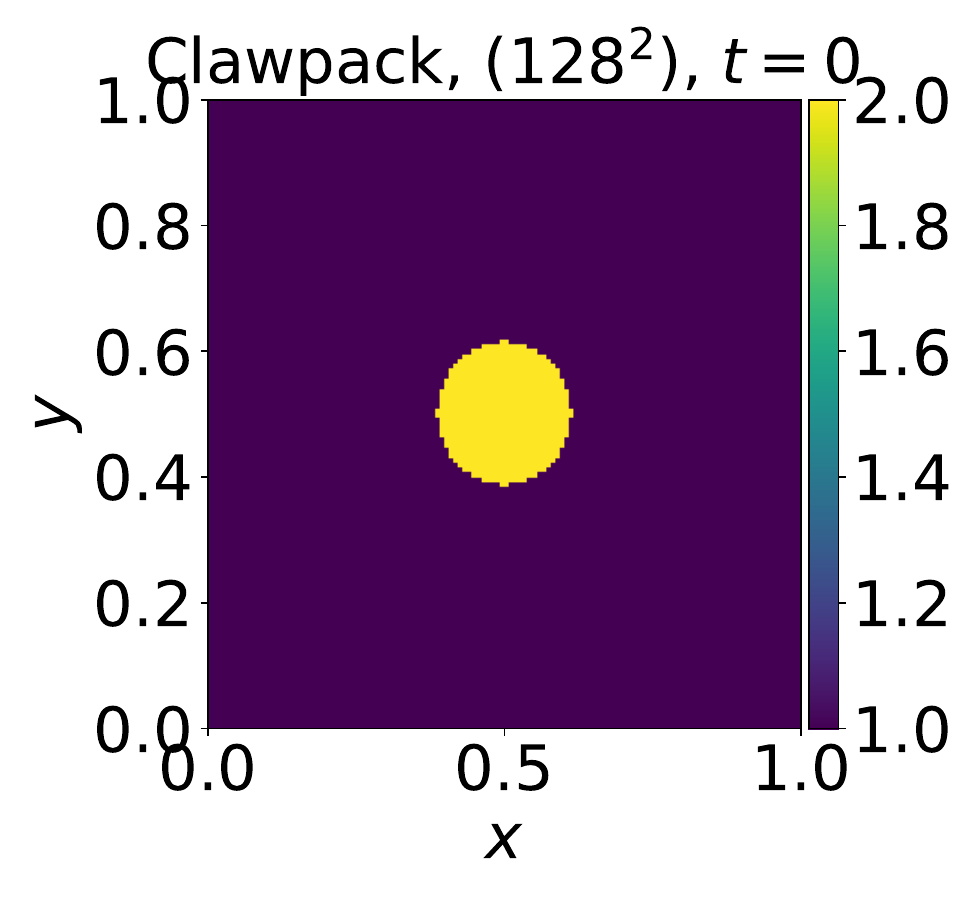}
         \caption{}
     \end{subfigure}
     \hfill
     \begin{subfigure}{0.19\textwidth}
         \centering
         \includegraphics[width=\textwidth]{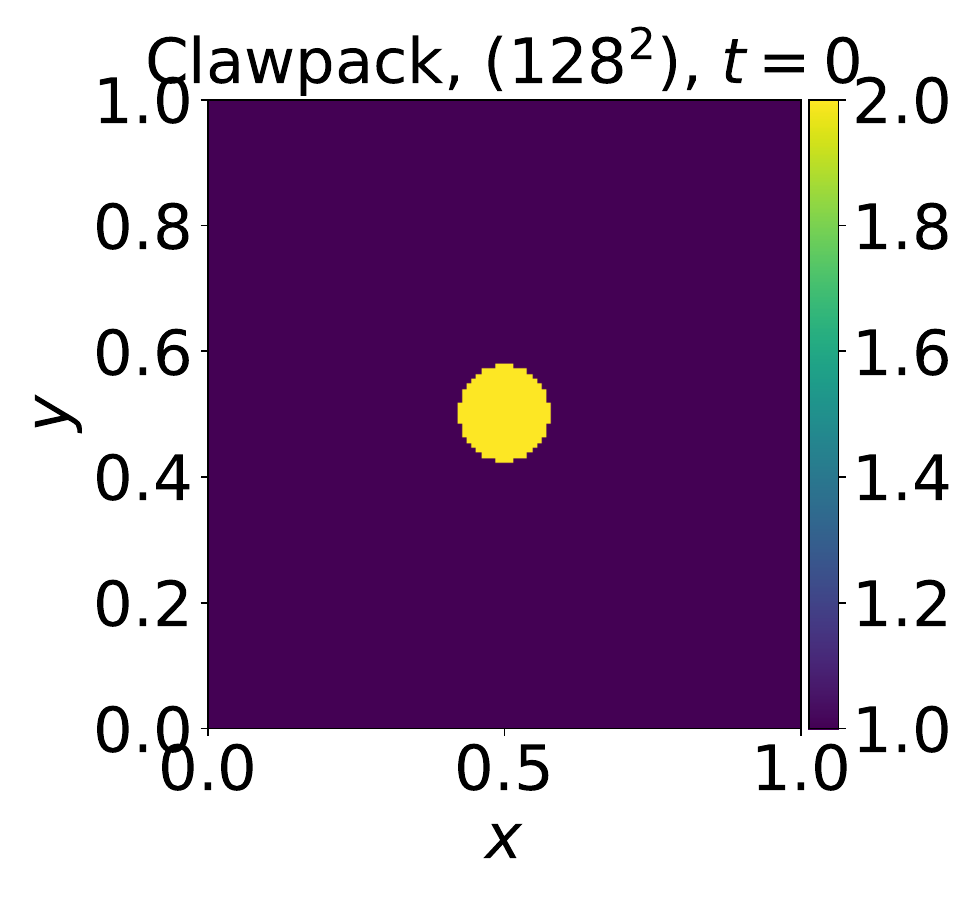}
         \caption{}
     \end{subfigure}
     \vspace{-2em} % REMOVING SUBFIGURE LABELS
     \caption{Visualization of the different radius of the initial perturbation used for the 2D shallow-water equations data.}
    \label{fig:vis_rdb_init}
\end{figure}

\begin{figure}[!]
    \centering
        \begin{subfigure}{0.19\textwidth}
         \centering
         \includegraphics[width=\textwidth]{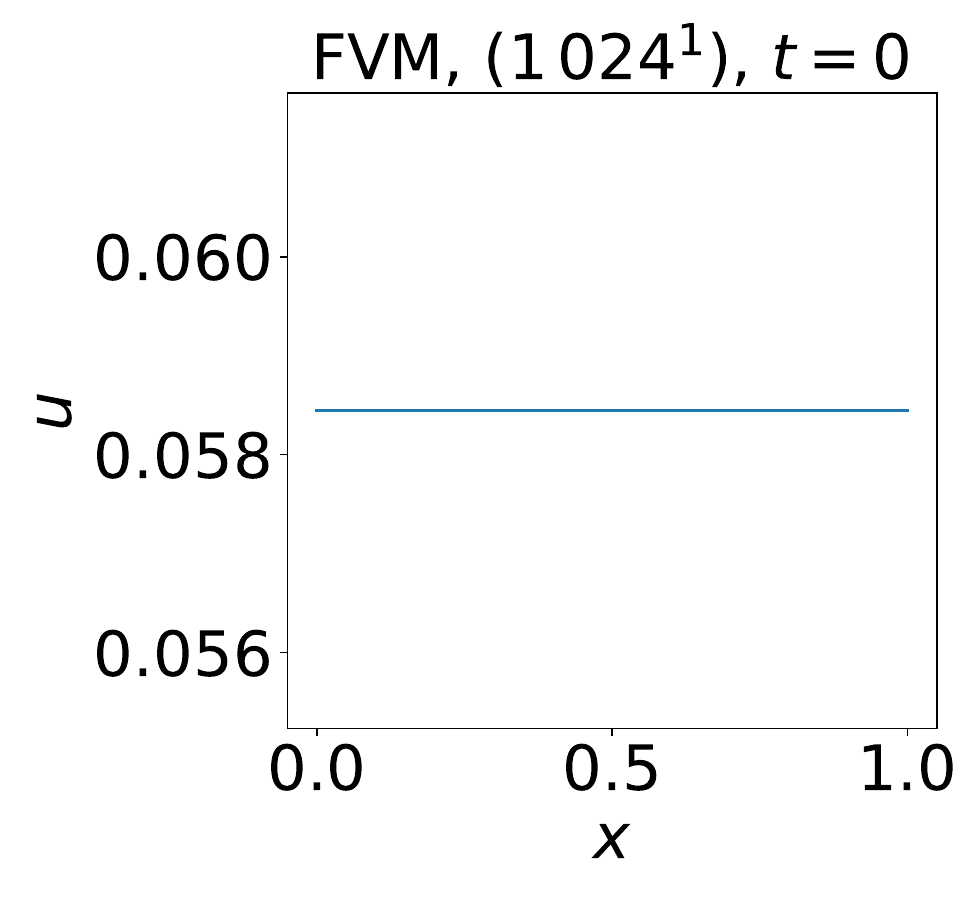}
         \caption{}
     \end{subfigure}    
    \hfill
    \begin{subfigure}{0.19\textwidth}
         \centering
         \includegraphics[width=\textwidth]{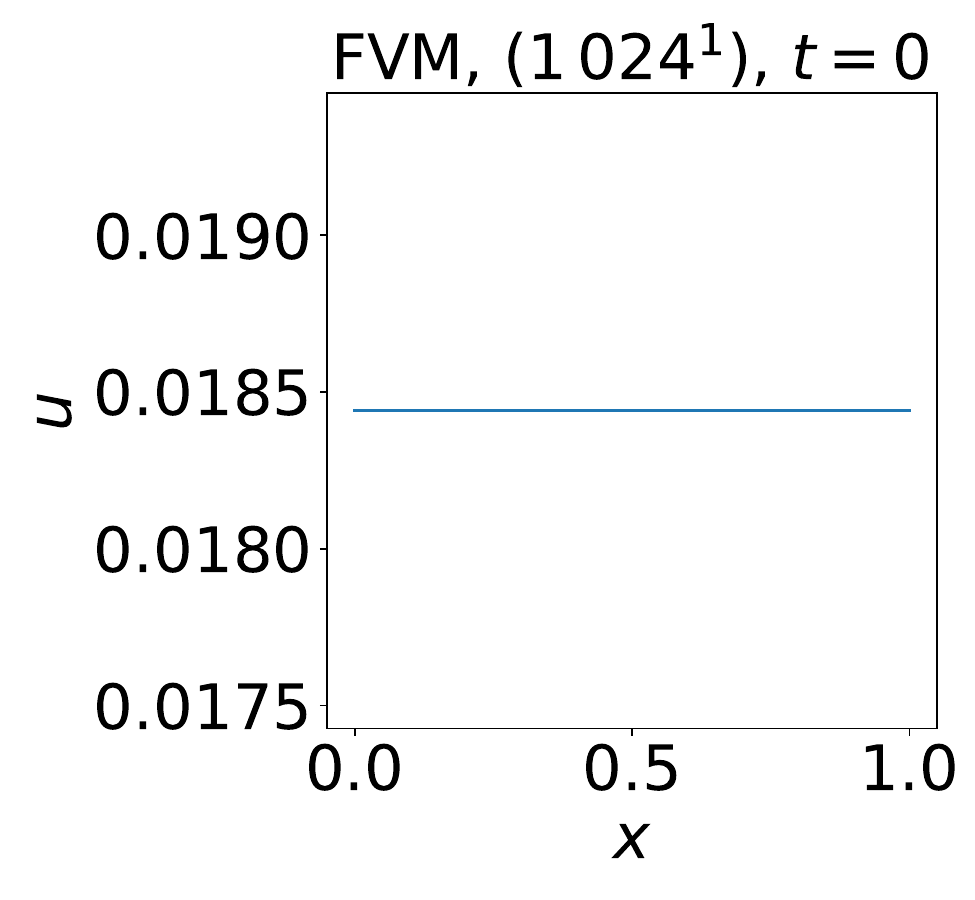}
         \caption{}
     \end{subfigure}
     \hfill
     \begin{subfigure}{0.19\textwidth}
         \centering
         \includegraphics[width=\textwidth]{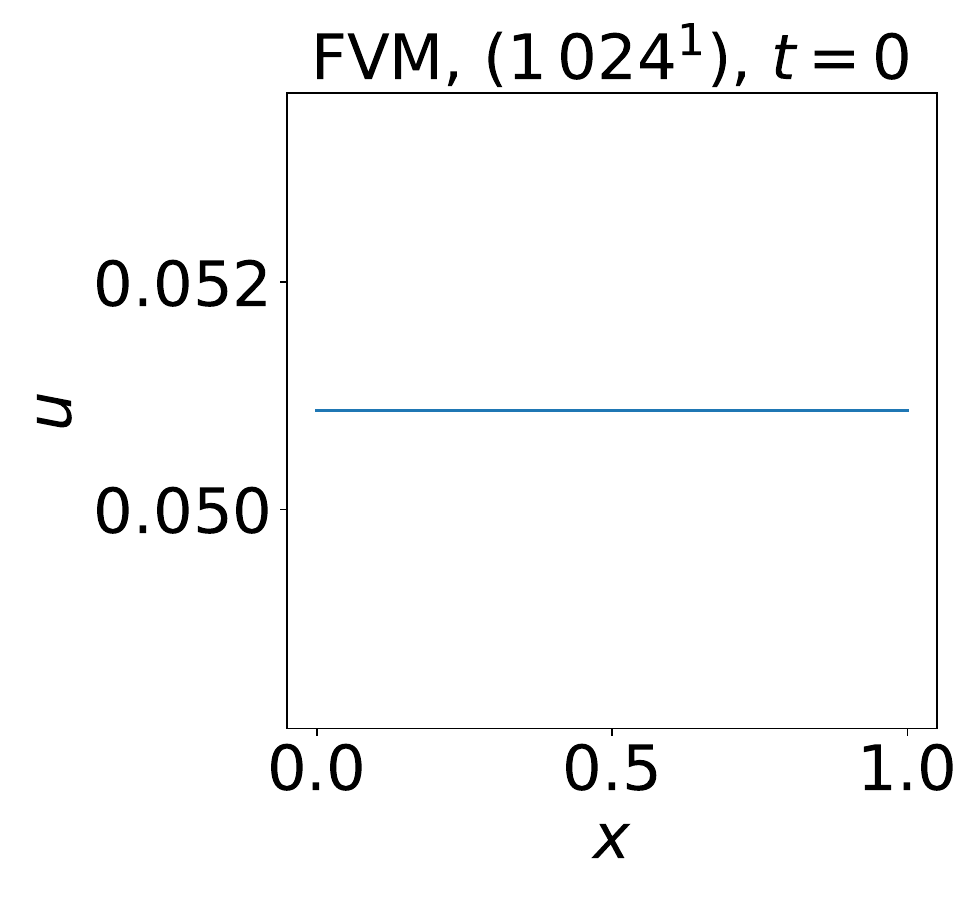}
         \caption{}
     \end{subfigure}
     \hfill
     \begin{subfigure}{0.19\textwidth}
         \centering
         \includegraphics[width=\textwidth]{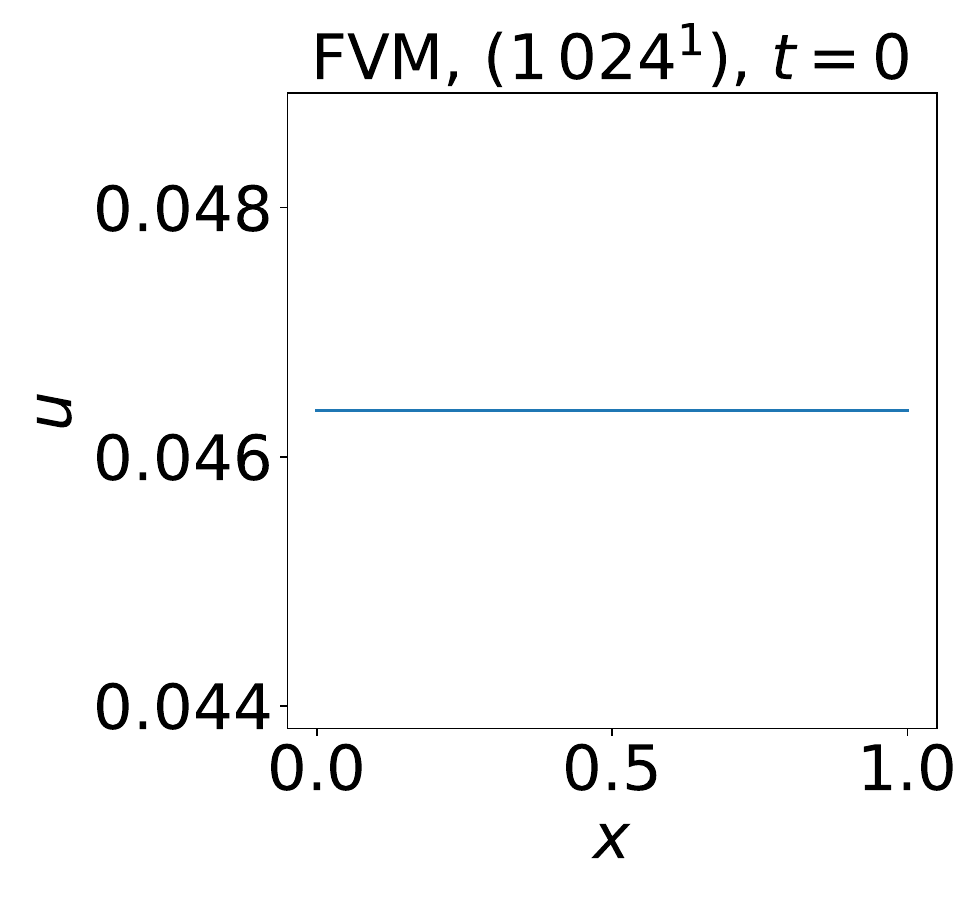}
         \caption{}
     \end{subfigure}
     \hfill
     \begin{subfigure}{0.19\textwidth}
         \centering
         \includegraphics[width=\textwidth]{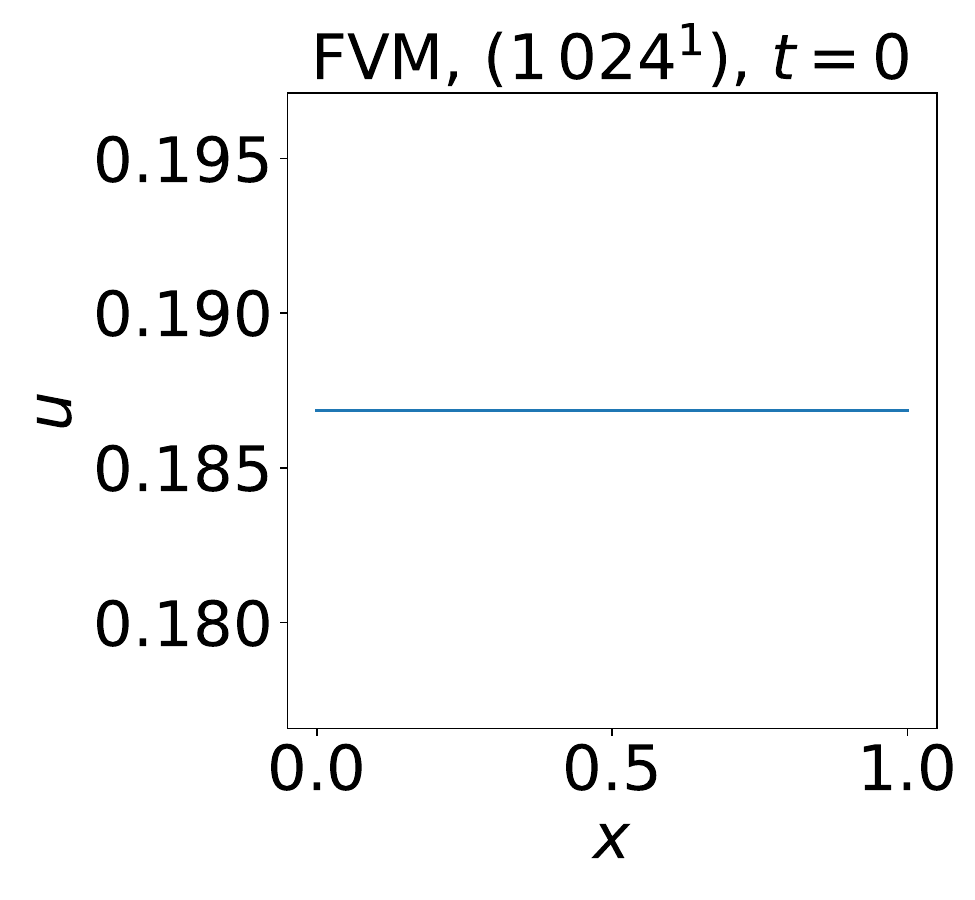}
         \caption{}
     \end{subfigure}
     \vspace{-2em} % REMOVING SUBFIGURE LABELS
     \caption{Visualization of the random uniform initial conditions used for the 1D diffusion-sorption equations data.}
    \label{fig:vis_diff-sorp_init}
\end{figure}

\begin{figure}[!]
    \centering
        \begin{subfigure}{0.19\textwidth}
         \centering
         \includegraphics[width=\textwidth]{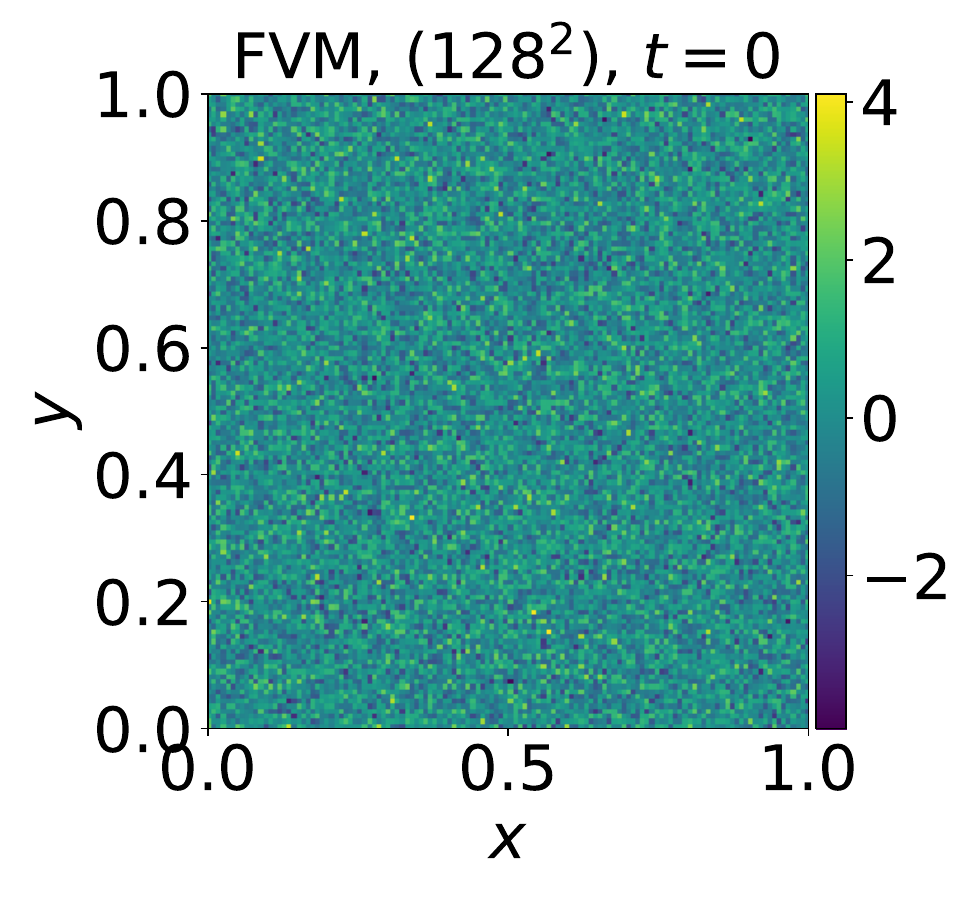}
         \caption{}
     \end{subfigure}    
    \hfill
    \begin{subfigure}{0.19\textwidth}
         \centering
         \includegraphics[width=\textwidth]{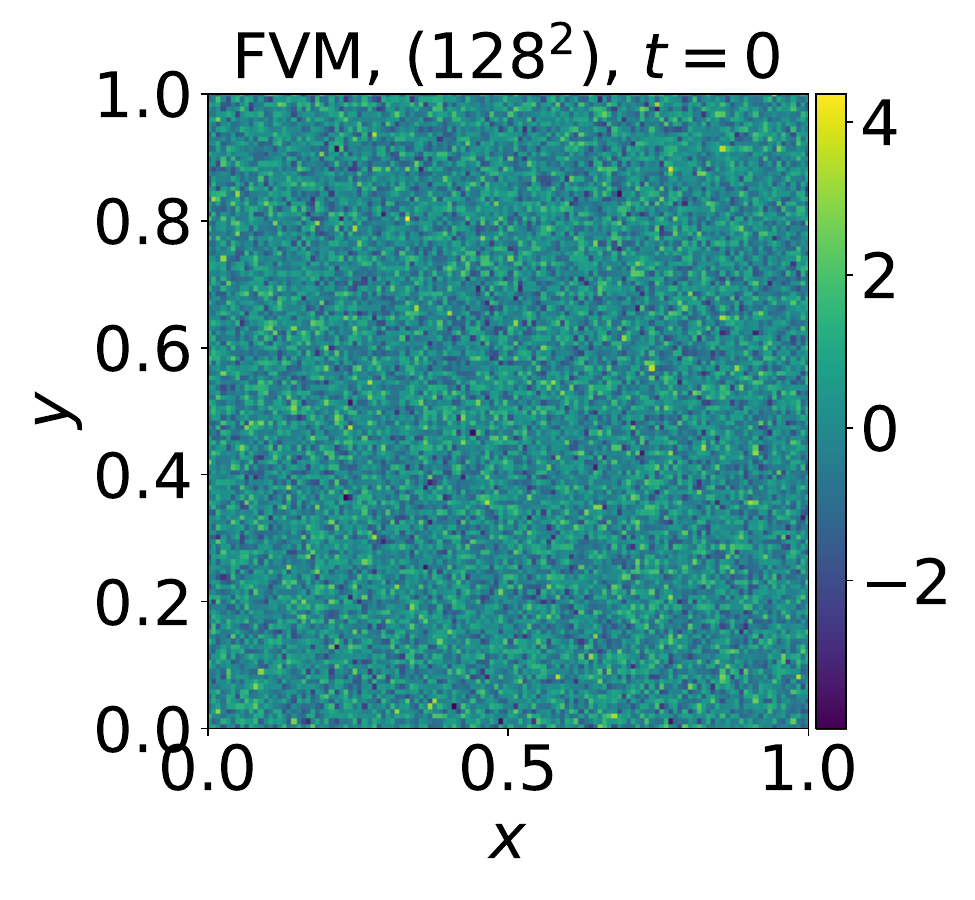}
         \caption{}
     \end{subfigure}
     \hfill
     \begin{subfigure}{0.19\textwidth}
         \centering
         \includegraphics[width=\textwidth]{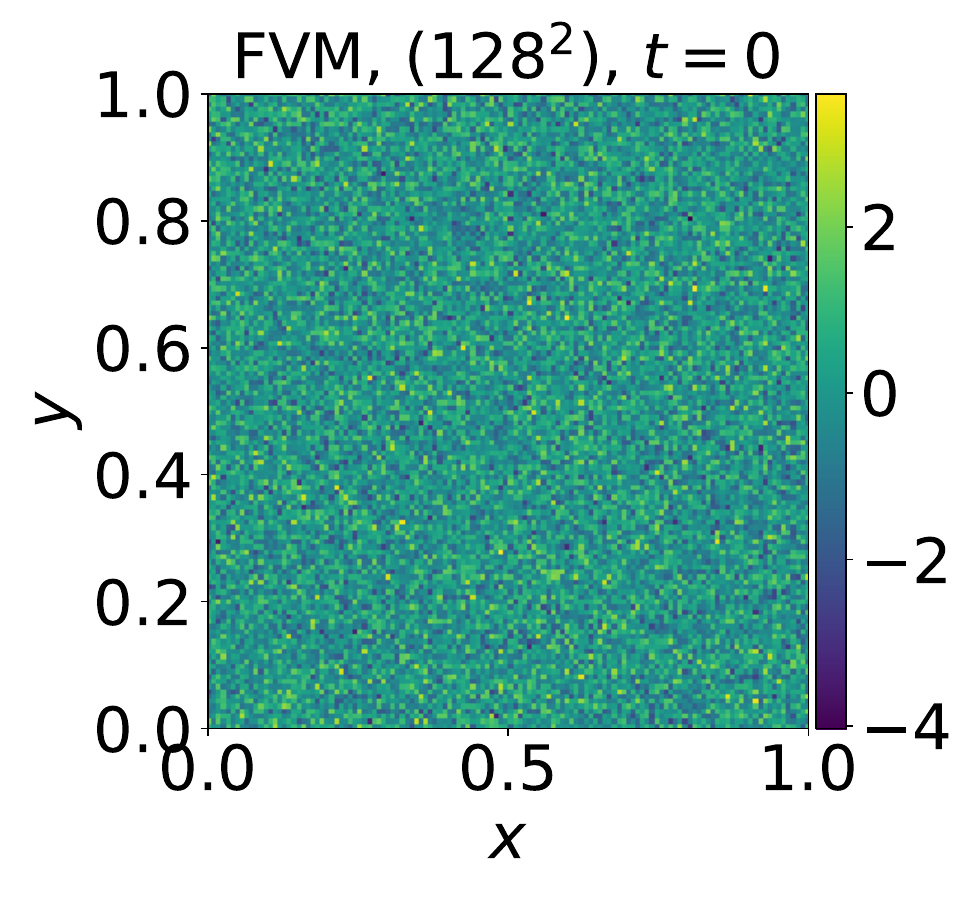}
         \caption{}
     \end{subfigure}
     \hfill
     \begin{subfigure}{0.19\textwidth}
         \centering
         \includegraphics[width=\textwidth]{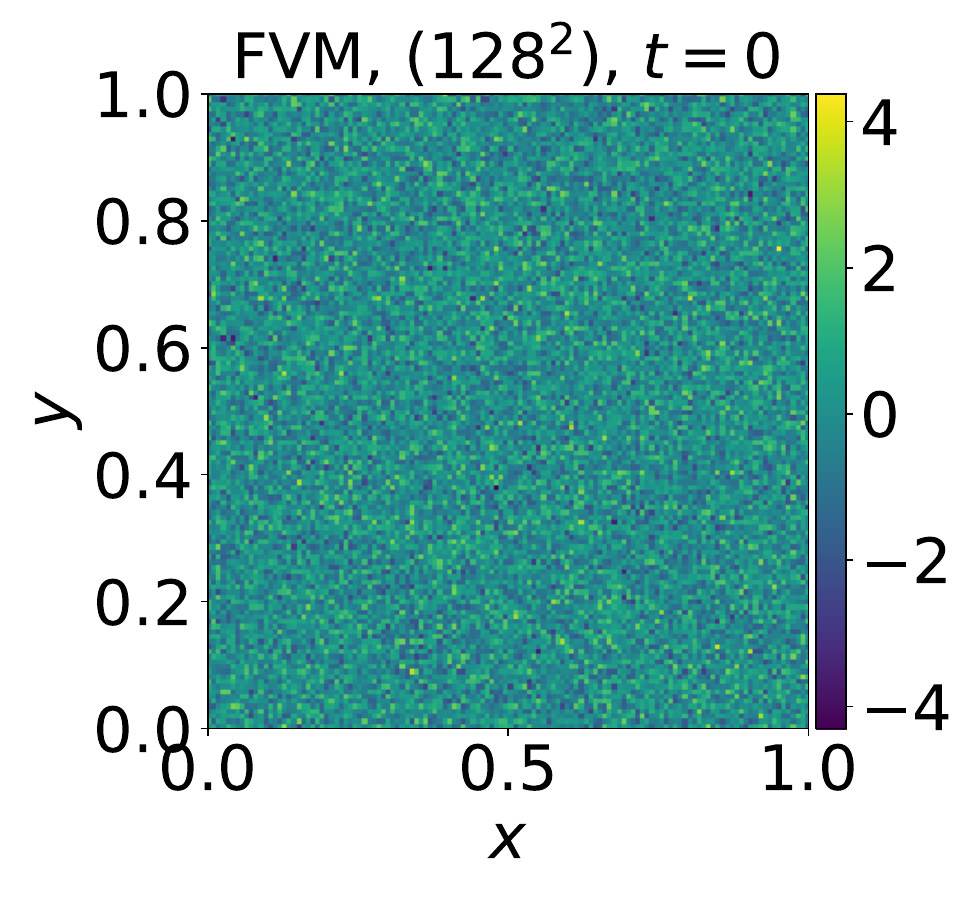}
         \caption{}
     \end{subfigure}
     \hfill
     \begin{subfigure}{0.19\textwidth}
         \centering
         \includegraphics[width=\textwidth]{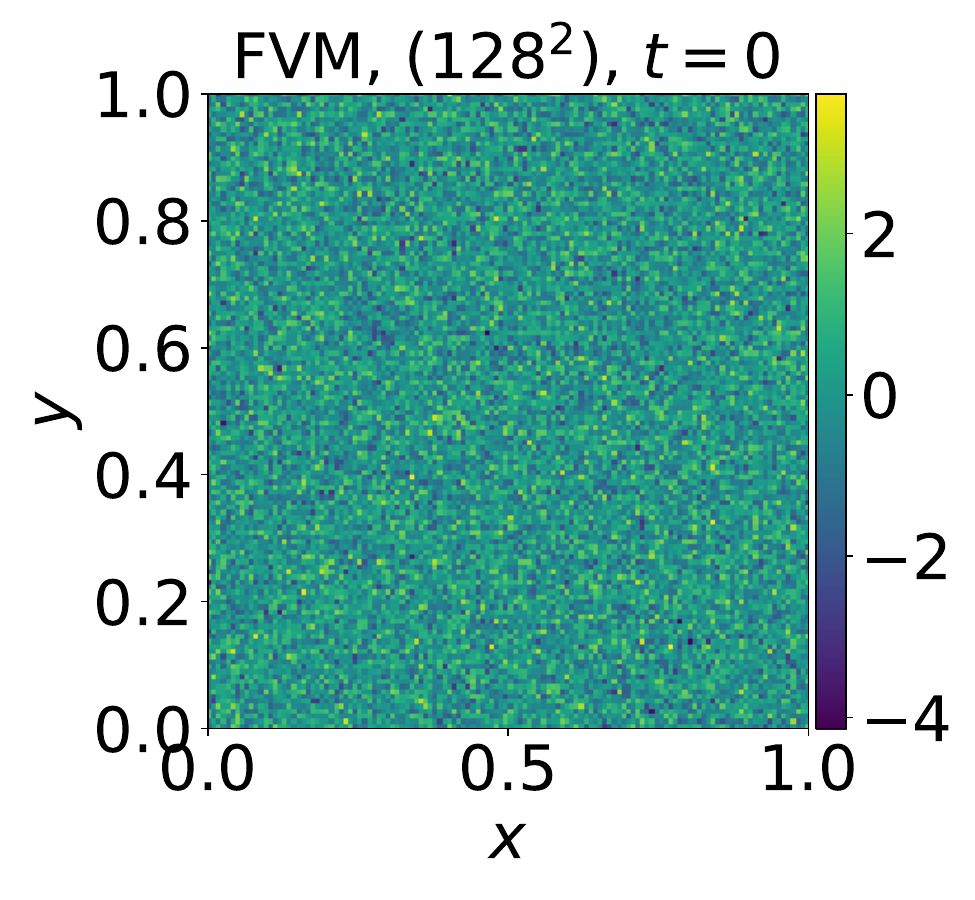}
         \caption{}
     \end{subfigure}
     \vspace{-2em} % REMOVING SUBFIGURE LABELS
     \caption{Visualization of the random initial conditions used for the 2D diffusion-reaction equations data.}
    \label{fig:vis_diff-react_init}
\end{figure}

In this section, we provide a collection of initial condition visualizations for each problem. \autoref{fig:vis_rdb_init} shows different radius of the initial perturbation used as the initial condition for five different samples of the 2D shallow-water equation data. \autoref{fig:vis_diff-sorp_init} shows different random uniform initial condition used for five different samples of the 1D diffusion-sorption equation data. \autoref{fig:vis_diff-react_init} shows different random noise used as the initial condition for five different samples of the 2D diffusion-reaction equation data.

\begin{figure}[!]
    \centering
    \includegraphics[width=\textwidth]{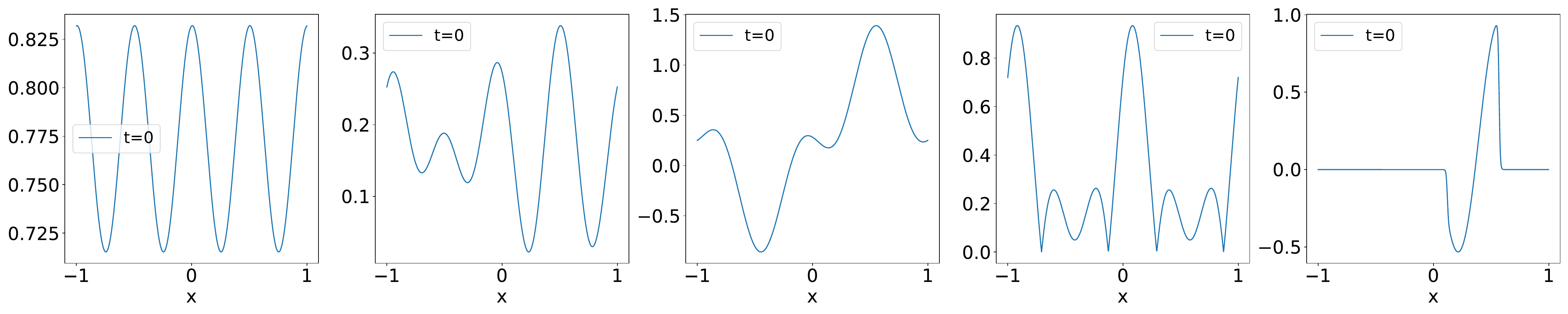}
     \caption{Visualization of the random initial conditions used for the 1D Advection/Burgers equations data.}
    \label{fig:vis_1D_init}
\end{figure}

\begin{figure}[!]
    \centering
    \includegraphics[width=\textwidth]{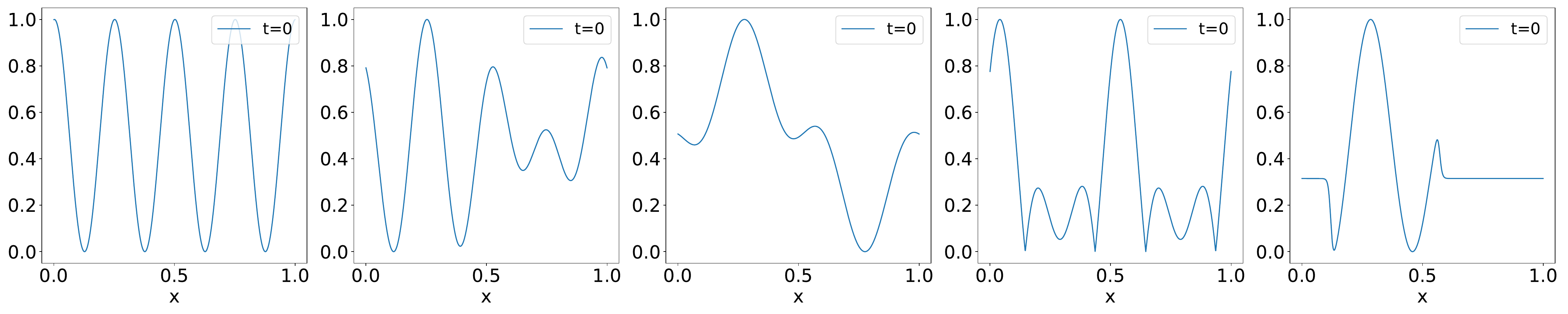}
     \caption{Visualization of the random initial conditions used for the 1D Reaction-Diffusion equations data.}
    \label{fig:vis_1D_reac_diff_init}
\end{figure}

\begin{figure}[!]
    \centering
    \includegraphics[width=\textwidth]{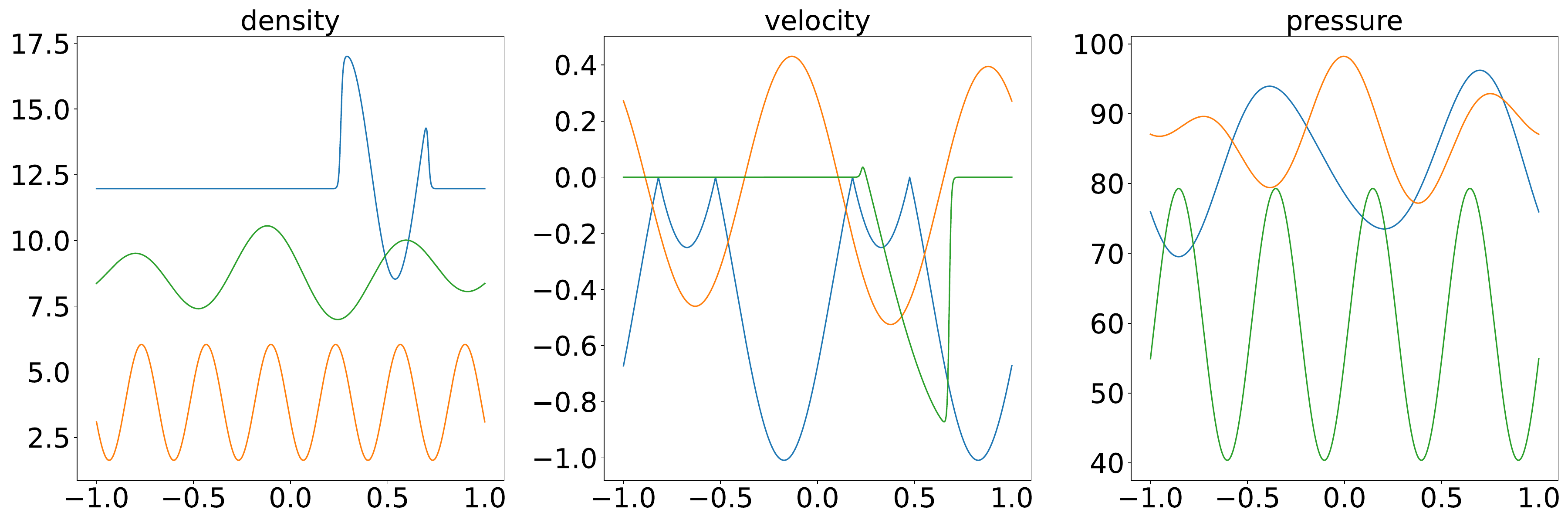}
     \caption{Visualization of the random initial conditions used for the 1D CFD data. The different colors mean the different samples. }
    \label{fig:vis_1D_CFD_init}
\end{figure}

\begin{figure}[!]
    \centering
    \includegraphics[width=\textwidth]{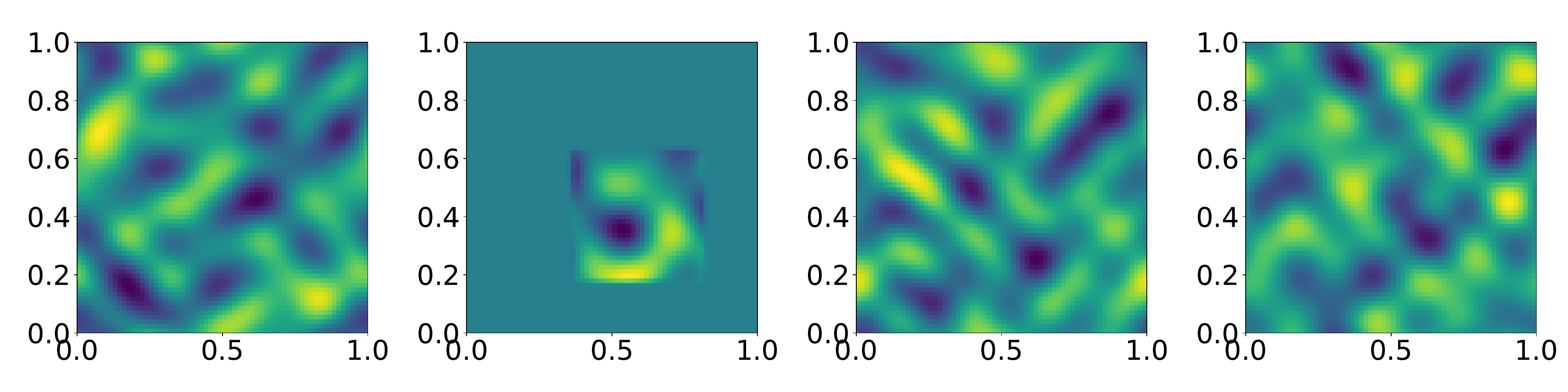}
     \caption{Visualization of the initial conditions of density used for the 2D CFD data.}
    \label{fig:vis_2D_CFD_init}
\end{figure}

In \autoref{fig:vis_1D_init}, we plotted the several samples of the initial condition for 1D Advection and Burgers equations. \autoref{fig:vis_1D_reac_diff_init} is also the similar plot of the initial condition for 1D Diffusion-Reaction equation. Note that in this case the value of the scalar function is limited between 0 to 1 because of the form of the source term. 
Finally, we provided several samples of the 1D and 2D CFD cases in \autoref{fig:vis_1D_CFD_init} and \autoref{fig:vis_2D_CFD_init}.

\color{black}

% \clearpage
\FloatBarrier

\section{\projectname{}'s Data Sheet}\label{app:technical} 

In this section, we provide more detailed information about the DaRUS Dataverse 
 which \projectname{} uses to store its datasets. 
As explained in \autoref{sec:data_access}, 
DaRUS is the University of Stuttgart’s data repository based on the OpenSource Software DataVerse3, 
providing a place to archive, share, and publish the research data, scripts and source codes. 
All data uploaded to DaRUS gets a DOI as a persistent identifier, a license, and can be described with an extensive set of metadata, organized in metadata blocks. A dedicated team ensures that DARUS is continuously maintained.
\projectname{} obtained a permanent DOI (\href{https://doi.org/10.18419/darus-2986}{doi:10.18419/darus-2986}) \cite{PDEBenchDataset}
through DaRUS.

\subsection{Motivation}
\begin{enumerate}
    \item \textbf{For what purpose was the dataset created?} \textit{Was there a specific task in mind? Was there a specific gap that needed to be filled? Please provide a description.}
    
    \projectname{}'s datasets were created to provide a more challenging and representative benchmark for scientific ML approaches, that is, machine learning methods to approximately simulate physical systems governed by PDEs. More details can be found in \autoref{sec:motivation}.
    
    \item \textbf{Who created this dataset (e.g. which team, research group) and on behalf of which entity (e.g., company, institution, organization)?}
    
    This dataset was created by Makoto Takamoto and Francesco Alesiani (senior researchers at NEC Labs Europe), Dan MacKinlay (postdoctoral researcher at CSIRO's Data61), Timothy Praditia and Raphael Leiteritz (PhD students at the University of Stuttgart), Dirk Pfl{\"u}ger and Mathias Niepert (professors at the University of Stuttgart).
    
    \item \textbf{Who funded the creation of the dataset?} \textit{If there is an associated grant, please provide the name of the grantor and the grant name and number.)}
    
    Funding for the University of Stuttgart researchers was provided by the Deutsche Forschungsgemeinschaft (DFG, German Research Foundation) under Germany's Excellence Strategy - EXC 2075 – 390740016, for the Cluster of Excellence on ``Data-Integrated Simulation Science (SimTech)". 
    
    \item \textbf{Any other comments?}
    
    No
    
\end{enumerate}

\subsection{Composition}

\begin{enumerate}
    \item \textbf{What do the instances that comprise the dataset represent (e.g. documents, photos, people, countries)?} \textit{Are there multiple types of instances (e.g. movies, users, and ratings; people and interactions between them; nodes and edges)? Please provide a description.)}
    
    Each instance of the dataset is a simulation-generated array, containing the solutions (i.e. values of the variables of interest) of a specific partial differential equation (PDE). The details of the physical systems goverened by PDEs can be found in \autoref{sec:ap:prob_desc} and the format of the data can be found in \autoref{sec:ap:data_format}.
    
    \item \textbf{How many instances are there in total (of each type, if appropriate)?}
    An overview of the datasets and instances are provided in \autoref{tab:data_overview}. 
    For the 1D diffusion-sorption equation, there is one file containing $10\,000$ samples, with $1\,024$ spatial cells, $101$ time steps, and $1$ variable. For the 2D diffusion-reaction equation, there is one file containing $1\,000$ samples, with $128 \times 128$ spatial cells, $101$ time steps, and $2$ variables. For the 2D shallow-water equation, there is one file containing $1\,000$ samples, with $128 \times 128$ spatial cells, $101$ time steps, and $1$ variable.
    For the advection, Burgers, and 1D-diffusion-reaction equation, each file contains $10\,000$ samples, with $1\,024$ spatial cells, $201$ time steps, and $1$ variable. The advection equation has 6 files for different parameters (advection velocity), the Burgers equation has 12 files for different parameters (diffusion coefficient), and the 1D diffusion-reaction equationa has 16 files for different parameters (diffusion and source term coefficients). 
    For the DarcyFlow, each file contains $10\,000$ samples, with $128 \times 128$ spatial cells, $1$ time step, and $2$ variables. 
    For the incompressible Navier-Stokes equations, the file contains $1\,000$ samples, with $256 \times 256$ spatial cells, $1\,000$ time steps, and $2$ variables.
    For the 1D compressible Navier-Stokes equations, the file contains $10\,000$ samples, with $1\,024$ spatial cells, $100$ time steps, and $3$ variables. 
     For the 2D compressible Navier-Stokes equations, the file contains $1\,000$ samples, with $512 \times 512$ spatial cells, $21$ time steps, and $4$ variables. 
     For the 3D compressible Navier-Stokes equations, the file contains $1\,00$ samples, with $128 \times 128 \times 128$ spatial cells, $21$ time steps, and $5$ variables. 
     The variation of parameters, initial conditions, and boundary conditions of the compressible Navier-Stokes equations are provided in \autoref{sec:cfd-intro}. 
    
    \item \textbf{Does the dataset contain all possible instances or is it a sample (not necessarily random) of instances from a larger set?} \textit{If the dataset is a sample, then what is the larger set? Is the sample representative of the larger set (e.g. geographic coverage)? If so, please describe how this representativeness was validated/verified. If it is not representative of the larger set, please describe why not (e.g., to cover a more diverse range of instances, because instances were withheld or unavailable).)}
    
    The datasets that we provide are intended to be representative of challenging physical systems. Since this is a very broad and general scope, it is difficult, if not impossible, to represent it wholly. However, we provide datasets that contain longer time steps, non-linearly coupled variables, various challenging boundary conditions, different parameters, and more applicable to real-world scenarios than currently available. For more details, see \autoref{sec:data_overview} and \autoref{sec:ap:prob_desc}.
    
    \item \textbf{What data does each instance consist of?} \textit{("Raw" data (e.g. unprocessed text or images) or features? In either case, please provide a description.)}
    
    Each instance consists of numerical values in the form of an HDF5 file that correspond to the variables of interest in each specific physical system. See \autoref{sec:ap:prob_desc} for further details.
    
    \item \textbf{Is there a label or target associated with each instance? If so, please provide a description.}
    
    Each instance is an array, which elements are the values of a specific variable at a specific space and time (for the data format please refer to \autoref{sec:ap:data_format}). These values are the target for the regression task performed by the baseline models. We also provide the full information of physical parameters for each dataset. 
    
    \item \textbf{Is any information missing from individual instances?} \textit{(If so, please provide a description, explaining why this information is missing (e.g., because it was unavailable). This does not include intentionally removed information, but might include, e.g., redacted text.)}
    
    No.
    
    \item \textbf{Are relationships between individual instances made explicit (e.g., users' movie ratings, social network links)?} \textit{(If so, please describe how these relationships are made explicit.)}
    
    There are no relationships between different instances which were generated using randomly created initial conditions.
    
    \item \textbf{Are there recommended data splits (e.g., training, development/validation, testing)?} \textit{(If so, please provide a description of these splits, explaining the rationale behind them.}
    
    There are no recommendations for the data split. Users can experiment defining the data split as a hyperparameter. However, for the paper we use the split of $90\%$ for training and the rest for validation and testing.
    
    \item \textbf{Are there any errors, sources of noise, or redundancies in the dataset?} \textit{(If so, please provide a description.)}
    
    No. The only possible source of error might come from the discretization error, which should be minimal because our dataset was generated with a relatively high resolution. However, users can also generate their own dataset with the provided source code with even higher resolution.
    
    \item \textbf{Is the dataset self-contained, or does it link to or otherwise rely on external resources (e.g., websites, tweets, other datasets)?} \textit{(If it links to or relies on external resources, a) are there guarantees that they will exist, and remain constant, over time; b) are there official archival versions of the complete dataset (i.e., including the external resources as they existed at the time the dataset was created); c) are there any restrictions (e.g., licenses, fees) associated with any of the external resources that might apply to a future user? Please provide descriptions of all external resources and any restrictions associated with them, as well as links or other access points, as appropriate.)}
    
    The dataset is available via a data repository (DaRUS) \citep{dataverse}, which is maintained by the University of Stuttgart, and therefore will be available permanently.
    
    \item \textbf{Does the dataset contain data that might be considered confidential (e.g., data that is protected by legal privilege or by doctor-patient confidentiality, data that includes the content of individuals' non-public communications)?} \textit{(If so, please provide a description.)}
    
    No.
    
    \item \textbf{Does the dataset contain data that, if viewed directly, might be offensive, insulting, threatening, or might otherwise cause anxiety?} \textit{(If so, please describe why.)}
    
    No.
    
    \item \textbf{Does the dataset relate to people?} \textit{(If not, you may skip the remaining questions in this section.)}
    
    No.
    
    \item \textbf{Does the dataset identify any subpopulations (e.g., by age, gender)?} \textit{(If so, please describe how these subpopulations are identified and provide a description of their respective distributions within the dataset.)}
    
    N/A.
    
    \item \textbf{Is it possible to identify individuals (i.e., one or more natural persons), either directly or indirectly (i.e., in combination with other data) from the dataset?} \textit{(If so, please describe how.)}
    
    N/A.
    
    \item \textbf{Does the dataset contain data that might be considered sensitive in any way (e.g., data that reveals racial or ethnic origins, sexual orientations, religious beliefs, political opinions or union memberships, or locations; financial or health data; biometric or genetic data; forms of government identification, such as social security numbers; criminal history)?} \textit{(If so, please provide a description.)}
    
    N/A.
    
    \item \textbf{Any other comments?}
    
    No.
    
\end{enumerate}

\subsection{Collection Process}

\begin{enumerate}

    \item {\bf How was the data associated with each instance acquired?} (Was the data directly observable (e.g., raw text, movie ratings), reported by subjects (e.g., survey responses), or indirectly inferred/derived from other data (e.g., part-of-speech tags, model-based guesses for age or language)? If data was reported by subjects or indirectly inferred/derived from other data, was the data validated/verified? If so, please describe how.)

    All the dataset were obtained by performing computationally expensive numerical simulations for each PDEs. 

    \item {\bf What mechanisms or procedures were used to collect the data (e.g., hardware apparatus or sensor, manual human curation, software program, software API)?} (How were these mechanisms or procedures validated?)

    The numerical simulations were performed either using CPUs or GPUs with freely available Python libraries, such as Scipy. 

    \item {\bf If the dataset is a sample from a larger set, what was the sampling strategy (e.g., deterministic, probabilistic with specific sampling probabilities)?}

    This is already mentioned above (F.2 "Composition"'s 3rd item). 

    \item {\bf Who was involved in the data collection process (e.g., students, crowdworkers, contractors) and how were they compensated (e.g., how much were crowdworkers paid)?}

    All the data generation process (numerical simulation) was performed by the authors, mainly conducted by Makoto Takamoto, Timothy Praditia, Raphael Leiteritz, Dan MacKinlay and Francesco Alesiani. For a part of the simulations we received help by people at ANU Techlauncher who are acknowledged in "Acknowledgements". 

   \item {\bf Over what timeframe was the data collected?} (Does this timeframe match the creation timeframe of the data associated with the instances (e.g., recent crawl of old news articles)? If not, please describe the timeframe in which the data associated with the instances was created.)

    The dataset was generated from early to mid 2022. Because of the eternal nature of PDEs, our dataset is irrelevant to the timeframe of creation.  

    \item {\bf Were any ethical review processes conducted (e.g., by an institutional review board)?} (If so, please provide a description of these review processes, including the outcomes, as well as a link or other access point to any supporting documentation.)

    No.

    \item {\bf Does the dataset relate to people?} (If not, you may skip the remaining questions in this section.)

    No

    \item {\bf Did you collect the data from the individuals in question directly, or obtain it via third parties or other sources (e.g., websites)?}

    No.

    \item {\bf Were the individuals in question notified about the data collection?} (If so, please describe (or show with screenshots or other information) how notice was provided, and provide a link or other access point to, or otherwise reproduce, the exact language of the notification itself.)

    N/A.

    \item {\bf Did the individuals in question consent to the collection and use of their data?} (If so, please describe (or show with screenshots or other information) how consent was requested and provided, and provide a link or other access point to, or otherwise reproduce, the exact language to which the individuals consented.)

    N/A.

    \item {\bf If consent was obtained, were the consenting individuals provided with a mechanism to revoke their consent in the future or for certain uses?} (If so, please provide a description, as well as a link or other access point to the mechanism (if appropriate).)

    N/A.

    \item {\bf Has an analysis of the potential impact of the dataset and its use on data subjects (e.g., a data protection impact analysis) been conducted?} (If so, please provide a description of this analysis, including the outcomes, as well as a link or other access point to any supporting documentation.)

    N/A.

    \item Any other comments?

    None.

\end{enumerate}

\subsection{Preprocessing/cleaning/labeling}

\begin{enumerate}
    \item {\bf Was any preprocessing/cleaning/labeling of the data done (e.g., discretization or bucketing, tokenization, part-of-speech tagging, SIFT feature extraction, removal of instances, processing of missing values)?} (If so, please provide a description. If not, you may skip the remainder of the questions in this section.)

    No. At most we rejected meaningless solutions because of the failure of numerical simulations, such as data with all zeros or Nan values. 

    \item {\bf Was the "raw" data saved in addition to the preprocessed/cleaned/labeled data (e.g., to support unanticipated future uses)?} (If so, please provide a link or other access point to the "raw" data.)

    N/A. 

    \item {\bf Is the software used to preprocess/clean/label the instances available?} (If so, please provide a link or other access point.)

    N/A.

    \item {\bf Any other comments?}

    None.

\end{enumerate}

\subsection{Uses}

\begin{enumerate}
    \item {\bf Has the dataset been used for any tasks already?} (If so, please provide a description.)

    Not yet. %A part of the authors are going to use the dataset to test the ML models ability to fit out-of-distribution data in terms of the physical parameters. 
    
    \item {\bf Is there a repository that links to any or all papers or systems that use the dataset?} (If so, please provide a link or other access point.)

    No.

    \item {\bf What (other) tasks could the dataset be used for?}

    The dataset could possibly be used for developing or testing ML models for fitting the out-of-distribution data, such as unseen physical parameters.

    \item {\bf Is there anything about the composition of the dataset or the way it was collected and preprocessed/cleaned/labeled that might impact future uses?} (For example, is there anything that a future user might need to know to avoid uses that could result in unfair treatment of individuals or groups (e.g., stereotyping, quality of service issues) or other undesirable harms (e.g., financial harms, legal risks) If so, please provide a description. Is there anything a future user could do to mitigate these undesirable harms?)

    Basically no. But high-dimensional data, in particular 3D data, has a relatively small number of samples, only 100, due of storage limitations and the very large   size of single data points. This might result in ML models with insufficient accuracy, but it is realistic for scientific simulations, where extensive computation time can be required for a single data point. However, we emphasize that we provide the data generation source code. This allows future users to increase their dataset size as much as they need. 

    \item {\bf Are there tasks for which the dataset should not be used?} (If so, please provide a description.)

    No. %(\textcolor{red}{should we mention that we do not want our dataset to be used for developing ML models with inhumanity purpose, such as developing weapon??})

    \item {\bf Any other comments?}

    None.

\end{enumerate}

\subsection{Distribution}

\begin{enumerate}
    \item  {\bf Will the dataset be distributed to third parties outside of the entity (e.g., company, institution, organization) on behalf of which the dataset was created?} (If so, please provide a description.)

    Yes, the dataset is freely and publicly available and accessible.

    \item {\bf How will the dataset will be distributed (e.g., tarball on website, API, GitHub)?} (Does the dataset have a digital object identifier (DOI)?)

    The dataset is free for download either by directly accessing the DaRUS page 
    \url{https://doi.org/10.18419/darus-2986}
    or making use of a Python API, "pyDaRUS", provided by DaRUS. 
    An example of a code snippet for downloading the data is listed in \autoref{sec:data_access}.

    \item {\bf When will the dataset be distributed?}

    The dataset is distributed as of June 2022 in its first version.

    \item {\bf Will the dataset be distributed under a copyright or other intellectual property (IP) license, and/or under applicable terms of use (ToU)?} (If so, please describe this license and/or ToU, and provide a link or other access point to, or otherwise reproduce, any relevant licensing terms or ToU, as well as any fees associated with these restrictions.)

    %\textcolor{red}{%
    The dataset is licensed under CC BY license.
    %}%

    \item {\bf Have any third parties imposed IP-based or other restrictions on the data associated with the instances?} (If so, please describe these restrictions, and provide a link or other access point to, or otherwise reproduce, any relevant licensing terms, as well as any fees associated with these restrictions.)

    No

    \item {\bf Do any export controls or other regulatory restrictions apply to the dataset or to individual instances?} (If so, please describe these restrictions, and provide a link or other access point to, or otherwise reproduce, any supporting documentation.)

    No

    \item {\bf Any other comments?}

    None.

\end{enumerate}

\subsection{Maintenance}

\begin{enumerate}
    \item {\bf Who is supporting/hosting/maintaining the dataset?}
    
    The storage infrastructure, DaRUS, is maintained by the University of Stuttgart and a dedicated team of DaRUS. 
    The dataset themselves are maintained by the authors. 

    \item {\bf How can the owner/curator/manager of the dataset be contacted (e.g., email address)?}

    E-mail addresses are provided in the main-body of the paper.

    \item {\bf Is there an erratum?} (If so, please provide a link or other access point.)

    Currently, no. As errors are encountered, future versions of the dataset may be released. They will all be provided in the same github and DaRUS location.

    \item {\bf Will the dataset be updated (e.g., to correct labeling errors, add new instances, delete instances')?} (If so, please describe how often, by whom, and how updates will be communicated to users (e.g., mailing list, GitHub)?)

    Same as previous.

    \item {\bf If the dataset relates to people, are there applicable limits on the retention of the data associated with the instances (e.g., were individuals in question told that their data would be retained for a fixed period of time and then deleted)?} (If so, please describe these limits and explain how they will be enforced.)

    N/A. 

    \item {\bf Will older versions of the dataset continue to be supported/hosted/maintained?} (If so, please describe how. If not, please describe how its obsolescence will be communicated to users.)

    % \textcolor{red}{Yes; all data will be versioned. (I think the version will be of the full dataset.??)}
    
    Versioning of the dataset is managed via DaRUS. Data can only be added. For additional information see  \url{https://guides.dataverse.org/en/5.9/user/dataset-management.html#dataset-versions}

    \item {\bf If others want to extend/augment/build on/contribute to the dataset, is there a mechanism for them to do so?} (If so, please provide a description. Will these contributions be validated/verified? If so, please describe how. If not, why not? Is there a process for communicating/distributing these contributions to other users? If so, please provide a description.)
    
    Errors may be submitted via the bugtracker on github. 

    \item {\bf Any other comments?}

    None.
\end{enumerate}

\subsection{Reproducibility of the baseline score}

We provide all the pretrained model in our DaRUS repository:  \url{https://darus.uni-stuttgart.de/privateurl.xhtml?token=cd862f8c-8e1b-49d2-b4da-b35f8df5ac85}.
So users can freely check our result or even continue longer training with the help of our APIs. 

\subsection{Reading and using the dataset}

The downloaded HDF5 files can easily be read by a standard software to read HDF5 files. 
Concerning the data size, 1D data have 4 - 20 GB, 2D data have 6 - 100 GB, 3D data have 60 - 80 GB, depending on PDEs and parameters. 
In our project home page \url{https://github.com/pdebench/PDEBench/blob/main/README.md}, 
we also provide APIs to read and store the data into PyTorch data loader, 
which could be directly used in the user's code or our providing ML model training APIs. 

An example of snipped code for reading data is introduced in \autoref{sec:data_access}. 
Information on how to use the baseline and read the dataset is provided in the project home page.

\begin{figure}[t]
    \centering
    \includegraphics[width=0.9\linewidth]{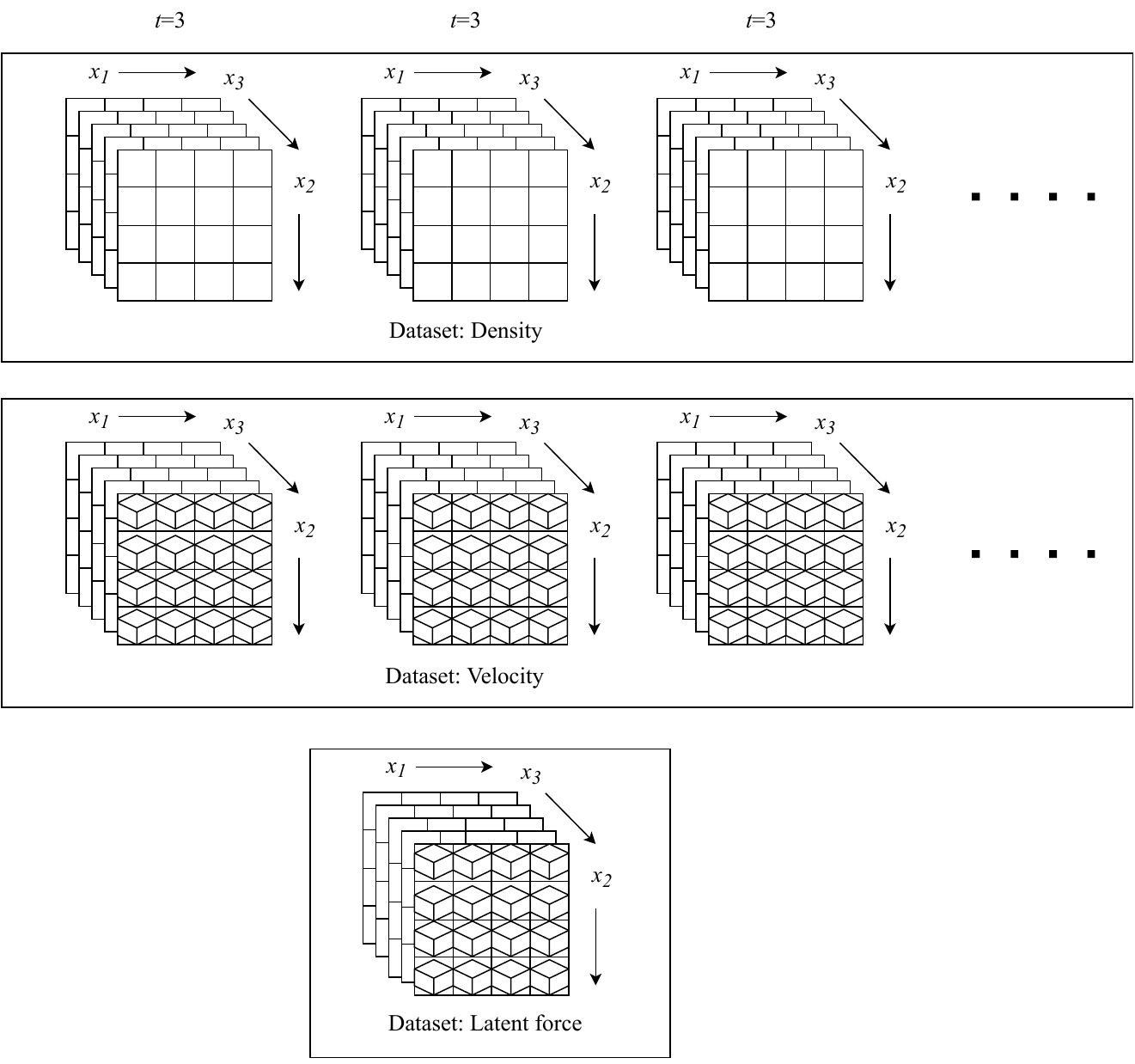}
    \caption{Example of data format for a PDE with density and velocity terms}
    \label{fig:dataformat}
\end{figure}

\subsection{Data Format}
\label{sec:ap:data_format}
\dan{move to `README.md`}

Fig.\ref{fig:dataformat} visualizes the data format of \projectname{}.

Realizations are stored in HDF5 format, allowing typed storage of dense arrays on disk.
File naming format is \texttt{\$\{PDE name\}{-}{-}\$\{Parameters\}{-}{-}\{\$Config\}.h5}.
The HDF file contains a single group, documented in the source code for the generating PDE, which may contain multiple HDF5 datasets, corresponding to tensors.

%Code includes HTTP download URLs.

Arrays are packed according to the convention
\((b\times t  \times x_1 \dots \times x_d \times v)\), where \(b\) denotes the samples' dimension, \(t\) the time dimension, \(x_1, \dots, x_d\) the \(d\) spatial dimensions, and \(v\) the dimensions needed to encode the state of the system (e.g. a scalar density field has dimension one, a 2d velocity field might have dimension 2)\footnote{%
  Concerning the compressible Navier-Stokes equations' dataset, each variable is provided independently, not summarized into a channel dimension (i.e. extending the array by $1$ dimension), such as "density" and "pressure". This is to avoide too big array sizes, allowing memory efficient data I/O by sub-sampling spatial cells. 
}%
.
Not all dimensions are present in all data sets; for example, some parameters are be time-invariant, in which case the time dimension does not exist.
For each file, simulation parameters are stored alongside the simulation runs as HDF5  attribute strings in YAML format encoded as UTF8.

% \bibliographystyleweb{plainnat}
% \bibliographyweb{web}
% \bibliographystyle{abbrvnat}
% \bibliography{physics_benchmark}

% Following lines for the bib for the supplementary material
% \bibliographystyle{abbrvnat}
% \bibliography{physics_benchmark}

\end{document}